\documentclass[twoside]{article}

\usepackage[accepted]{aistats2019}
\usepackage{tikz}
\tikzset{mynode/.style={draw,circle, minimum size = 0.7cm}}
\usetikzlibrary{positioning,shapes,arrows}
\usepackage{subcaption}

\usepackage{bbm}
\usepackage{amsmath,amssymb,amsfonts,mathrsfs}
\usepackage{algorithmicx}
\usepackage{algpseudocode}
\usepackage{algorithm}

\usepackage{wrapfig}
\usepackage[utf8]{inputenc}
\usepackage[amsmath,thmmarks]{ntheorem}

\usepackage[round]{natbib}

\begin{document}
%% Theorem-like environments

%% This can be changed according to language. You can comment out the ones you
%% don't need.

%\numberwithin{equation}{chapter}

%% German theorems
%\newtheorem{satz}{Satz}[chapter]
%\newtheorem{beispiel}[satz]{Beispiel}
%\newtheorem{bemerkung}[satz]{Bemerkung}
%\newtheorem{korrolar}[satz]{Korrolar}
%\newtheorem{definition}[satz]{Definition}
%\newtheorem{lemma}[satz]{Lemma}
%\newtheorem{proposition}[satz]{Proposition}

%% English variants
\newtheorem{theorem}{Theorem}
\newtheorem{example}[theorem]{Example}
\newtheorem{remark}[theorem]{Remark}
\newtheorem{corollary}[theorem]{Corollary}
\newtheorem{definition}[theorem]{Definition}
\newtheorem{lemma}[theorem]{Lemma}
\newtheorem{proposition}[theorem]{Proposition}

%% Proof environment with a small square as a "qed" symbol
\theoremstyle{nonumberplain}
\theorembodyfont{\normalfont}
\theoremsymbol{\ensuremath{\square}}
\newtheorem{proof}{Proof}
%\newtheorem{beweis}{Beweis}

% If your paper is accepted and the title of your paper is very long,
% the style will print as headings an error message. Use the following
% command to supply a shorter title of your paper so that it can be
% used as headings.
%
\runningtitle{FGPGM for parameter identification in systems of nonlinear ODEs}

% If your paper is accepted and the number of authors is large, the
% style will print as headings an error message. Use the following
% command to supply a shorter version of the authors names so that
% they can be used as headings (for example, use only the surnames)
%
%\runningauthor{Surname 1, Surname 2, Surname 3, ...., Surname n}

\twocolumn[

\aistatstitle{Fast Gaussian process based gradient matching for parameter identification in systems of nonlinear ODEs}

\aistatsauthor{Philippe Wenk$^{1, 2}$ \And Alkis Gotovos$^1$ \And  Stefan Bauer$^{2, 3}$}
\aistatsauthor{Nico S. Gorbach$^1$ \And Andreas Krause$^1$ \And Joachim M. Buhmann$^1$}
\runningauthor{Wenk, Gotovos, Bauer, Gorbach, Krause and Buhmann}
\aistatsaddress{$^1$ETH Zürich \And $^2$Max Planck ETH CLS \And  $^3$ MPI-IS Tübingen} ]

\begin{abstract}
Parameter identification and comparison of dynamical systems is a challenging task in many fields. Bayesian approaches based on Gaussian process regression over time-series data have been successfully applied to infer the parameters of a dynamical system without explicitly solving it. While the benefits in computational cost are well established, the theoretical foundation has been criticized in the past. We offer a novel interpretation which leads to a better understanding and improvements in state-of-the-art performance in terms of accuracy, robustness and a decrease in run time due to a more efficient setup for general nonlinear dynamical systems.
\end{abstract}

\section{INTRODUCTION}
The underlying mechanism of many processes in science and engineering can often be described by ordinary differential equations (ODE). While the form of dynamical systems, the ODEs, can often be derived using expert knowledge, the parameters are usually unknown, can not be directly measured and have to be estimated from empirical time series data.  Since nonlinear ODEs typically do not have a closed form solution, standard methods for statistical inference require the computationally expensive numerical integration of the ODEs every time the parameters are changed \citep{Calderhead}.

To circumvent the high computational cost of numerical integration, gradient matching techniques have been proposed \citep[e.g.][]{ramsay2007, Dondelinger, niu2016fast, VGM}. Gradient matching is based on minimizing the difference between a model interpolating the dynamics of the state variables and the time derivatives provided by the ODEs. The first steps of this approach go back to spline-based methods, with an overview given by \citet{ramsay2007}. \citet{Calderhead} then proposed a fully probabilistic treatment by using Gaussian process models, increasing data efficiency and providing final uncertainty estimates. To match the gradients of the GP model to the gradients provided by the ODEs, \citet{Calderhead} introduce a product of experts heuristics (PoE). This heuristic has since become the state-of-the-art explanation, being reused, e.g., in \citet{Dondelinger} and \citet{VGM}. The advantages of the method have been well established, with applications in biomedical domains e.g., systems biology or neuroscience  \citep{babtie2014topological, macdonald2015gradient, pfister2018identifying}. Given these applications, a solid understanding and theoretical framework of theses methods is of critical importance.

However, there are two important open issues. Regarding the theoretical framework, there has been some criticism regarding the use of the PoE in this context, e.g., by \citet{WangAndBarber}, who introduced a different modeling paradigm. Regarding empirical performance, \citet{VGM} recently introduced a mean field variational scheme to decrease the run time. However, despite the expected run time savings, \citet{VGM} also reported a significantly increased accuracy of the variational approach compared to the MCMC sampling based counterparts.

While the criticism of the product of experts approach \citep{WangAndBarber} lead to the controversy of mechanistic modeling with Gaussian processes \citep{ControversyPaper} where the theoretical inconsistencies of the modeling proposal of \citet{WangAndBarber} have been outlined, a similar analysis  investigating the theoretical foundation of the state of the art approach underlying all current modeling proposals \citep{Calderhead, Dondelinger, VGM} has been missing.\\

\noindent
\textbf{Our contributions.} In this work, we
%\vspace{-1em}
\begin{enumerate}
\setlength{\itemsep}{1pt}
        \item analyze the Product of Experts (PoE) heuristic, discovering and explaining theoretical inconsistencies of the state of the art approach,
        \item provide a theoretical framework, replacing the criticized product of experts heuristic,
        \item identify the cause of the performance gains of the variational approach of \citet{VGM} over sampling-based methods,
        \item combine these insights to create a novel algorithm improving on state-of-the-art performance for nonlinear systems in terms of accuracy and robustness, while providing a more computationally efficient sampling scheme reducing its run time by roughly 35\%.\footnote{Code publicly available at\\ \hspace*{0.5cm} https://github.com/wenkph/FGPGM/}
\end{enumerate}

\section{PRELIMINARIES}
\subsection{Problem Formulation}
\label{sec:ProblemFormulation}
In this work, we analyze an arbitrary system of parametric ordinary differential equations, whose state derivatives can be parametrized by a time independent parameter vector $\boldsymbol{\theta}$. In this setting, the true evolution of the dynamical system is given by
\begin{equation}
\label{eq:ODEModel}
\mathbf{\dot{x}} = \mathbf{f}(\mathbf{x}, \boldsymbol{\theta})
\end{equation}
where $\mathbf{x}$ are the time dependent states of the system and $\mathbf{f}$ is an arbitrary, potentially nonlinear vector valued function. 

While Equation \eqref{eq:ODEModel} is meant to represent the dynamics of the system at all time points, $\mathbf{y}$, $\mathbf{x}$ and $\mathbf{\dot{x}}$ will be used throughout this paper to denote the vector containing the time evolution of one state or one state derivative at the observation times $\mathbf{t}$, i.e. $\mathbf{x} = [ x_0(t_0), ..., x_0(t_N)]$ etc.

At N discrete time points $\mathbf{t}$, the state trajectories are observed with additive, i.i.d. Gaussian noise $\boldsymbol{\epsilon}(t_i) \sim \mathcal{N}(\mathbf{0}, \sigma^2 \boldsymbol{I})$, i.e.,
\begin{equation}
\mathbf{y}(t_i) = \mathbf{x}(t_i) + \boldsymbol{\epsilon}(t_i), \qquad i=1\dots \textrm{N},
\end{equation}
or equivalently
\begin{equation}
\label{eq:ObsModel}
p(\mathbf{y} | \mathbf{x}, \sigma) = \mathcal{N}(\mathbf{y} | \mathbf{x}, \sigma^2 \boldsymbol{I}).
\end{equation}
Given these noisy observations $\mathbf{y}$ and the functional form of $\mathbf{f}$, the goal is to infer the true parameters $\boldsymbol{\theta}$. %\subsubsection{Notation}

For the sake of clarity, we present the theory for a one-dimensional system only and assume that all observations were created using one realization of the experiment. The extension to multidimensional systems (as done in the experiments, Section \ref{sec:Experiments}) or repeated experiments is straightforward but omitted to simplify the notation. 

\subsection{Modeling}
\label{sec:Modeling}

The key idea of GP-based gradient matching is to build a GP regression model mapping the time points to the corresponding state values. For this, one needs to choose an appropriate kernel function $k_\phi(t_i, t_j)$, which is parametrized by the hyperparameters $\boldsymbol{\phi}$. Both, the choice of kernels as well as how to fit its hyperparameters is discussed in Section \ref{sec:HyperAndKernelSelection}.

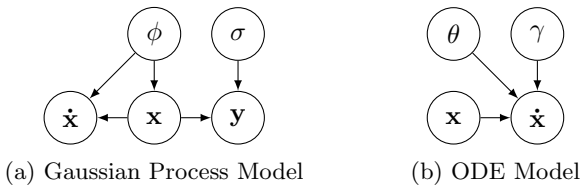
\begin{figure}[h]%{.5\textwidth}
    \begin{tabular}[c]{cc}
        \begin{subfigure} [t]{.24\textwidth}
            \centering
            \begin{tikzpicture}[
            node distance=0.4 cm and 0.4 cm,
            ]
            \node[mynode] (phi) {$\phi$};
            \node[mynode, below=of phi](x){$\mathbf{x}$};
            \node[mynode, left=of x] (xDot){$\mathbf{\dot{x}}$};
            \node[mynode, right=of x](y){$\mathbf{y}$};
            \node[mynode, above=of y](sigma){$\sigma$};
            \path
            (phi) edge[-latex] (xDot)
            (x) edge[-latex] (xDot)
            (phi) edge[-latex] (x)
            (x) edge[-latex] (y)
            (sigma) edge[-latex] (y)
            ;
            \end{tikzpicture}
            \caption{Gaussian Process Model}
            \label{subfig:GPModel}
        \end{subfigure}&
        \begin{subfigure}[t]{.24\textwidth}
            \centering
            \begin{tikzpicture}[
            node distance=0.4 cm and 0.4 cm,
            ]
            \node[mynode] (xDot){$\mathbf{\dot{x}}$};
            \node[mynode, left=of xDot](x){$\mathbf{x}$};
            \node[mynode, above=of xDot](gamma){$\gamma$};
            \node[mynode, left=of gamma] (theta) {$\theta$};
            \path
            (x) edge[-latex] (xDot)
            (gamma) edge[-latex] (xDot)
            (theta) edge[-latex] (xDot)
            ;
            \end{tikzpicture}
            \caption{ODE Model}
            \label{subfig:ODEModel}
        \end{subfigure}
    \end{tabular}
    \centering
    \caption{Modeling assumptions of Gaussian Process based gradient matching.}
    \label{fig:CalderheadModelInitial}
\end{figure}

Once the kernel and its hyperparameters are fixed, the covariance matrix $\mathbf{C}_\phi$, whose elements are given by $\mathbf{C}_\phi(i, j) = k(t_i, t_j)$, can be constructed and used to define a standard zero mean Gaussian process prior on the states:
\begin{equation}
\label{eq:GPPrior}
p(\mathbf{x} | \boldsymbol{\phi}) = \mathcal{N}(\mathbf{x} | \boldsymbol{0}, \mathbf{C}_\phi).
\end{equation}
As differentiation is a linear operation, the derivative of a Gaussian process is again a Gaussian process. Using probabilistic calculus (see supplementary material, Section \ref{sec:AppendixDerivs} for details), this fact leads to a distribution over the derivatives conditioned on the states at the observation points:
\begin{equation}
\label{eq:GPDerivs}
p(\dot{\mathbf{x}} | \mathbf{x}, \boldsymbol{\phi}) = \mathcal{N}(\dot{\mathbf{x}} | \mathbf{D} \mathbf{x}, \mathbf{A}).
\end{equation}
Lastly, the information provided by the differential equation is used as well. For known states and parameters, one can calculate the derivatives using equation \eqref{eq:ODEModel}. A potential modeling mismatch between the output of the ODEs and the derivatives of the GP model is accounted for by introducing isotropic Gaussian noise with standard deviation $\gamma$, leading to the following Gaussian distribution over the derivatives:
\begin{equation}
\label{eq:ODEDerivs}
p(\dot{\mathbf{x}} |\mathbf{x}, \boldsymbol{\theta}, \gamma)  = \mathcal{N} (\dot{\mathbf{x}} | f(\mathbf{x}, \boldsymbol{\theta}), \gamma \boldsymbol{I}).
\end{equation}
The modeling assumptions are summarized in the graphical models shown in Figure \ref{fig:CalderheadModelInitial}.

\subsection{Inference}
\label{sec:TraditionalInference}
As stated in Section \ref{sec:ProblemFormulation}, the main goal of the inference process is to learn the parameters $\boldsymbol{\theta}$ using the noisy observations $\mathbf{y}$. Thus, it is necessary to connect the two graphical models shown in Figure \ref{fig:CalderheadModelInitial}. As shown in the previous section,  $\mathbf{x}$ and the $\mathbf{\dot{x}}$ represent the same variables in both models. However, it is not straightforward to  use this fact to combine the two. While it is intuitive to use the probability density over $\mathbf{x}$ of the Gaussian process model directly as the prior for $\mathbf{x}$ in the ODE response model, handling $\mathbf{\dot{x}}$ is more challenging. In both models, $\mathbf{\dot{x}}$ is a dependent variable. Thus, some heuristic is needed to combine the two conditional distributions $p(\dot{\mathbf{x}} | \mathbf{x}, \boldsymbol{\phi})$ and $p(\dot{\mathbf{x}} |\mathbf{x}, \boldsymbol{\theta}, \gamma)$.

\subsubsection{Product of experts heuristic}
The main idea of the product of experts, originally introduced by \citet{ProductOfExperts}, is to infer the probability density of a variable by normalizing the product of multiple expert densities. \citet{Calderhead} use this to connect the two distributions over $\mathbf{\dot{x}}$, leading to
\begin{equation}
\label{eq:PoE}
p(\dot{\mathbf{x}} | \mathbf{x}, \boldsymbol{\phi}, \boldsymbol{\theta}, \gamma) \propto p(\dot{\mathbf{x}} | \mathbf{x}, \boldsymbol{\phi})p(\dot{\mathbf{x}} |\mathbf{x}, \boldsymbol{\theta}, \gamma)
\end{equation}
The idea of this approach is that the resulting density only assigns high probability if both experts assign high probabilities. Hence, it considers only cases in which both experts agree. It is thus based on the intuition that the true $\boldsymbol{\theta}$ should correspond to a model that agrees both with the ODE model and the observed data. While this is intuitively well-motivated, we will show that the product of experts heuristic leads to theoretical difficulties and offer an alternative.

\subsubsection{Markov Chain Monte Carlo based methods}
\citet{Calderhead} combine the product of experts with equations \eqref{eq:ObsModel}, \eqref{eq:GPPrior} and \eqref{eq:GPDerivs} and some suitable prior over $\boldsymbol{\theta}$ to obtain a joint distribution $p(\mathbf{x}, \mathbf{\dot{x}}, \boldsymbol{\theta}, \boldsymbol{\phi}, \sigma | \mathbf{y})$. After integrating out $\mathbf{\dot{x}}$, which can be done analytically since Gaussian processes are closed under linear operators (and using some proportionality arguments), a sampling scheme was derived that consists of two MCMC steps. First, the hyperparameters of the GP, $\boldsymbol{\phi}$ and $\sigma$, are sampled from the conditional distribution $p(\boldsymbol{\phi}, \sigma | \mathbf{y})$. Then, a second MCMC scheme is deployed to infer the parameters of the ODE model, $\boldsymbol{\theta}$ and $\gamma$, by sampling from the conditional distribution $p(\boldsymbol{\theta}, \gamma | \mathbf{x}, \boldsymbol{\phi}, \sigma)$.

\citet{Dondelinger} then reformulated the approach by directly calculating the joint distribution
\begin{align}
p(\mathbf{y}, \mathbf{x}, \boldsymbol{\theta}, \boldsymbol{\phi}, \gamma, \sigma) &\propto
p(\boldsymbol{\theta})\nonumber\\
&\times \mathcal{N}(\mathbf{x} | \mathbf{0}, \boldsymbol{C_\phi})\nonumber\\
&\times \mathcal{N}(\mathbf{y} | \mathbf{x}, \sigma^2 \mathbf{I})\nonumber\\
&\times \mathcal{N}(\mathbf{f}(\mathbf{x}, \boldsymbol{\theta}) | \mathbf{D} \mathbf{x}, \mathbf{A} + \gamma \mathbf{I}),\label{eq:DondelingerDensity}
\end{align}
where the proportionality is meant to be taken w.r.t. the latent states $\mathbf{x}$ and the ODE parameters $\boldsymbol{\theta}$. Here $p(\boldsymbol{\theta})$ denotes some prior on the ODE parameters. This approach was named Adaptive Gradient Matching ($\mathrm{AGM}$).

\subsubsection{Variational inference}
The main idea of Variational Gradient Matching ($\mathrm{VGM}$), introduced by \citet{VGM}, is to substitute the MCMC inference scheme of $\mathrm{AGM}$ with a mean field variational inference approach, approximating the density in Equation \eqref{eq:DondelingerDensity} with a fully factorized Gaussian over the states $\mathbf{x}$ and the parameters $\boldsymbol{\theta}$. To obtain analytical solutions, the functional form of the ODEs is restricted to locally linear functions that could be written as
\begin{equation}
\label{eq:VGMFunctionalForm}
f(\mathbf{x}, \boldsymbol{\theta}) = \sum_{i} \theta_i \prod_{j \in \mathcal{M}_i} x_j \qquad \mathrm{where} \quad \mathcal{M}_i \subseteq \{1, ..., K\}.
\end{equation}
As perhaps expected, $\mathrm{VGM}$ is magnitudes faster than the previous sampling approaches. However, despite being a variational approach, $\mathrm{VGM}$ was also able to provide significantly more accurate parameter estimates than both sampling-based approaches of \citet{Calderhead} and \citet{Dondelinger}. In Section \ref{sec:OwnAlgorithm}, we provide justification for these surprising performance differences.

%\subsection{Gradient Matching Without Product of Experts}
%Up to our knowledge, there was only one approach that tried to do Gaussian Process based gradient matching without the product of experts formulation, namely \cite{WangAndBarber}. However, this approach had its own theoretical problems, which were intensively discussed in \cite{ControversyPaper}.

\section{THEORY}
\label{sec:Theory}
\newcommand{\gp}{\textrm{GP}}
\newcommand{\ode}{\textrm{ODE}}

\subsection{Analysis Of The Product Of Experts Approach}
\label{sec:ProblemsWithPoE}
As previously stated, \citet{Calderhead}, \citet{Dondelinger} and \citet{VGM} all use a product of experts to obtain $p(\dot{\mathbf{x}} | \mathbf{x}, \boldsymbol{\phi}, \boldsymbol{\theta}, \gamma)$ as stated in Equation \eqref{eq:PoE}.

In this section, we will first provide an argument based on graphical models and then an argument based on the original mathematical derivation to illustrate challenges arising from this heuristic.  

\begin{figure*}[!h]
    \centering
    \begin{tabular}[c]{cc}
        \begin{subfigure} [t]{.5\textwidth}
            \centering
            \begin{tikzpicture}[
            node distance=0.5 cm and 0.5 cm,
            mynode/.style={draw,circle,minimum size=0.9cm}  % needs to be that big for F_1 to fit
            ]%[
            %node distance=0.4 cm and 0.4 cm,
            %]
            \node[mynode] (phi) {$\phi$};
            \node[mynode, below=of phi](x){$\mathbf{x}$};
            \node[mynode, left=of x] (xDot){$\mathbf{\dot{x}}$};
            \node[mynode, right=of x](y){$\mathbf{y}$};
            \node[mynode, above=of y](sigma){$\sigma$};
            \node[mynode, above=of xDot](gamma){$\gamma$};
            \node[mynode, left=of xDot] (theta) {$\theta$};
            \path
            (phi) edge[-latex] (xDot)
            (x) edge[-latex] (xDot)
            (phi) edge[-latex] (x)
            (x) edge[-latex] (y)
            (sigma) edge[-latex] (y)
            (theta) edge[-latex] (xDot)
            (gamma) edge[-latex] (xDot)
            ;
            \end{tikzpicture}
            \caption{After Product of Experts}
            \label{subfig:AfterPoE}
        \end{subfigure}& 
        \begin{subfigure} [t]{.5\textwidth}
            \centering
            \begin{tikzpicture}[
            node distance=0.5 cm and 0.5 cm,
            mynode/.style={draw,circle,minimum size=0.9cm}  % needs to be that big for F_1 to fit
            ]%[
            %node distance=0.4 cm and 0.4 cm,
            %]
            \node[mynode] (phi) {$\phi$};
            \node[mynode, below=of phi](x){$\mathbf{x}$};
            \node[mynode, right=of x](y){$\mathbf{y}$};
            \node[mynode, above=of y](sigma){$\sigma$};
            \node[mynode, left=of phi](gamma){$\gamma$};
            \node[mynode,  on grid, below left=1.1cm and .9cm of gamma, very thick,draw=red] (theta) {$\theta$};
            \path
            (phi) edge[-latex] (x)
            (x) edge[-latex] (y)
            (sigma) edge[-latex] (y)
            ;
            \end{tikzpicture}
            \caption{After Marginalizing $\mathbf{\dot{x}}$}
            \label{subfig:AfterMargin}
        \end{subfigure}
    \end{tabular}
    \caption{The product of experts approach as a graphical model. After marginalization of $\mathbf{\dot{x}}$, the parameters $\boldsymbol{\theta}$ we would like to infer are independent of the observations $\mathbf{y}$.}
    \label{fig:PoEDestroyer}
\end{figure*}
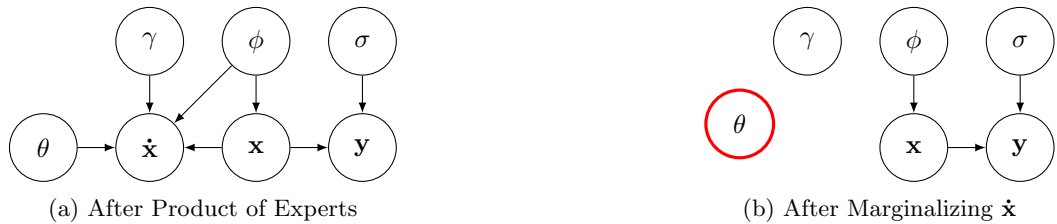

Figure \ref{fig:PoEDestroyer} depicts what is happening if the product of experts approach is applied in the gradient matching framework. Figure \ref{subfig:AfterPoE} depicts the graphical model after the two models have been merged using the product of experts heuristic of Equation \eqref{eq:PoE}. Using the distribution over $\mathbf{x}$ of the Gaussian process model \ref{subfig:GPModel} as a prior for the $\mathbf{x}$ in the ODE response model \ref{subfig:ODEModel}, effectively leads to merging the two nodes representing $\mathbf{x}$.  Furthermore, the product of experts heuristic implies by its definition that after applying Equation \eqref{eq:PoE}, $\mathbf{\dot{x}}$ is only depending on $\mathbf{x}$, $\boldsymbol{\phi}$, $\boldsymbol{\theta}$ and $\gamma$.

In the graphical model in Figure \ref{subfig:AfterPoE}, the problem is already visible. The ultimate goal of merging the two graphical models is to create a probabilistic link between the observations $\mathbf{y}$ and the ODE parameters $\boldsymbol{\theta}$. However, the newly created connection between these two variables is given via $\mathbf{\dot{x}}$, which has no outgoing edges and of which no observations are available. Marginalizing out $\mathbf{\dot{x}}$ as proposed in the traditional approaches consequently leads to the graphical model in Figure \ref{subfig:AfterMargin}. As there is no directed path connecting other variables via $\mathbf{\dot{x}}$, all the different components are now independent. Consequently, the posterior over $\boldsymbol{\theta}$ is now given by the prior we put on $\boldsymbol{\theta}$ in the first place.

This problem can further be illustrated by the mathematical derivations in the original paper of \citet{Calderhead}. After calculating the necessary normalization constants, the last equation in the third chapter is equivalent to stating
\begin{equation}
p(\boldsymbol{\theta}, \gamma | \mathbf{x}, \phi, \sigma) = \int p(\boldsymbol{\theta})p(\gamma)p(\dot{\boldsymbol{x}} | \mathbf{x}, \boldsymbol{\theta}, \gamma, \phi, \sigma) d\dot{\mathbf{x}}.
\end{equation}
It is clear that this equation should simplify to
\begin{equation}
p(\boldsymbol{\theta}, \gamma | \mathbf{x}, \phi, \sigma) = p(\boldsymbol{\theta}) p(\gamma).
\end{equation}
Thus, one could argue that any links that are not present in the graphical model of Figure \ref{subfig:AfterMargin} but found by \citet{Calderhead} and reused in \citet{Dondelinger} and \citet{VGM} were created by improper normalization of the density $p(\dot{\boldsymbol{x}} | \mathbf{x}, \boldsymbol{\theta}, \gamma, \phi, \sigma)$.

\subsection{Adapting The Original Graphical Model}
\label{sec:ReformulatingTraditionalPoE}
Despite these technical difficulties arising from the PoE heuristic, the approaches provide good empirical results and have been used in practice, e.g., by \citet{babtie2014topological}. In what follows, we derive an alternative model and mathematical justification for Equation \eqref{eq:DondelingerDensity} to provide a theoretical framework explaining the good empirical performance of Gaussian process-based gradient matching approaches, especially from \citet{VGM}, which uses only weak or nonexistent priors. 

\begin{figure}[h]
    \centering
    \begin{tikzpicture}[
    node distance=0.5 cm and 0.5 cm,
    mynode/.style={draw,circle,minimum size=0.9cm}  % needs to be that big for F_1 to fit
    ]
    \node[mynode](sigma) {$\sigma$};
    \node[mynode, right=of sigma] (phi){$\boldsymbol{\phi}$};
    \node[mynode, below=of sigma] (y){$\mathbf{y}$};
    \node[mynode, right=of y] (x){$\mathbf{x}$};
    \node[mynode, right=of x] (xDot){$\mathbf{\dot{x}}$};
    \node[mynode,below=of y] (theta){$\theta$};
    \node[mynode, right=of theta] (F1){$\mathbf{F_1}$};
    \node[mynode, right= of F1] (F2){$\mathbf{F_2}$};
    \node[mynode, right= of F2] (gamma){$\gamma$};

    \path
    (phi) edge[-latex] (x)
    (x) edge[-latex] (y)
    (sigma) edge[-latex] (y)
    (phi) edge[-latex] (xDot)
    (x) edge[-latex] (xDot)
    (gamma) edge[-latex] (F2)
    (xDot) edge[-latex] (F2)
    ;
    
    \draw[-latex,
        preaction={
            draw, black!20!white, -,
            double=black!20!white,
            double distance=6\pgflinewidth}](theta) to (F1);
    \draw[-latex,
    preaction={
        draw, black!20!white, -,
        double=black!20!white,
        double distance=6\pgflinewidth}](x) to (F1);
    \draw[-,
    preaction={
        draw, black!20!white, -,
        double=black!20!white,
        double distance=6\pgflinewidth}](F1) to (F2);
    \end{tikzpicture}
    \caption{Alternative probabilistic model without PoE heuristic. Gray shaded connections are used to indicate a deterministic relationship.}
    \label{fig:AlternativeBNet}
\end{figure}
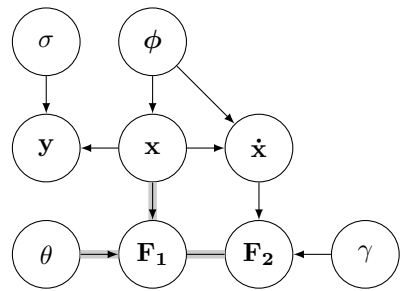
 
The graphical model shown in Figure \ref{fig:AlternativeBNet} offers an alternative approach to the product of experts heuristic. The top two layers are equivalent to a GP prior on the states, the induced GP on the derivatives and the observation model, as shown in Figure \ref{subfig:GPModel}.

The interesting part of the new graphical model is the bottom layer. Instead of adding a second graphical model like in Figure \ref{subfig:ODEModel} to account for the ODE response, two additional random variables are introduced.

$\mathbf{F_1}$ is the deterministic output of the ODEs, assuming the values of $\mathbf{x}$ and $\boldsymbol{\theta}$ are given, i.e. $\mathbf{F_1} = \mathbf{f}(\mathbf{x}, \boldsymbol{\theta})$. The deterministic nature of this equation is represented as a Dirac delta function:
\begin{equation}
p(\mathbf{F_1} | \mathbf{x}, \boldsymbol{\theta}) = \delta(\mathbf{F_1} - \mathbf{f}(\boldsymbol{\theta}, \mathbf{x}))
\end{equation}
If the GP model were able to capture the true states and true derivatives perfectly, this new random variable should be equivalent to the derivatives of the GP, i.e., $\mathbf{F_1}=\mathbf{\dot{x}}$. However, to compensate for a potential model mismatch and slight errors of both GP states and GP derivatives, this condition is relaxed to
\begin{align}
\label{eq:NewODEDensityDef}
&\mathbf{F_1} = \mathbf{\dot{x}} + \boldsymbol{\epsilon} =: \mathbf{F_2}, &\boldsymbol{\epsilon}\sim \mathcal{N}(\mathbf{0}, \gamma \mathbf{I})
\end{align}
In the graphical model, this intuitive argument is encoded via the random variable $\mathbf{F_2}$. Given the $\mathbf{\dot{x}}$ provided by the GP model, Gaussian noise with standard deviation $\gamma$ is added to create $\mathbf{F_2}$, whose probability density can thus be described as
\begin{equation}
\label{eq:definitionOfF2Density}
p(\mathbf{F_2} | \mathbf{\dot{x}}, \gamma) = \mathcal{N}(\mathbf{F_2} | \mathbf{\dot{x}}, \gamma \mathbf{I}).
\end{equation}
The equality constraint given by equation \eqref{eq:NewODEDensityDef} is represented in the graphical model by the undirected edge between $\mathbf{F_1}$ and $\mathbf{F_2}$. When doing inference, this undirected edge is incorporated in the joint density via a Dirac delta function ${\delta(\mathbf{F_2} - \mathbf{F_1})}$. Thus, the joint density of the graphical model represented in Figure \ref{fig:AlternativeBNet} can be written as
%\begin{align}
%p(\mathbf{x}, \mathbf{\dot{x}}, &\mathbf{y}, \mathbf{F_1}, \mathbf{F_2}, \boldsymbol{\theta} | \boldsymbol{\phi}, \sigma, \gamma) = \nonumber \\
%&p(\boldsymbol{\theta}) %parameter priors
%p(\mathbf{x} | \boldsymbol{\phi}) % GP prior
%p(\mathbf{\dot{x}} | \mathbf{x}, \boldsymbol{\phi})% derivative prior
%p(\mathbf{y}|\mathbf{x}, \sigma) \nonumber \\% observation
%\times \hspace{.5em} &p(\mathbf{F_1} | \boldsymbol{\theta}, \mathbf{x}) % F_1 density
%p(\mathbf{F_2} | \mathbf{\dot{x}}, \gamma \mathbf{I})%F_2 density
%\delta(\mathbf{F_1} - \mathbf{F_2}). \label{eq:AlternativeGraphModelEquations}
%\end{align}
\begin{align}
p(\mathbf{x}, \mathbf{\dot{x}}, \mathbf{y}, &\mathbf{F_1}, \mathbf{F_2}, \boldsymbol{\theta} | \boldsymbol{\phi}, \sigma, \gamma) = 
p(\boldsymbol{\theta}) \nonumber \\ %parameter priors 
&\times p(\mathbf{x} | \boldsymbol{\phi}) % GP prior
p(\mathbf{\dot{x}} | \mathbf{x}, \boldsymbol{\phi})% derivative prior
p(\mathbf{y}|\mathbf{x}, \sigma) \nonumber \\% observation
&\times p(\mathbf{F_1} | \boldsymbol{\theta}, \mathbf{x}) % F_1 density
p(\mathbf{F_2} | \mathbf{\dot{x}}, \gamma \mathbf{I})%F_2 density
\delta(\mathbf{F_1} - \mathbf{F_2}). \label{eq:AlternativeGraphModelEquations}
\end{align}
%\begin{align}
%p(\boldsymbol{\theta}) %parameter priors
%p(\mathbf{x} | \boldsymbol{\phi}) % GP prior
%p(\mathbf{\dot{x}} | \mathbf{x}, \boldsymbol{\phi})% derivative prior
%p(\mathbf{y}|\mathbf{x}, \sigma)% observation
%p(\mathbf{F_1} | \boldsymbol{\theta}, \mathbf{x}) % F_1 density
%p(\mathbf{F_2} | \mathbf{\dot{x}}, \gamma \mathbf{I})%F_2 density
%\delta(\mathbf{F_1} - \mathbf{F_2}). \label{eq:AlternativeGraphModelEquations}
%\end{align}

\subsection{Inference In The New Model}
Given all the definitions in the previous section, inference can now be directly performed without the need for additional heuristics. The result is a theoretically sound justification of the main result of \citet{Calderhead} and \citet{Dondelinger}:

\begin{theorem}
    \label{th:InferenceInNewModel}
    Given the modeling assumptions summarized in the graphical model in Figure \ref{fig:AlternativeBNet},
    \begin{align}
    p(\mathbf{x}, \boldsymbol{\theta}|\mathbf{y}, \boldsymbol{\phi}, \gamma, \sigma) &\propto
    p(\boldsymbol{\theta}) \nonumber \\
    &\times \mathcal{N}(\mathbf{x} | \mathbf{0}, \boldsymbol{C_\phi}) \nonumber \\
    &\times \mathcal{N}(\mathbf{y} | \mathbf{x}, \sigma^2 \mathbf{I}) \nonumber \\
    &\times \mathcal{N}(\mathbf{f}(\mathbf{x}, \boldsymbol{\theta}) | \mathbf{D} \mathbf{x}, \mathbf{A} + \gamma \mathbf{I}).     \label{eq:jointDensityProof}
    \end{align}
\end{theorem}

The proof can be found in the supplementary material, section \ref{sec:proof1}.

\section{FAST GAUSSIAN PROCESS GRADIENT MATCHING}
\label{sec:OwnAlgorithm}
\subsection{Hyperparameters}
Using the theoretical framework of the previous section, it is clear that the performance of any algorithm based on equation \ref{eq:jointDensityProof} will heavily rely on the quality of the hyperparameters $\boldsymbol{\phi}$, as $\mathbf{C}_{\boldsymbol{\phi}}$, $\mathbf{D}$ and $\mathbf{A}$ are all depending on $\boldsymbol{\phi}$. In GP regression, it is common to fit the hyperparameters to the observations using a maximum likelihood scheme (see supplementary material, section \eqref{sec:HyperAndKernelSelection}). However, neither $\mathrm{VGM}$ nor $\mathrm{AGM}$ actually do this.

In $\mathrm{AGM}$, the hyperparameters are inferred concurrently to the states $\mathbf{x}$ and the ODE parameters $\boldsymbol{\theta}$ in one big MCMC scheme. Besides the need for clear hyperpriors on the hyperparameters and obvious drawbacks regarding running time, e.g. at each MCMC step $\mathbf{C}_{\boldsymbol{\theta}}$, $\mathbf{D}$ and $\mathbf{A}$ have to be recalculated and potentially inverted, this leads to a rough probability landscape requiring a complicated multi-chain setup \citep{Dondelinger}.

In $\mathrm{VGM}$, this problem was sidestepped by treating the hyperparameters as tuning parameters that had to be set by hand. As these hyperparameters are not learned from data, it is obviously not a fair comparison. In our experiments, we thus used a maximum likelihood scheme to infer the hyperparameters before using the implementation of \citet{VGM}. To indicate this modification, the modified VGM algorithm was called $\mathrm{MVGM}$.

\subsection{FGPGM}
However, $\mathrm{MVGM}$ is still outperforming $\mathrm{AGM}$ significantly as shown in the experiments in section \ref{sec:Experiments}. This suggests that the concurrent optimization of $\boldsymbol{\phi}$, $\mathbf{x}$ and $\boldsymbol{\theta}$ suggested by \citet{Dondelinger} is not helpful for the performance of the system. Based on this insight, we propose the sequential approach shown in Algorithm \ref{algo1}. In a first step, the Gaussian process model is fit to the standardized data by calculating the hyperparameters via equation \eqref{eq:marginalHyperparams}. Then, the states $\mathbf{x}$ and ODE parameters $\boldsymbol{\theta}$ are inferred using a one chain MCMC scheme on the density given by equation \eqref{eq:DondelingerDensity}.

\begin{algorithm}[!h]
\caption{FGPGM}
\label{algo1}
\begin{algorithmic}[1]
\State{\textbf{Input:}\quad  $\mathbf{y}, \mathbf{f}(\mathbf{x}, \boldsymbol{\theta}), \gamma, N_{MCMC}, N_{burnin}, \mathbf{t}, \sigma_{s}, \sigma_{p}$}
\State \emph{Step 1: Fit GP model to data}
 \ForAll{$ k \in K$}
\State $\boldsymbol{\mu}_{y, k} \leftarrow \textrm{mean}(\mathbf{y}_k)$
\State $\sigma_{y, k} \leftarrow \textrm{std}(\mathbf{y}_k)$
\State $\mathbf{\tilde{y}} \leftarrow (\mathbf{y_k} - \boldsymbol{\mu}_k) / \sigma_{y, k}$
\State $\begin{aligned}
 &\textrm{Find } \boldsymbol{\phi}_k \textrm{ and } \sigma_k \textrm{ by maximizing }\\[-4pt]
 &p(\mathbf{\tilde{y}} | \mathbf{t}, \boldsymbol{\phi}_k, \sigma_k) \textrm{ given by equation \eqref{eq:marginalHyperparams}}
 \end{aligned}$
\EndFor

\State \emph{Step 2: Infer $\mathbf{x}$ and $\boldsymbol{\theta}$ using MCMC}
\State $\mathcal{S} \leftarrow \emptyset$
\For{$i=1\rightarrow N_{MCMC} + N_{burnin}$}
\For{\textbf{each} state and parameter}
    \State $\mathcal{T} \leftarrow \emptyset$
    \State
    $\begin{aligned}
    &\textrm{Propose a new state or parameter value}\\[-4pt]
    &\textrm{adding a zero mean Gaussian increment}\\[-4pt]
    &\textrm{with standard deviation }\sigma_{s}\textrm{ or }\sigma_{p}.
    \end{aligned}$
    \State $\begin{aligned}
    &\textrm{Accept proposed value based on the}\\[-4pt]
    &\textrm{density given by equation \eqref{eq:DondelingerDensity}}
    \end{aligned}$
    \State Add current value to $\mathcal{T}$
\EndFor
\State Add the mean of $\mathcal{T}$ to $\mathcal{S}$
\EndFor
\State Discard the first $N_{burnin}$ samples of $\mathcal{S}$
\State Return the mean of $\mathcal{S}$
\State{\textbf{Return:} $\mathbf{x}, \boldsymbol{\theta}$}
\end{algorithmic}
\end{algorithm}

\section{EXPERIMENTS}
\label{sec:Experiments}
For all experiments involving $\mathrm{AGM}$, the R toolbox \mbox{deGradInfer} \citep{codeAGM} published alongside \citep{MacdonaldThesis} was used. The toolbox was provided with code to run two experiments, namely Lotka Volterra and Protein Transduction. Both of these systems are used in this paper, as they are the two standard benchmark systems used in all previous publications. It should be noted however that the applicability of $\mathrm{FGPGM}$ is not restricted to these systems. Unlike $\mathrm{AGM}$, $\mathrm{FGPGM}$ refrains from using hard to motivate hyperpriors and our publicly available implementation can easily be adapted to new settings.

For details about implementation and additional plots, refer to the supplementary material in section \ref{sec:AppendixExperiments}. It should be noted however that all algorithms were provided with one hyperparameter. While the toolbox of $\mathrm{AGM}$ had to be provided with the true standard deviation of the observation noise, $\mathrm{MVGM}$ and $\mathrm{AGM}$ were provided with $\gamma$. The $\gamma$ was determined by testing eight different values logarithmically spaced between 1 and $10^{-4}$ and comparing the results based on observation fit. For more details regarding the experimental setup, we refer to the appendix of \citet{wenk2019ODIN}.

%\subsection{Setup}
\subsection{Lotka Volterra}
\label{sec:LotkaVolterra}

The first system in question is the Lotka Volterra system originally proposed in \citet{LotkaVolterra}. It describes a two dimensional system whose dynamics are given by
\begin{align*}
\dot{x}_1(t) &=  \hspace{0.8em}\theta_1 x_1(t) - \theta_2 x_1(t) x_2(t)\\\dot{x}_2(t) &= -\theta_3 x_2(t) + \theta_4 x_1(t) x_2(t)
\end{align*}

We reproduce the experimental setup of \citet{VGM}, i.e., the system was observed in the time interval $[0, 2]$ at 20 evenly spaced observation times. The true states were initialized with $[5, 3]$ and Gaussian noise with standard deviation 0.1 (low noise) and standard deviation 0.5 (high noise) was added to create the observations.

\begin{figure*}[tbh!]
    \centering
    \begin{subfigure}[t]{.23\textwidth}
        \centering
        \includegraphics[width=\textwidth]{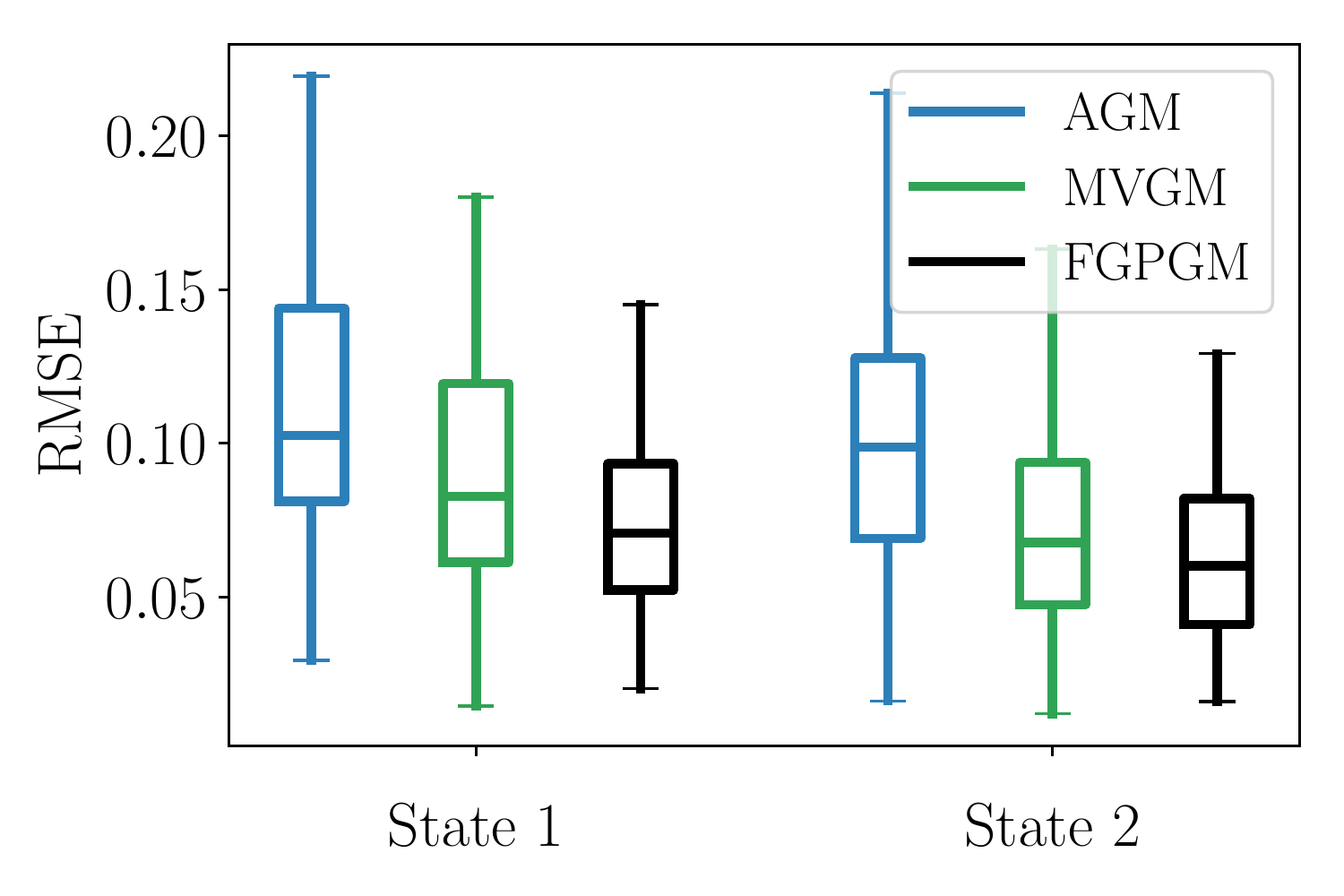}
        \caption{LV low noise}
    \end{subfigure}
    \begin{subfigure}[t]{.23\textwidth}
        \centering
        \includegraphics[width=\textwidth]{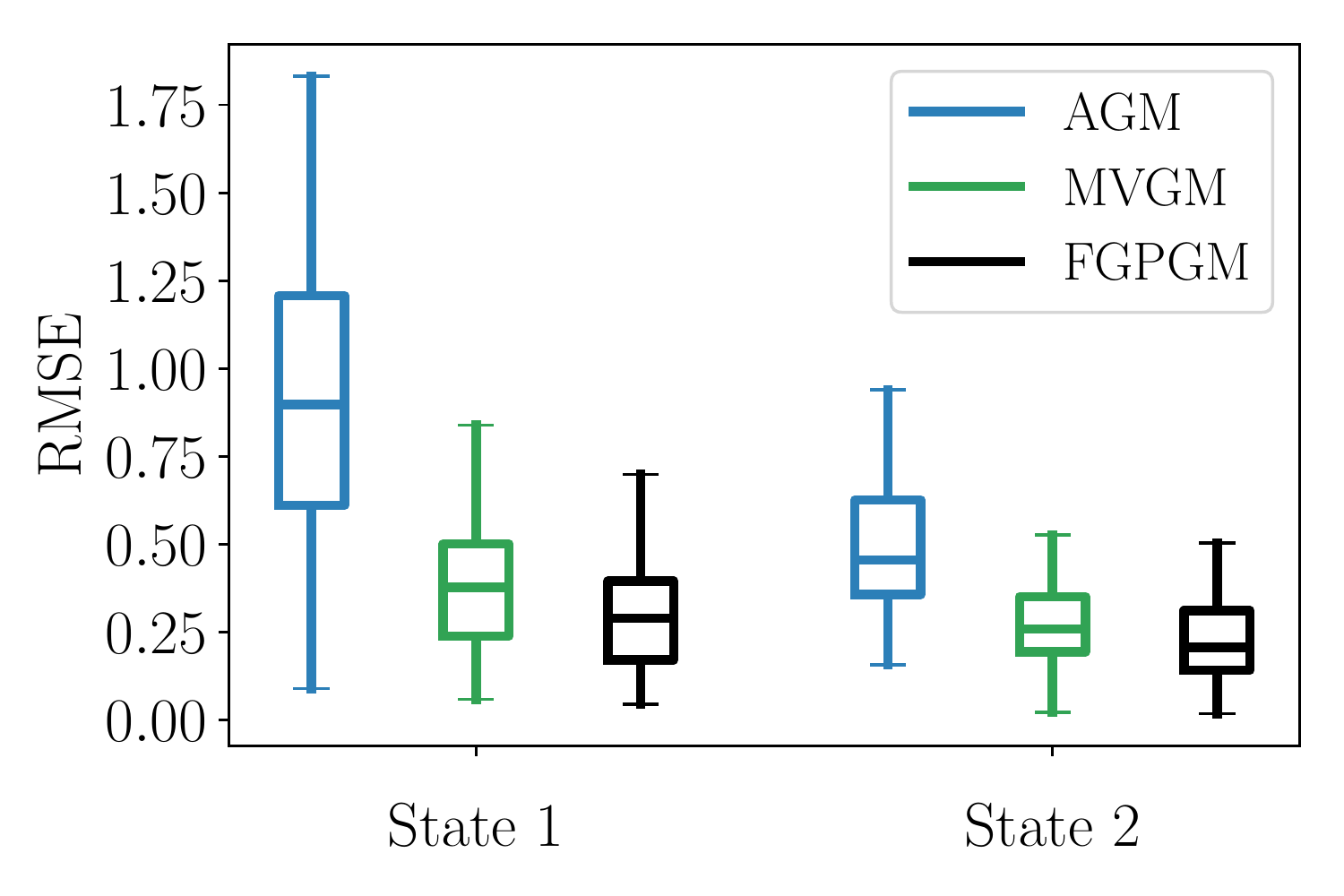}
        \caption{LV high noise}
    \end{subfigure}
    \begin{subfigure}[t]{.23\textwidth}
        \centering
        \includegraphics[width=\textwidth]{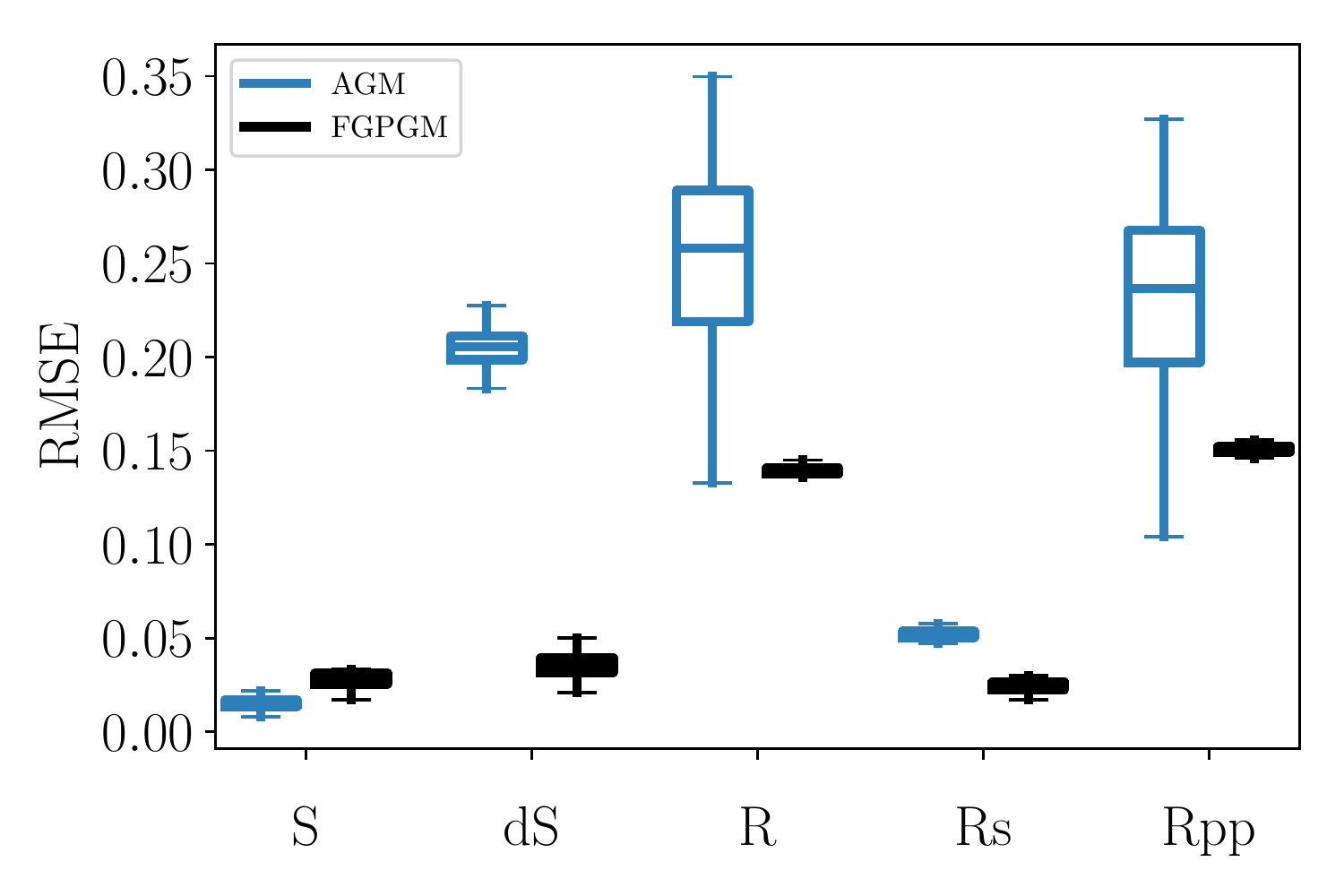}
        \caption{PT low noise}
        \label{fig:PTRMSELow}
    \end{subfigure}
    \begin{subfigure}[t]{.23\textwidth}
        \centering
        \includegraphics[width=\textwidth]{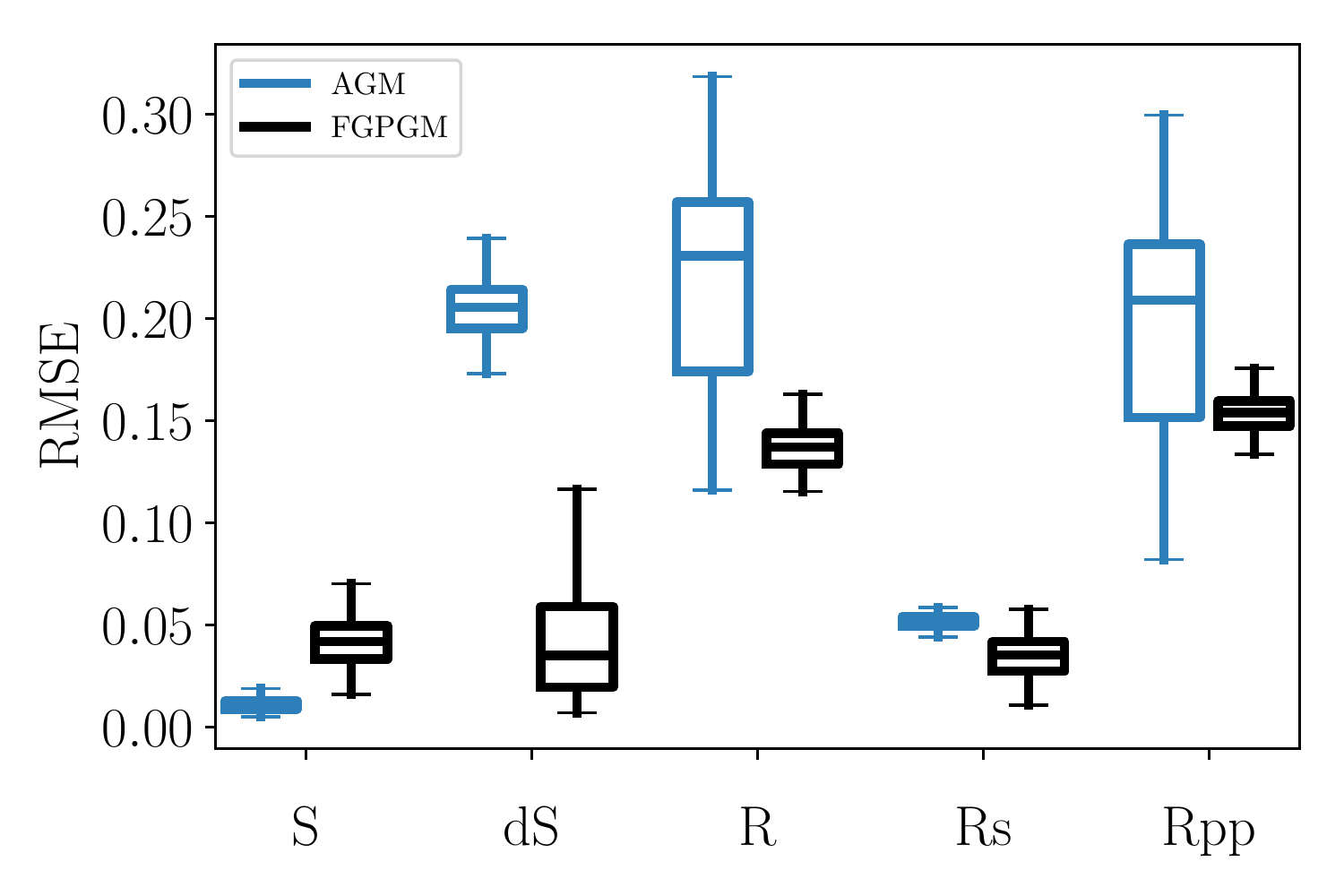}
        \caption{PT high noise}
        \label{fig:PTRMSEHigh}
    \end{subfigure}
    \caption{RMSE of the trajectories obtained by numerical integration compared to the ground truth. Boxplot with median (line), 50\% (box) and 75\% (whisker) quantiles over 100 independent noise realizations. Lotka Volterra is shown on the left, Protein Transduction on the right.}
    \label{fig:RMSE}
    \vspace{1.5em}
\end{figure*}

 \subsection{Protein Transduction}
 \label{sec:ProteinTransduction}
 The second system to be analyzed is called Protein Transduction and was originally proposed in \citet{ProteinTransductionPaper}. It is known to be notoriously difficult to fit with unidentifiable parameters \citep{Dondelinger}. The dynamics are given by:
 \begin{alignat}{2}
 &\dot{S} &&= -\theta_1 S - \theta_2 S R + \theta_3 R_S \nonumber \\
 &\dot{dS} &&= \theta_1 S \nonumber \\
 &\dot{R} &&= -\theta_2 S R + \theta_3 R_S + \theta_5 \frac{R_{pp}}{\theta_6 + R_{pp}} \nonumber \\
 &\dot{R_S} &&= \theta_2 S R - \theta_3 R_S - \theta_4 R_S \nonumber \\
 &\dot{R_{pp}} &&= \theta_4 R_S - \theta_5 \frac{R_{pp}}{\theta_6 + R_{pp}} 
 \end{alignat}
 It should be noted that these dynamics contain nonlinear terms violating the functional form assumption given in equation \eqref{eq:VGMFunctionalForm} of $\mathrm{VGM}$ inherited by $\mathrm{MVGM}$. Nevertheless, both $\mathrm{FGPGM}$ and $\mathrm{AGM}$ can still be applied. For $\mathrm{FGPGM}$, $\gamma$ was set to $10^{-4}$, while $\mathrm{AGM}$ was provided with the true observation noise standard deviations. The experimental setup of \citet{Dondelinger} was copied, i.e. the system was observed in the time interval $[0, 100]$ at the discrete observation times
 $\mathbf{t} = [0, 1, 2, 4, 5, 7, 10, 15, 20, 30, 40, 50, 60, 80, 100]$, the states were initialized with $\mathbf{x}(0) = [1,0,1,0,0]$
 and the parameters were set to
 $\boldsymbol{\theta} = [0.07,0.6,0.05,0.3,0.017,0.3]$.
 As in \citet{Dondelinger}, Gaussian noise was added with standard deviation 0.001 (low noise) and 0.01 (high noise). As in the previous papers, a sigmoid kernel was used to deal with the logarithmically spaced observation times and the typically spiky form of the dynamics.

\begin{figure*}
    \begin{minipage}{.75\textwidth}
        \centering
        \subcaptionbox{$\mathrm{AGM}$}{\includegraphics[width=.31\textwidth]{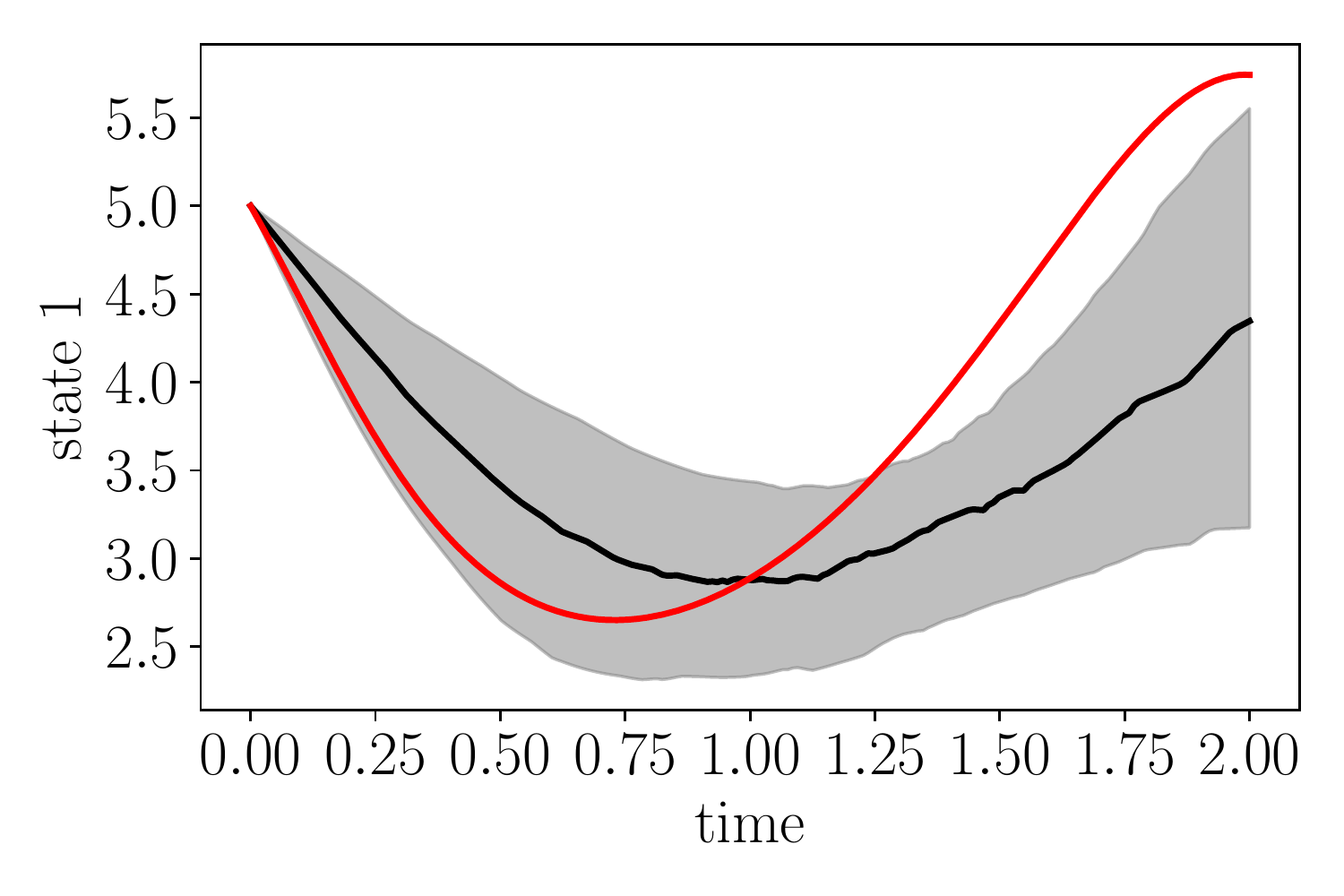}}
        \subcaptionbox{$\mathrm{MVGM}$}{\includegraphics[width=.31\textwidth]{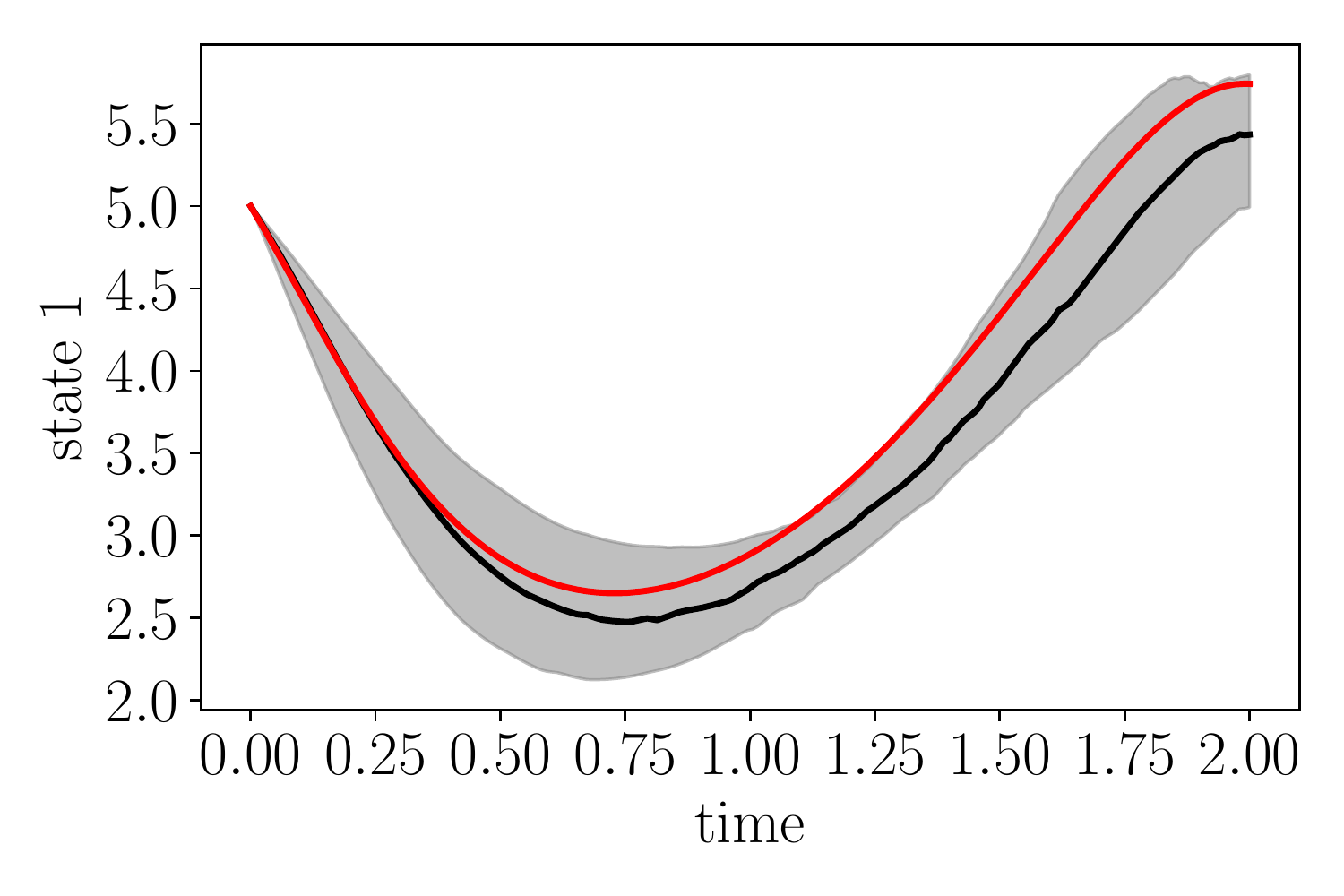}}        \subcaptionbox{$\mathrm{FGPGM}$}{\includegraphics[width=.31\textwidth]{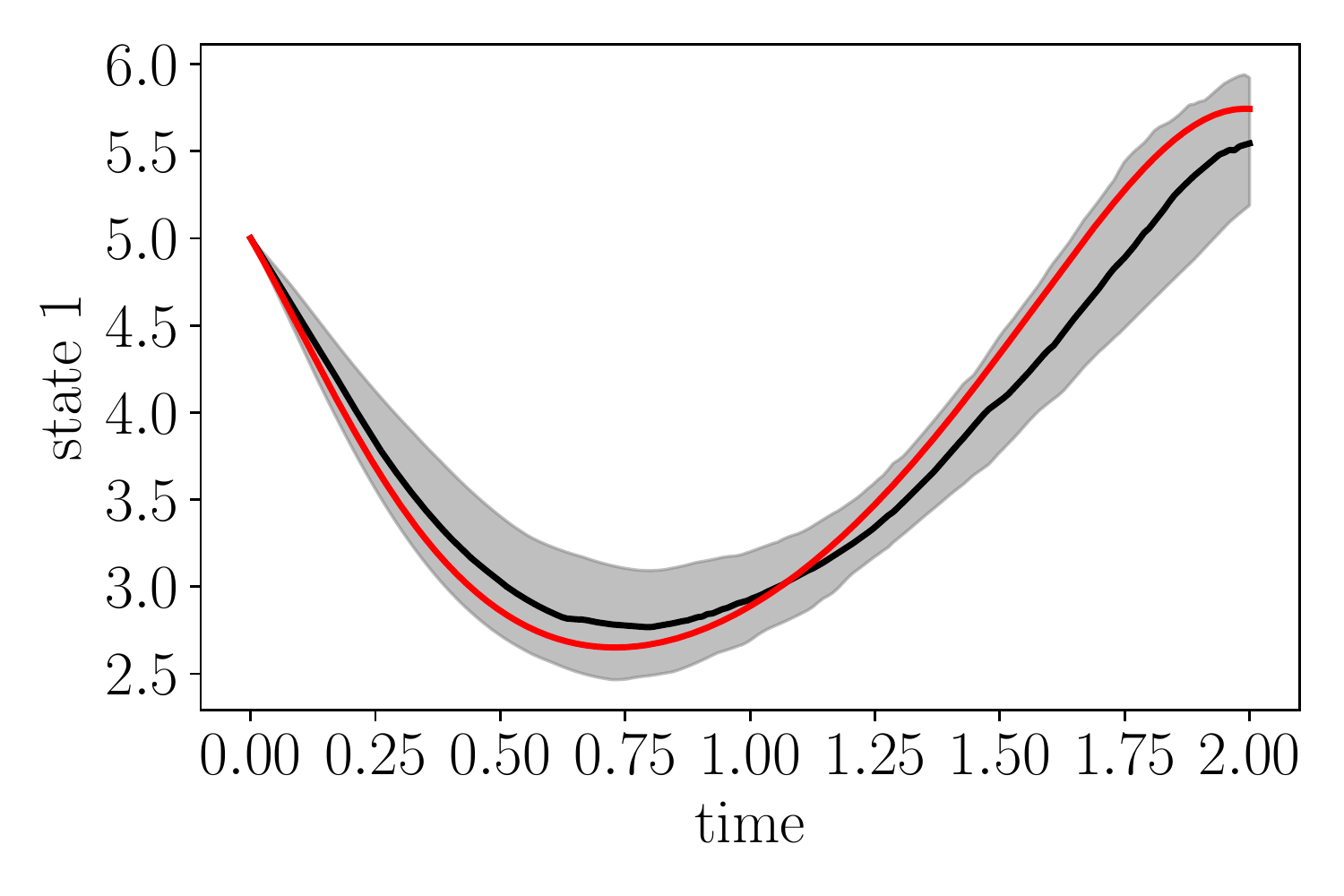}}                \subcaptionbox*{}{\includegraphics[width=.31\textwidth]{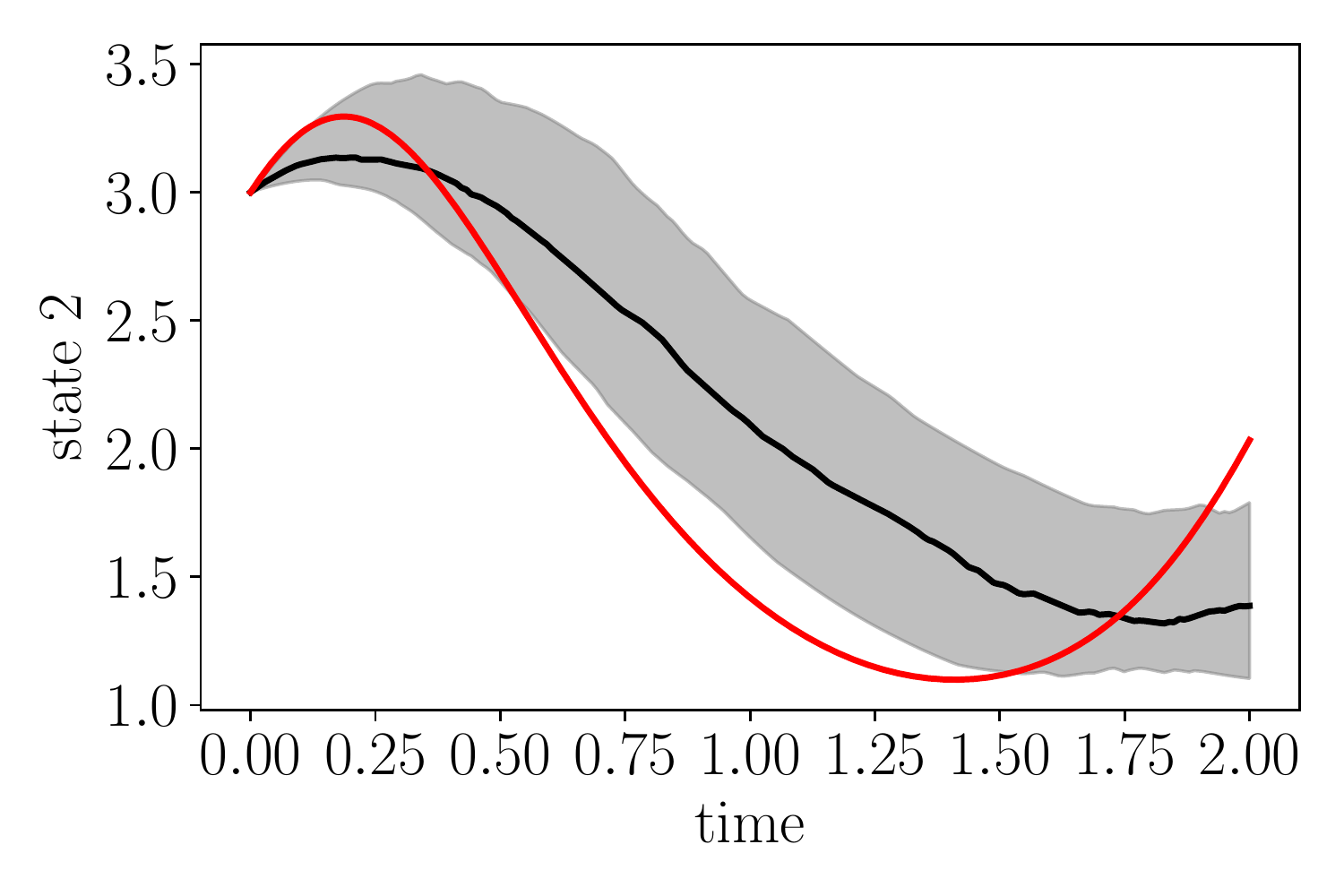}}
        \subcaptionbox*{}{\includegraphics[width=.31\textwidth]{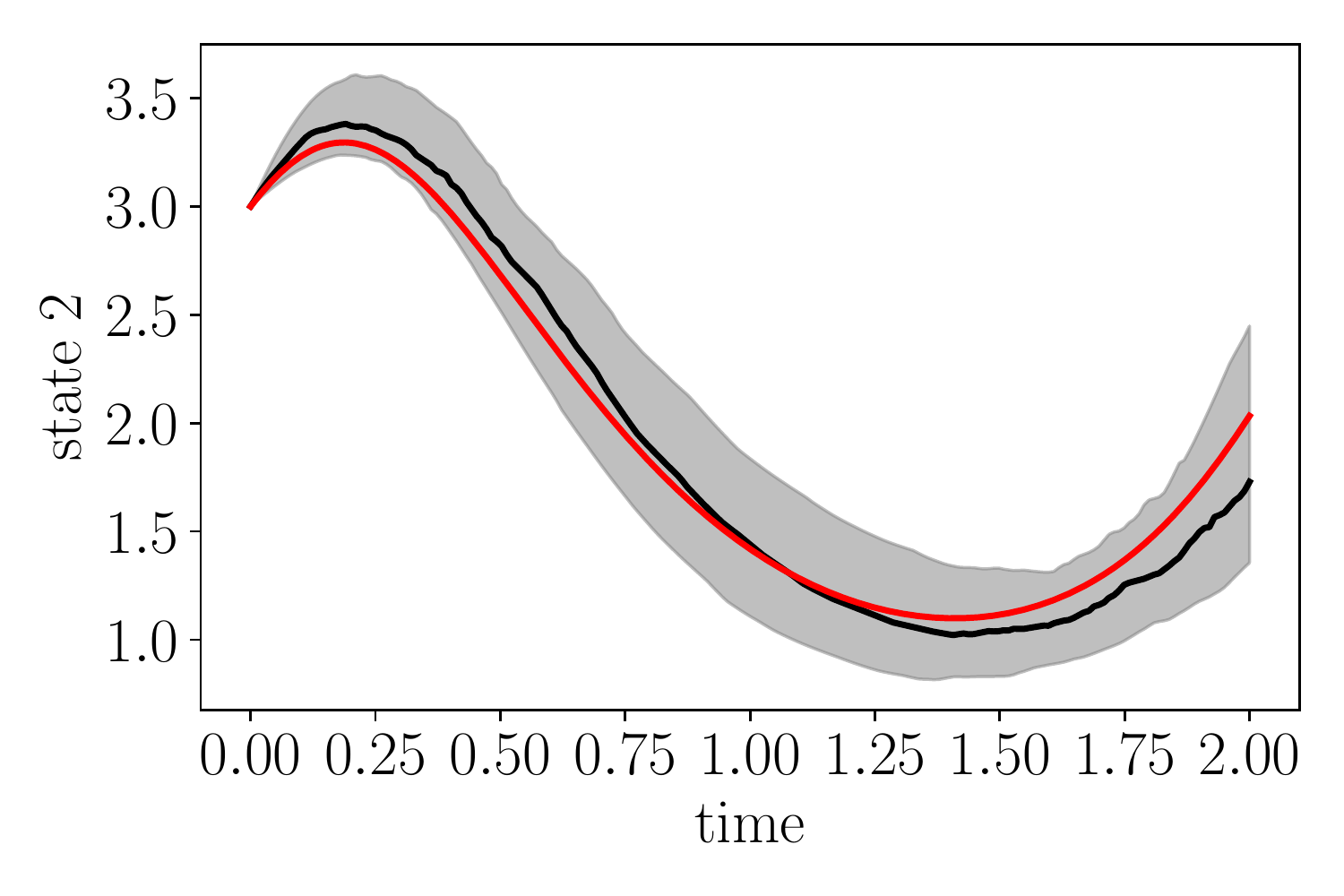}}        \subcaptionbox*{}{\includegraphics[width=.31\textwidth]{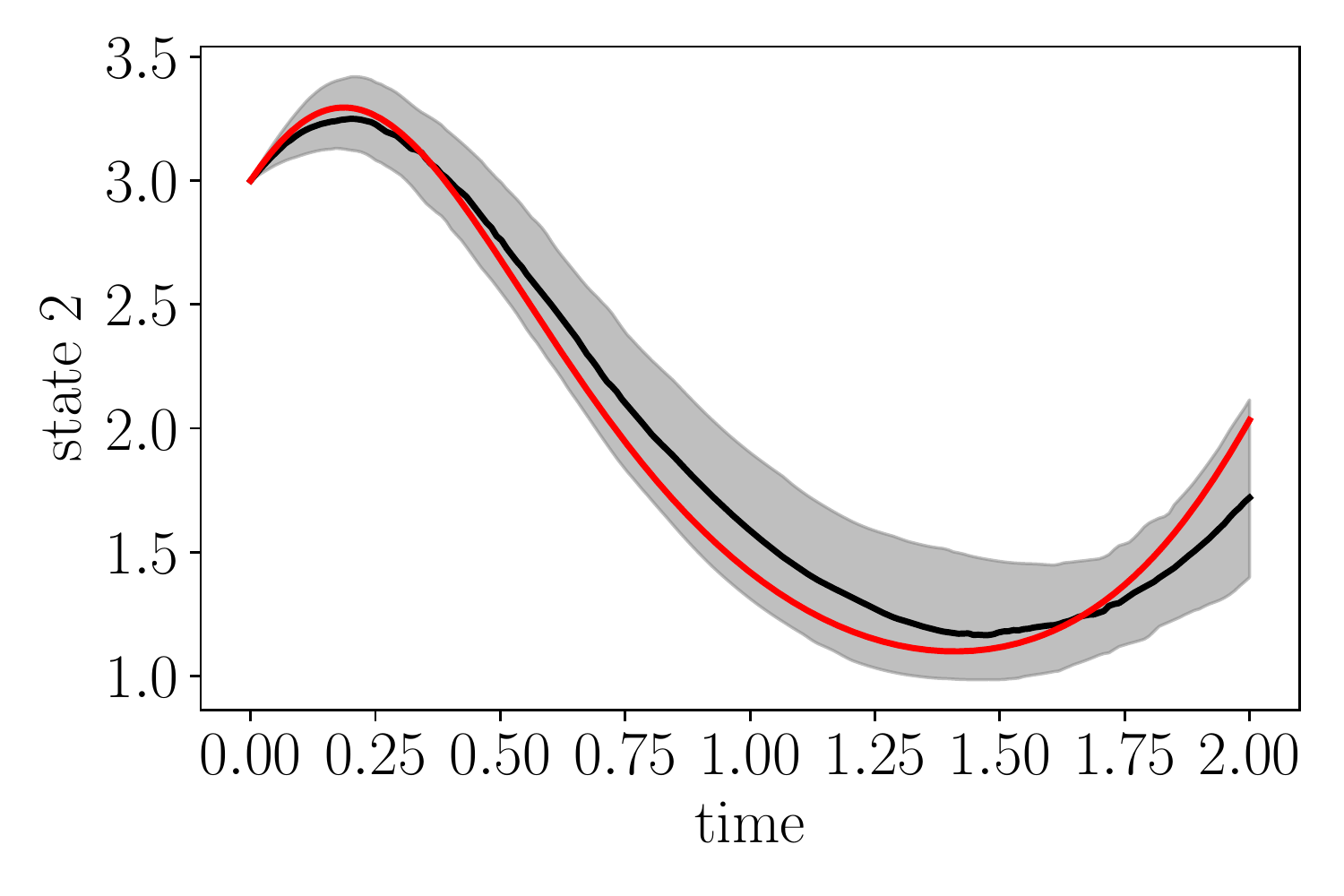}} 
        \vspace{-20pt}      
        \caption{States after numerical integration of the inferred parameters in the high noise case of Lotka Volterra. Ground truth (red), median (black) and 75\% quantiles (gray) over 100 independent noise realizations.}
        \label{fig:StateLVHighNoise}
    \end{minipage}\hfill
    \begin{minipage}{.22\textwidth}
        \centering
%        \resizebox{.22\textwidth}
        \captionof{table}{Median reduction of state RMSE of $\mathrm{FGPGM}$ compared to $\mathrm{AGM}$ and $\mathrm{VGM}$ as baseline.}
        \label{tab:RMSE}
        \begin{tabular}[c]{c|c}
            & $\mathrm{AGM}$ \\
            LV low& 35\%\\
            LV high& 62\%\\
            PT low& 50\% \\
            PT high & 43\% \\
            \\
            & $\mathrm{MVGM}$\\
            LV low & 13\% \\
            LV high & 31\%
        \end{tabular}
    \end{minipage}
\end{figure*}

\subsection{Evaluation}

\begin{figure}
    \centering
    \begin{subfigure}[t]{.22\textwidth}
        \centering
        \includegraphics[width=\textwidth]{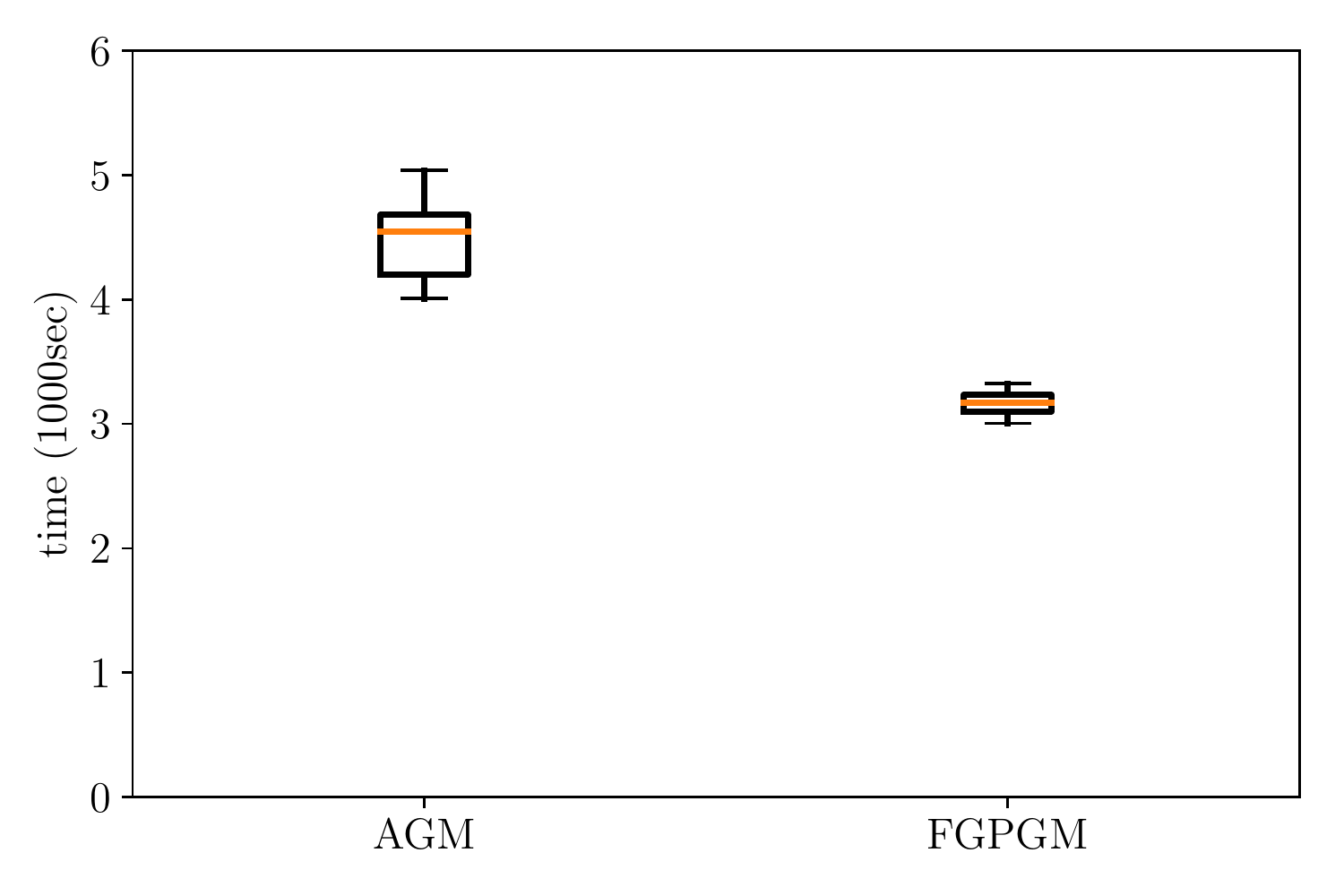}
        \caption{run time LV}
    \end{subfigure}
    \begin{subfigure}[t]{.22\textwidth}
        \centering
        \includegraphics[width=\textwidth]{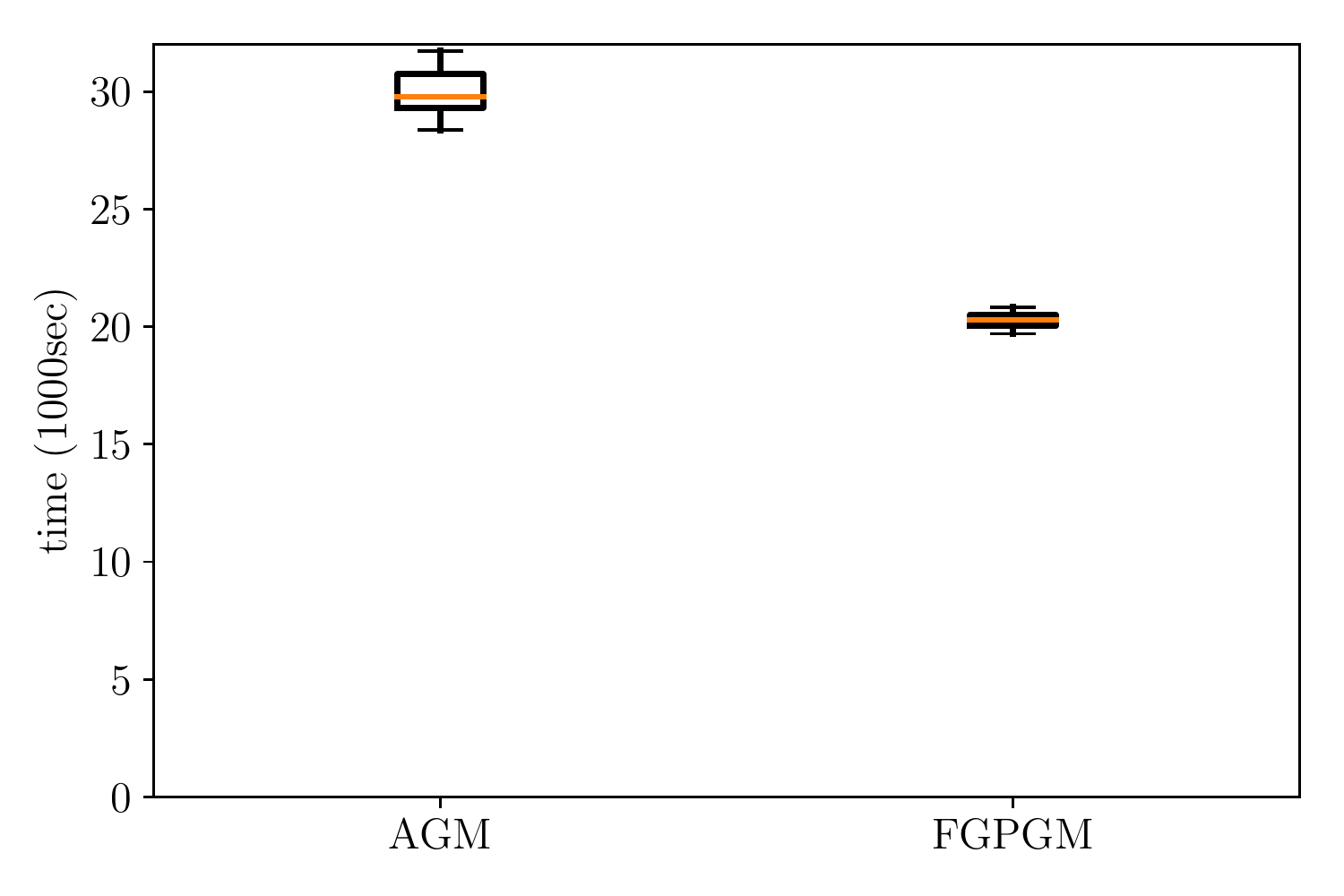}
        \caption{run time PT}
    \end{subfigure}
    \caption{$\mathrm{FGPGM}$ shows a clearly reduced run time compared to $\mathrm{AGM}$, saving roughly 35\% of computation time.}
    \label{fig:runTime}
\end{figure}

As the parameters of the Protein Transduction system are unidentifiable, comparing parameter values is not a good metric to rank approaches. Instead, after running the algorithms, we used a numerical integrator to obtain the trajectories corresponding to the dynamical system whose dynamics is given by the the inferred parameters. Then, the RMSE of these trajectories compared to the ground truth at the observation times were evaluated. For each experimental setting, this procedure was repeated for 100 different noise realizations. The results are shown in Figure \ref{fig:RMSE}. 
In Figure \ref{fig:StateLVHighNoise}, we show median plots for the high noise setting of Lotka Volterra, while the running time between the state of the art $\mathrm{AGM}$ and $\mathrm{FGPGM}$ is shown in Figure \ref{fig:runTime}.

While achieving run time savings of 35\%, $\mathrm{FGPGM}$ shows an increased accuracy, reducing the state RMSE up to 62 \% as shown in Table \ref{tab:RMSE} and shows much lower variability across noise realizations. The effect is especially striking for the nonlinear system Protein Transduction, where the variance of the states $R$ and $R_{pp}$ has been reduced by at least one order of magnitude, while reducing the bias by more than 30\%, see Figures \ref{fig:PTRMSELow} and \ref{fig:PTRMSEHigh}.

\begin{figure*}
	\begin{minipage}{\textwidth}
		\centering
		\subcaptionbox{input sample}{\includegraphics[width=.24\textwidth]{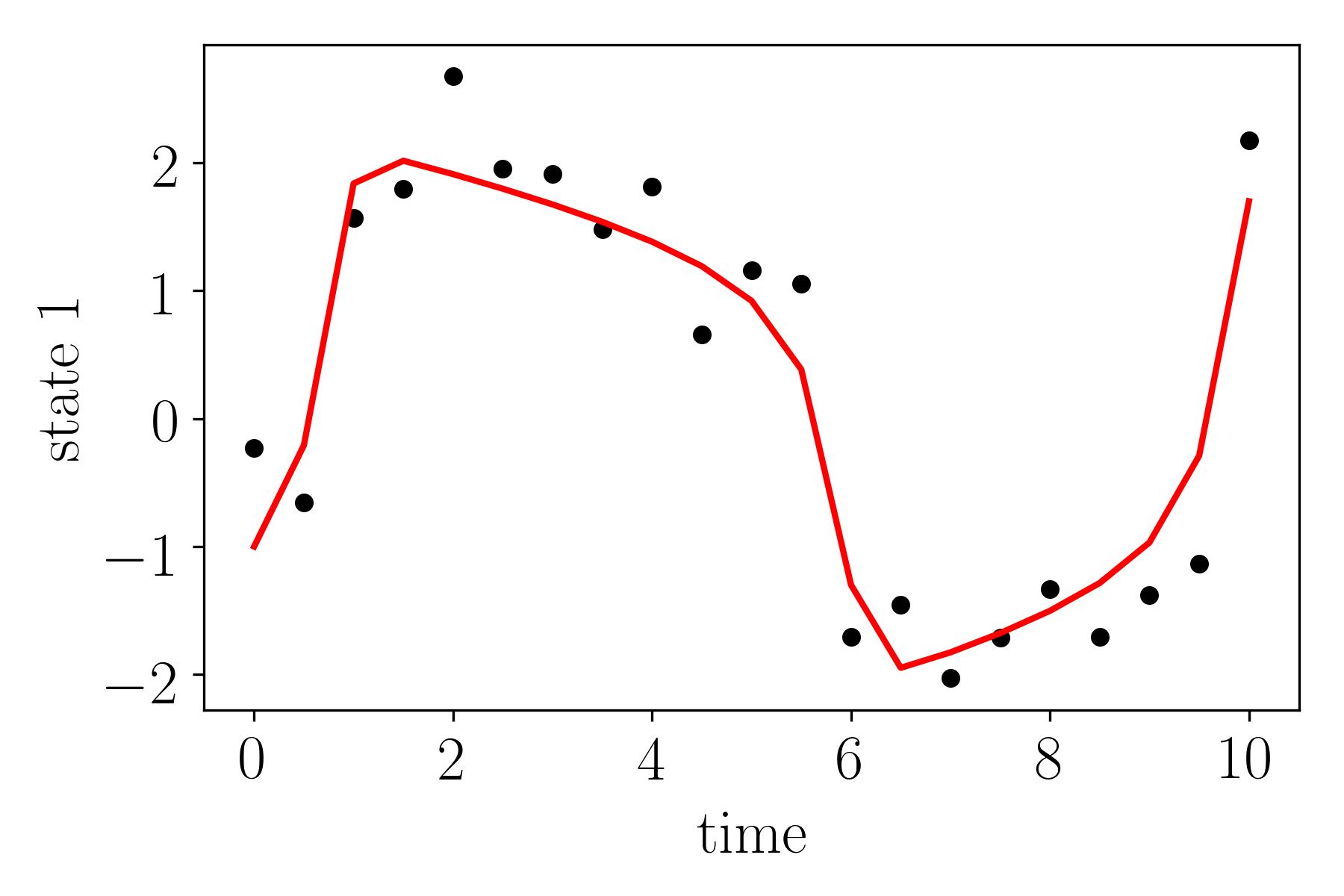}}
		\subcaptionbox{20 observations}{\includegraphics[width=.24\textwidth]{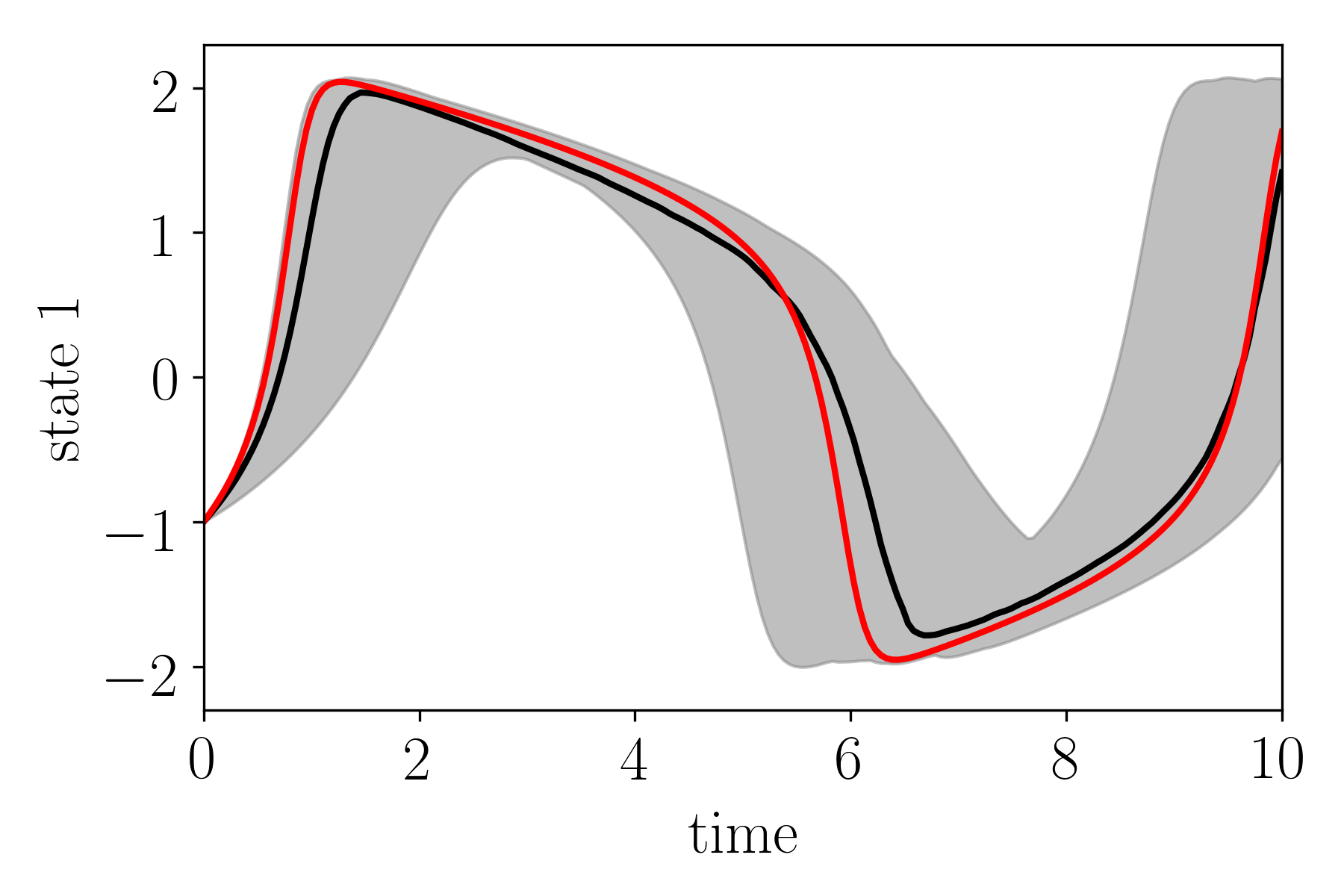}}
		\subcaptionbox{40 observations}{\includegraphics[width=.24\textwidth]{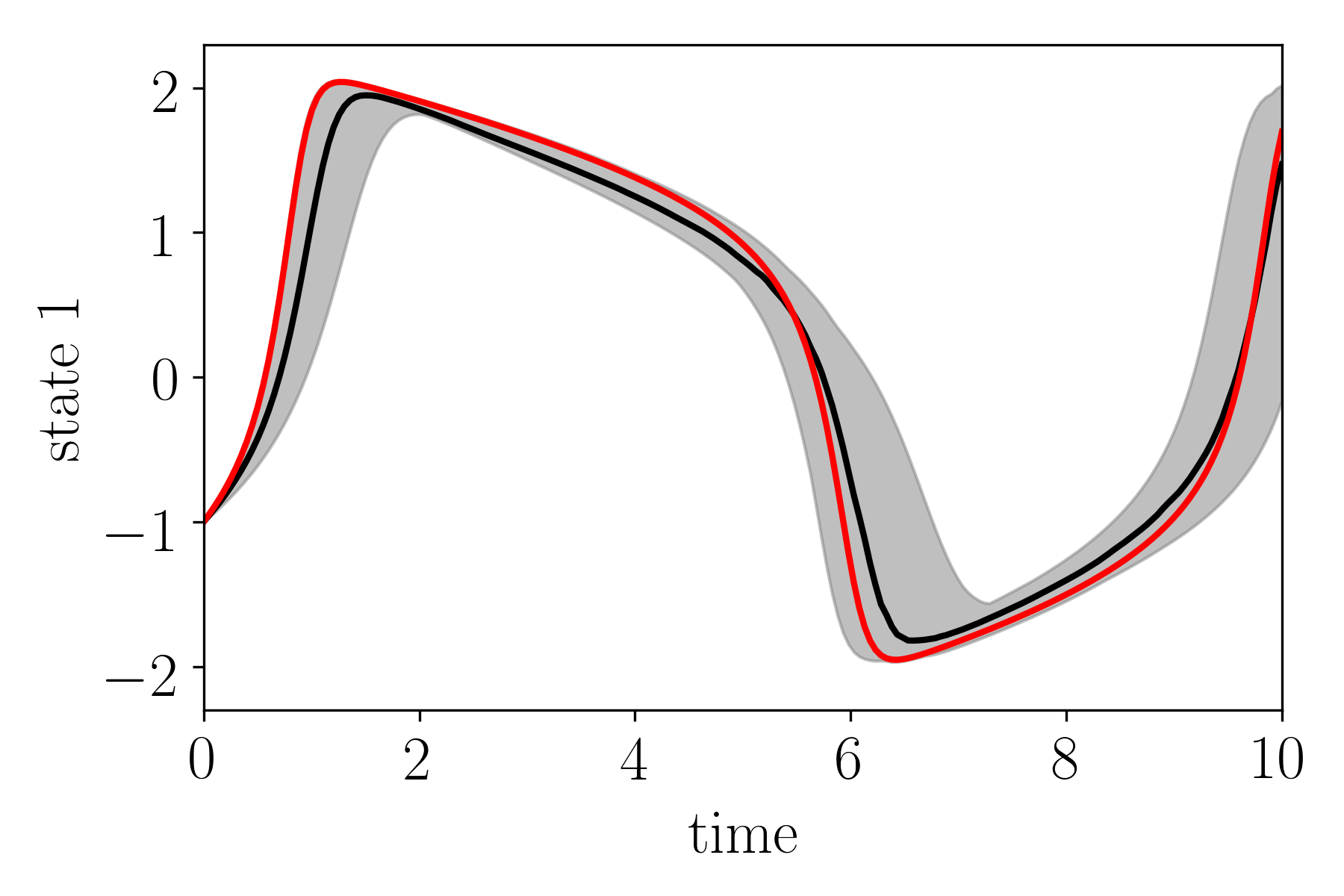}}
		\subcaptionbox{100 observations}{\includegraphics[width=.24\textwidth]{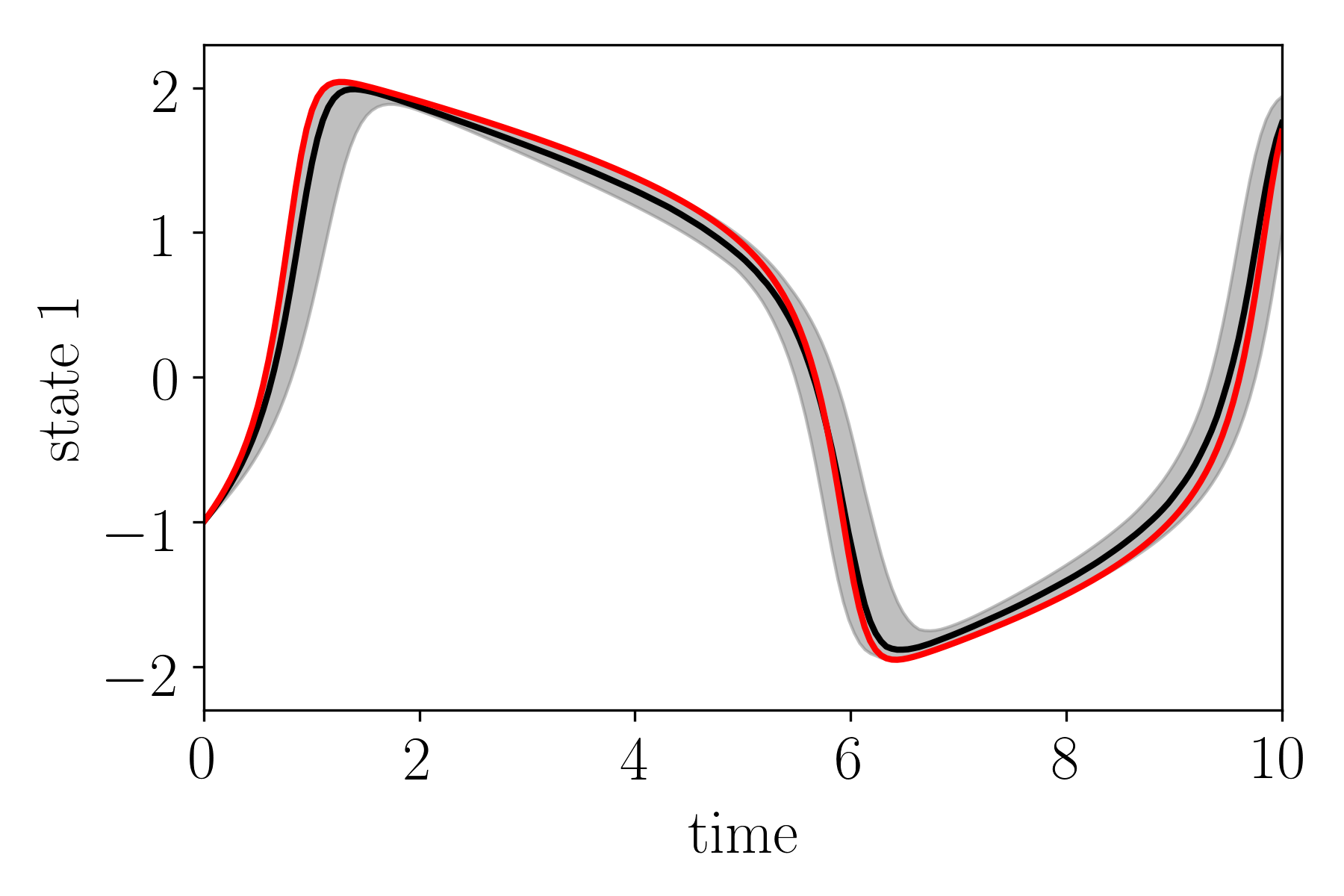}}
		\subcaptionbox*{}{\includegraphics[width=.24\textwidth]{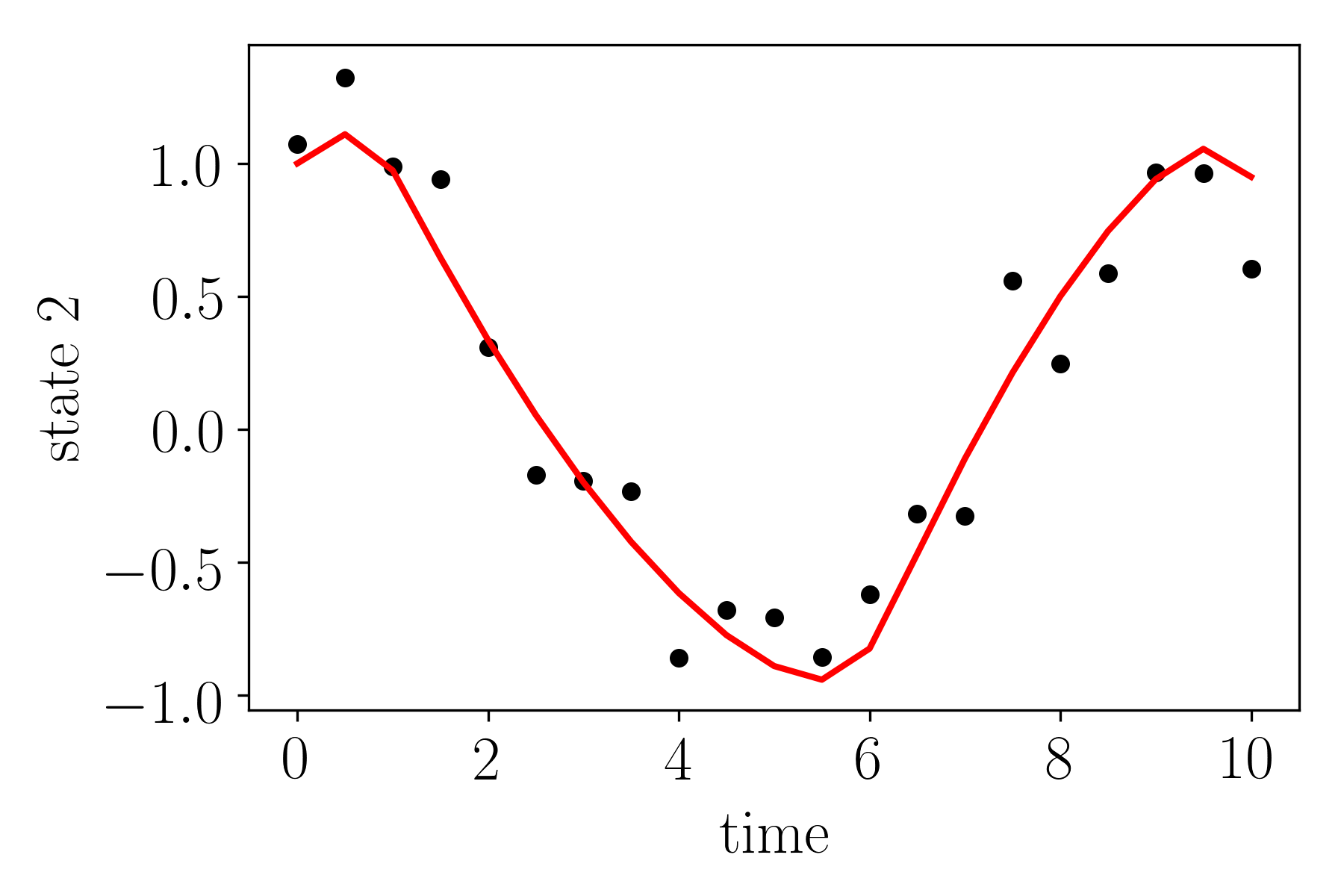}}
		\subcaptionbox*{}{\includegraphics[width=.24\textwidth]{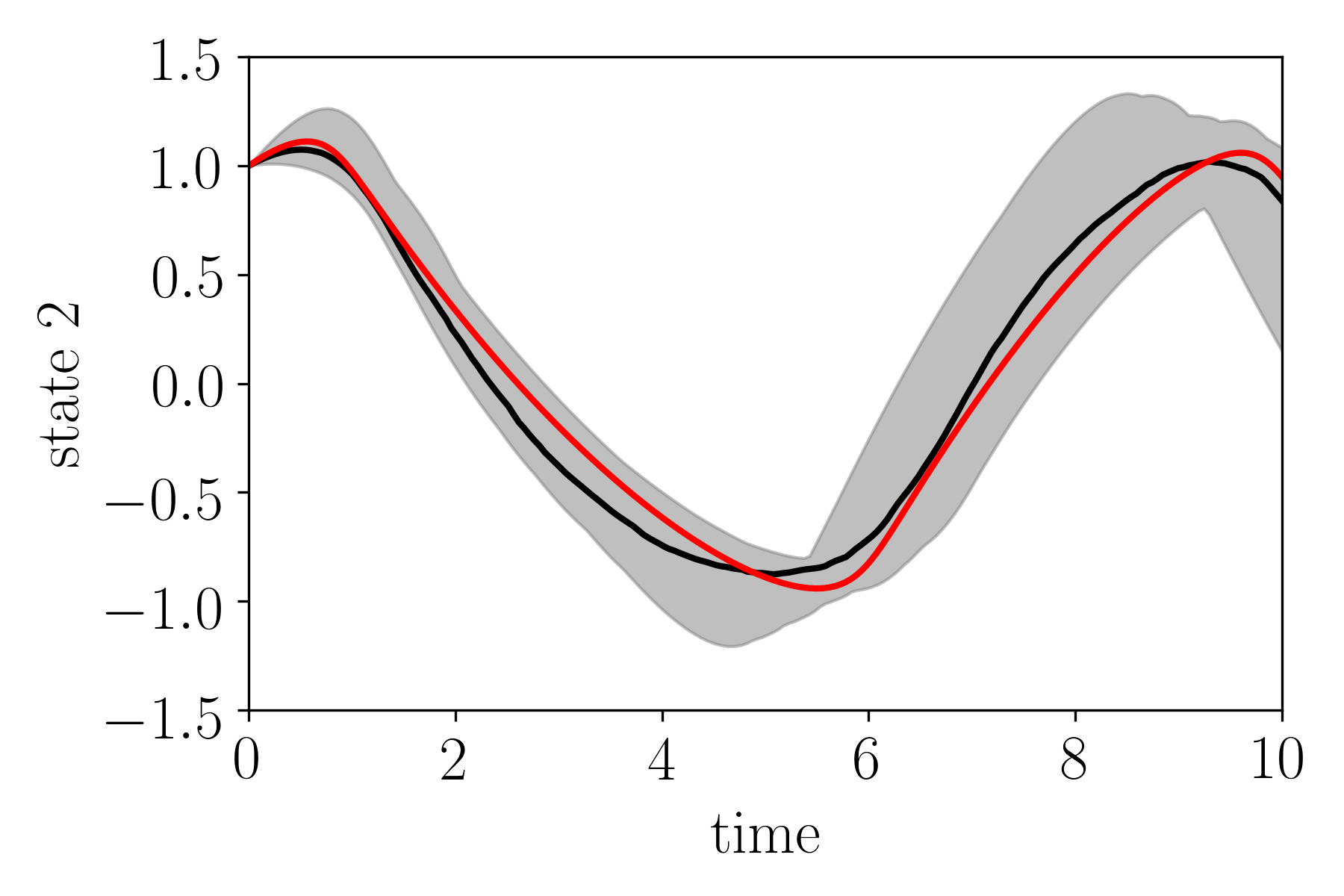}}
		\subcaptionbox*{}{\includegraphics[width=.24\textwidth]{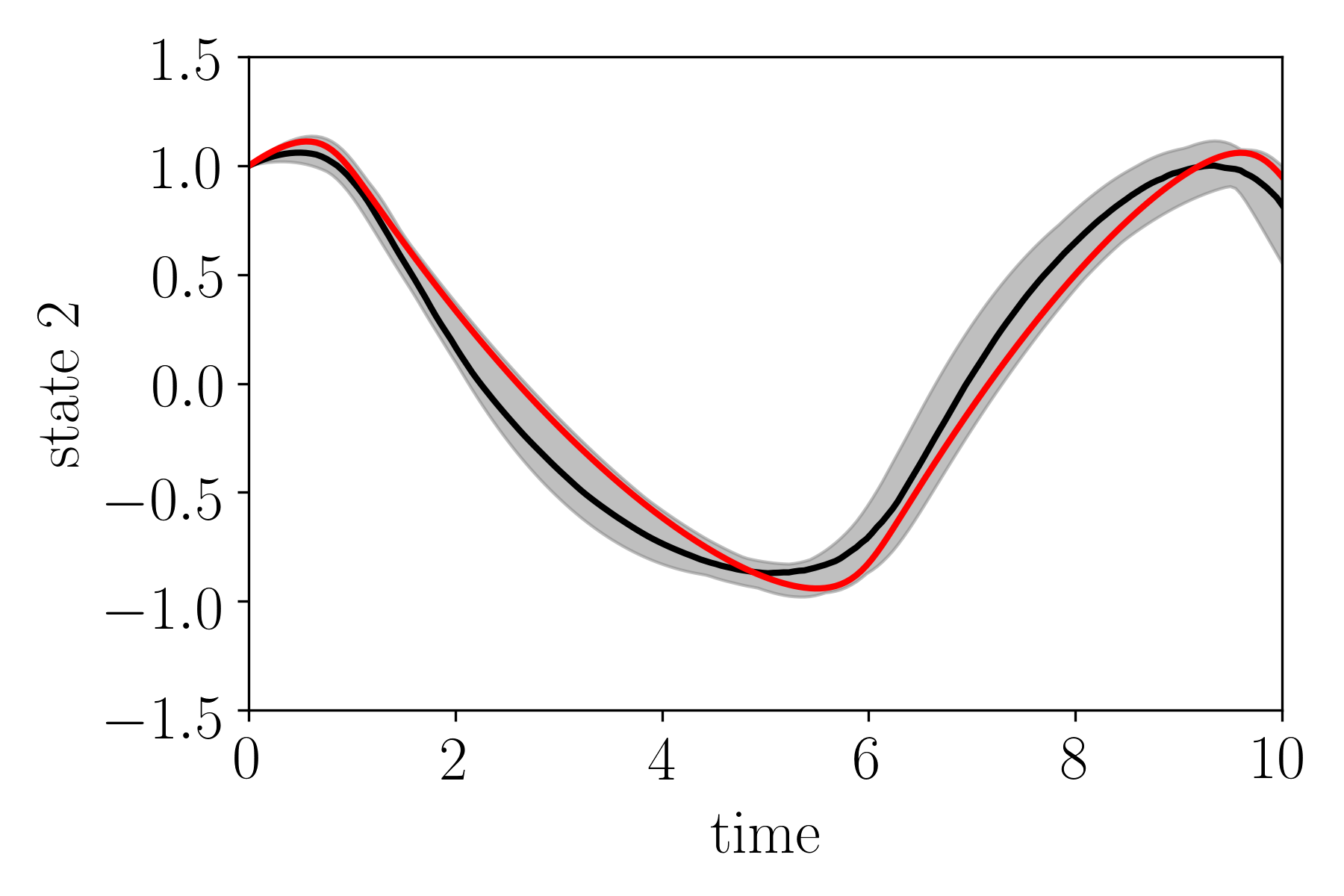}}
		\subcaptionbox*{}{\includegraphics[width=.24\textwidth]{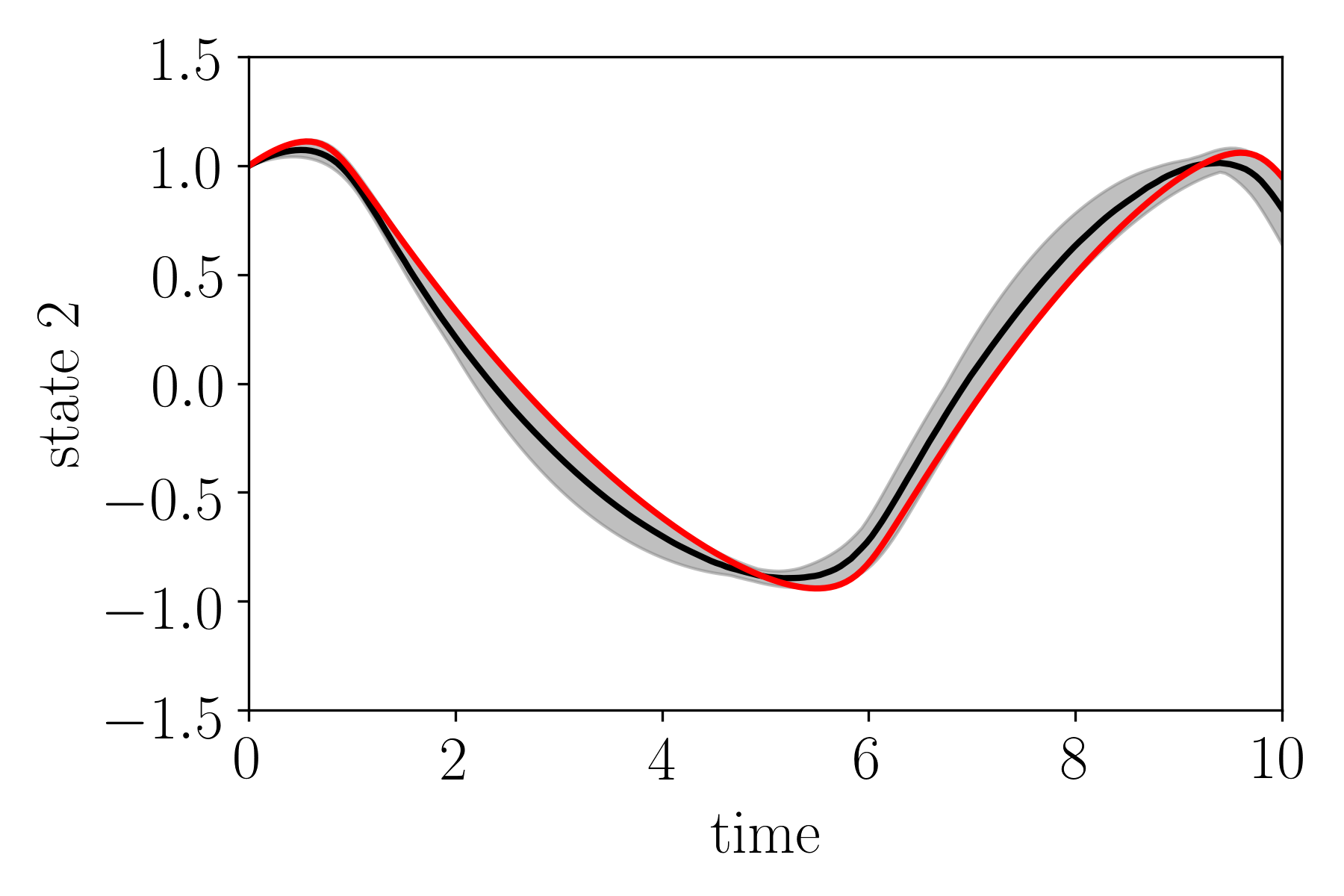}}
		\vspace{-20pt}      
		\caption{One input sample and median plots of the numerically integrated states after parameter inference for the FHN system with SNR 10. Ground truth (red), median (black) and 75\% quantiles (gray) over 100 independent noise realizations. In the most left plots, the black dots represent the noisy observations.}
		\label{fig:FHNHighNoise}
	\end{minipage}
	\vspace{0.2em}
\end{figure*}

\begin{figure*}
	\begin{minipage}{\textwidth}
		\centering
		\subcaptionbox*{}{\includegraphics[width=.24\textwidth]{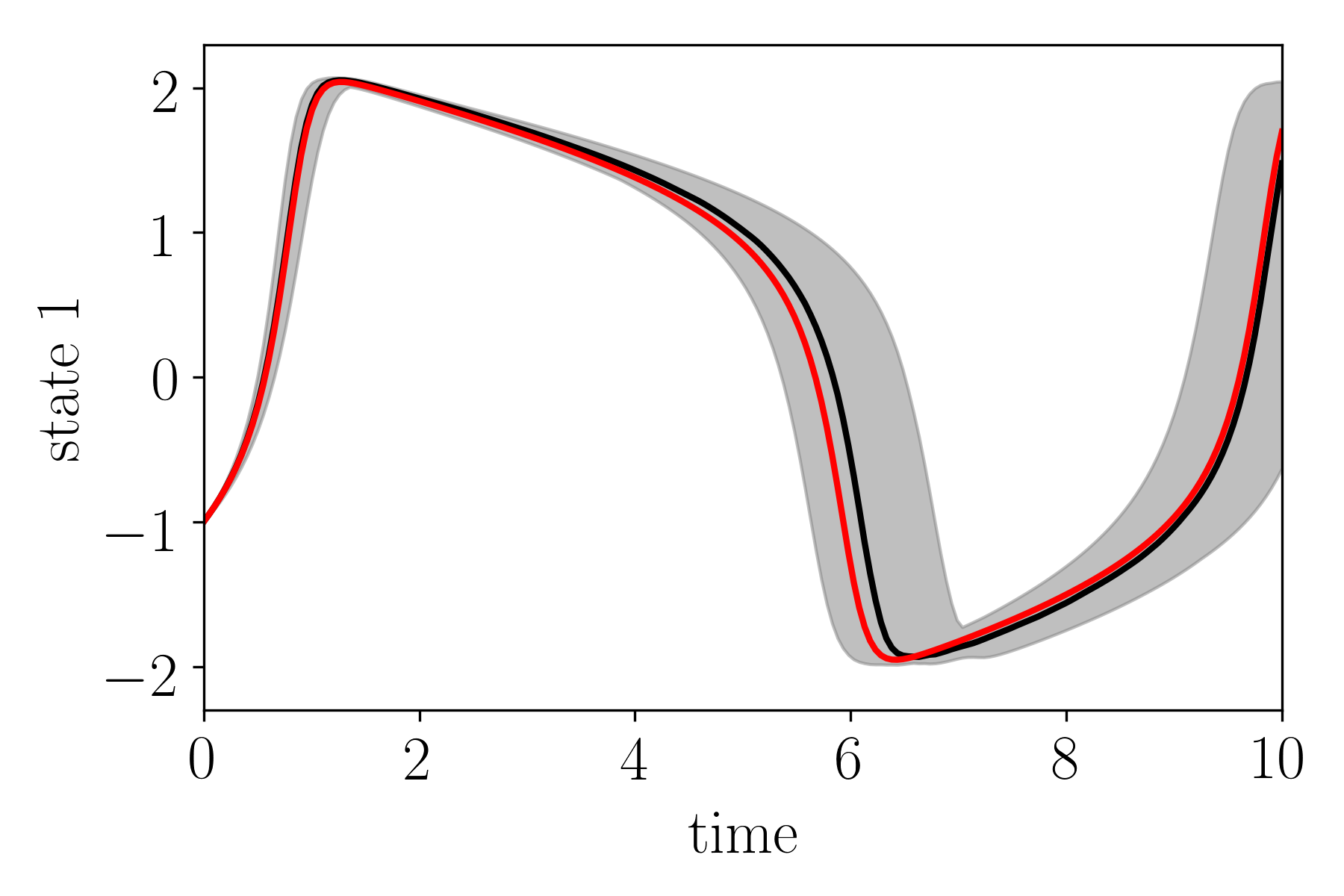}}
		\subcaptionbox*{}{\includegraphics[width=.24\textwidth]{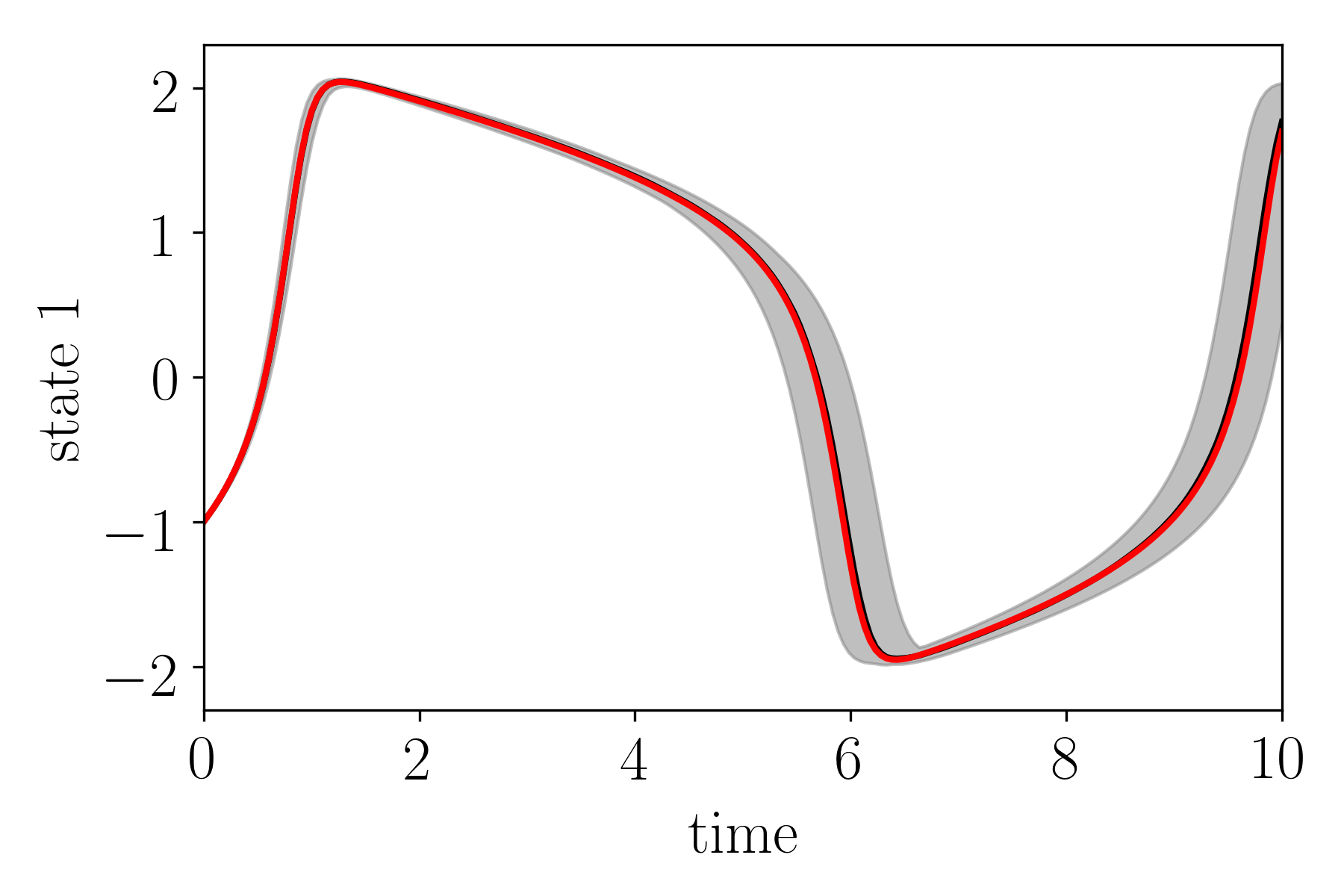}}
		\subcaptionbox*{}{\includegraphics[width=.24\textwidth]{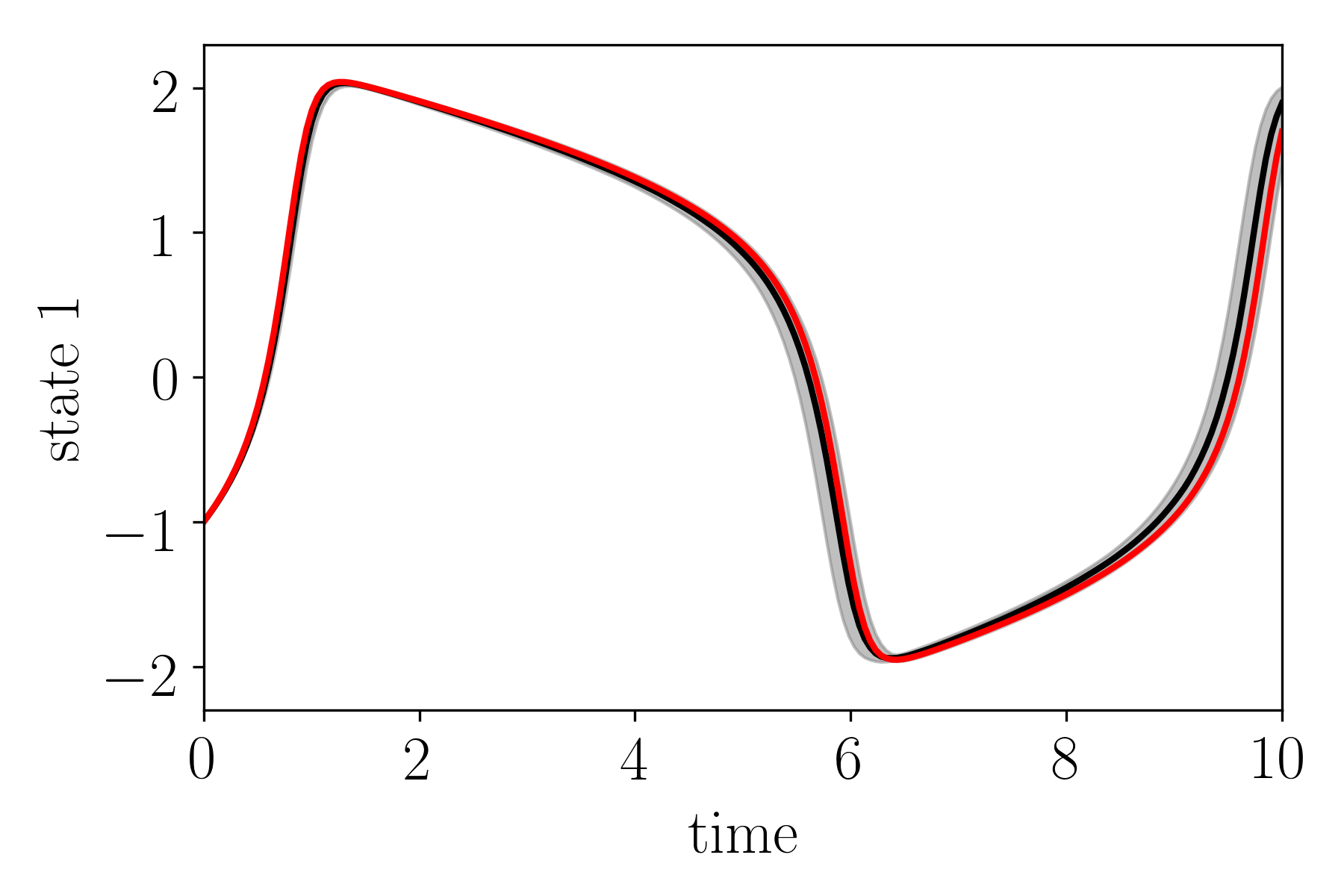}}
		\subcaptionbox*{}{\includegraphics[width=.24\textwidth]{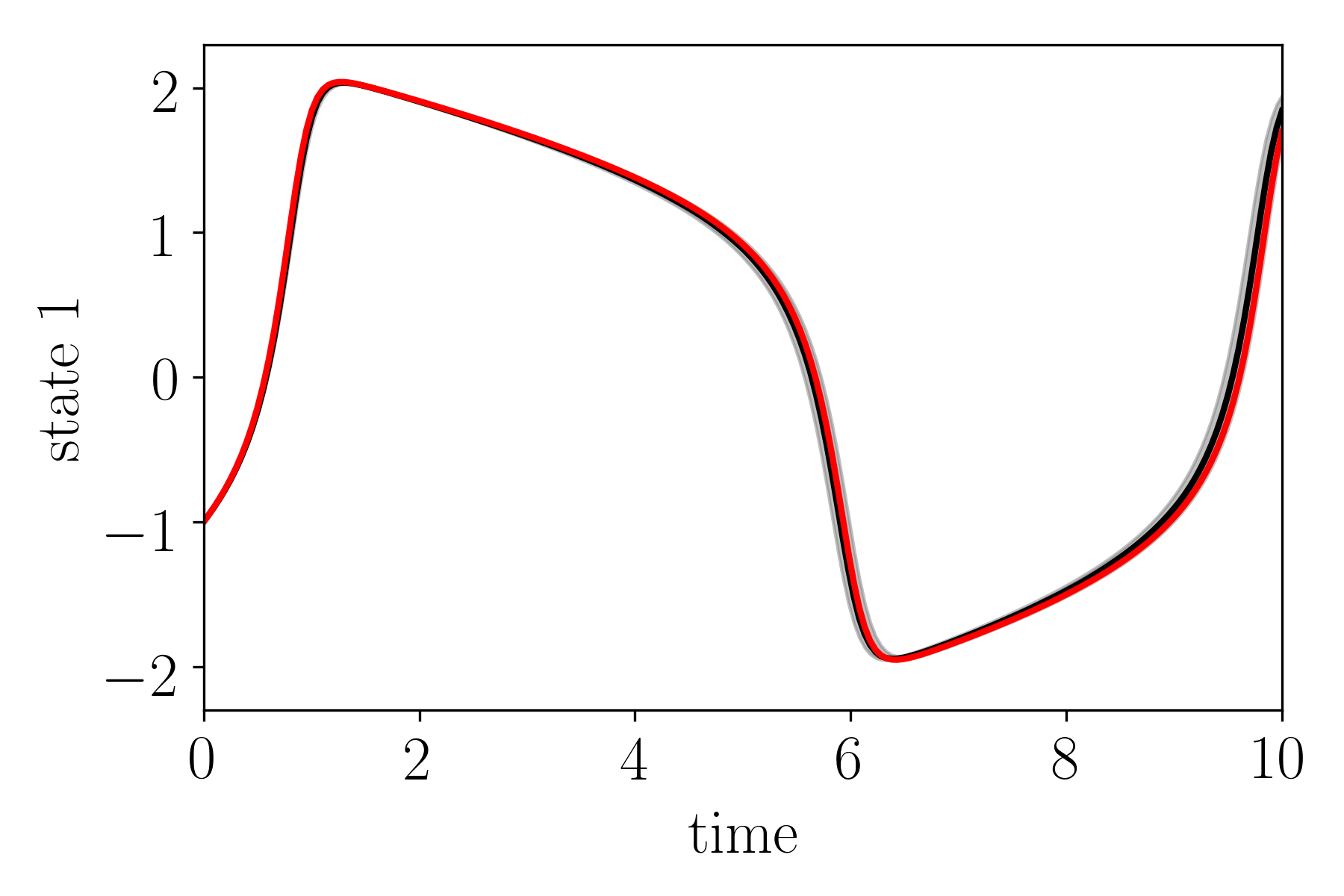}}
		\vspace{-20pt}      
		\caption{Median plots of the first state of FHN with SNR 100 in case of 10, 25, 50 and 100 observations. Same colors as in Figure \ref{fig:FHNHighNoise}. Plots of state 2 can be found in the supplementary material.}
		\label{fig:FHNLowNoise}
	\end{minipage}
\vspace{-8pt}
\end{figure*}

\subsection{Spiky Dynamics}
As can be seen in Figure \ref{fig:StateLVHighNoise}, all GP based gradient matching algorithms converge to parameter settings where the trajectories are smoother than the ground truth. While learning the hyperparameters in a pre-processing step clearly reduces this effect for $\mathrm{FGPGM}$ and $\mathrm{MVGM}$, there is still some bias. If only few observations are available, the GP prior on the states tends to smooth out part of the system dynamics. This "smoothing bias" is then passed on to the ODE parameters in the gradient matching scheme.

To investigate the practical importance of this effect, we evaluate $\mathrm{FGPGM}$ on a third system proposed by \citet{fitzhugh1961impulses} and \citet{nagumo1962active} for modeling giant squid neurons. Abbreviating the name of its inventors, we will refer to it as the FHN system. Its dynamics are given by the two ODEs
\begin{align}
\dot{V} &= \theta_1( V - \frac{V^3}{3} + R)\\
\dot{R} &= \frac{1}{\theta_1} ( V - \theta_2 + \theta_3 R)
\end{align}
Due to its highly nonlinear terms, the FHN system has notoriously fast changing dynamics. One example realization including noisy observations is shown in Figure \ref{fig:FHNHighNoise}. To account for the spikier behavior of the system, we used a Mat\'ern52 kernel. $\gamma$ was set to $3*10^{-4}$.

While the bias towards smoother trajectories is clearly visible, $\mathrm{FGPGM}$ finds parameters that tightly hug the ground truth. As to be expected, the smoothing bias gets smaller if we add more observations. Furthermore, increasing the SNR to 100 as shown in Figure \ref{fig:FHNLowNoise} leads to an impressive accuracy, even if we reduce the amount of observations to just 10, especially as $\mathrm{FGPGM}$ is a hyper-prior free, statistical method
\section{DISCUSSION}
\label{sec:Discussion}
Gradient matching is a successful  tool to circumvent the computational cost of numerical integration for Bayesian parameter identification in dynamical systems, especially if the dynamics, like in most real world systems, are reasonably smooth. Previous Gaussian process-based approaches used a criticized product of experts heuristics, which leads to technical difficulties. We illustrated these theoretical problems and provided a novel, sound formulation that does not rely on a PoE. We furthermore explained the surprising performance gains of variational over sampling-based approaches and then combined these insights to propose a new algorithm called $\mathrm{FGPGM}$, which jointly learns states and parameters with improved state-of-the-art performance in terms of accuracy, run time, robustness and reduction of "smoothing bias" for general nonlinear dynamical systems.

Unlike the previous MCMC approaches, $\mathrm{FGPGM}$ uses a one-chain Metropolis Hastings scheme, which is much easier to tune than the previously used, complicated multi chain setups. Due to the sequential fitting of the hyperparameters, the wild behavior of the probability density motivating the setup in \citet[][section 3]{Dondelinger} and the need for hard to motivate hyperpriors is successfully avoided. Consequently, the inference is significantly more accurate, faster, more robust and significantly decreases the smoothing bias of GP based gradient matching schemes.

\subsubsection*{Acknowledgements}

This research was partially supported by the Max Planck ETH Center for Learning Systems and SNSF grant 200020\_159557.

\nocite{*}

%\vskip 0.2in
\clearpage
\bibliographystyle{plainnat}
\bibliography{bibliography}

\clearpage
\section{Supplementary Material}
\subsection{Derivatives of a Gaussian process}
\label{sec:AppendixDerivs}
\subsubsection{Definitions}
Following \citet{papoulis2002probability}, we define stochastic convergence and stochastic differentiability.
\begin{definition}
The RV $x_n$ \textbf{converges} to x in the MS sense (limit in mean) if  for some x
\begin{equation}
\lim\limits_{n \to\infty}\mathbb{E} (|x_n - x|) =0
\end{equation}
\end{definition}
\begin{definition}
The stochastic process x(t) is MS \textbf{differentiable} if for some $x'(t)$
\begin{equation}
\lim\limits_{\delta t\to 0} \mathbb{E}\left|\frac{x(t+\delta t) - x(t)}{\delta t} - x'(t)\right|=0
\label{eq:differentiability}
\end{equation}
\end{definition}
\begin{definition}
A stochastic process x(t) is called a \textbf{Gaussian Process}, if any finite number of samples of its trajectory are jointly Gaussian distributed according to a previously defined mean function $\mu(t)$ and a covariance matrix, that can be constructed using a predefined kernel function $k_\phi(t_i, t_j)$
\end{definition}
\subsubsection{GP and its derivative are jointly Gaussian}
Let $t_0, \delta t \in \mathbb{R}$.
\\\noindent
Let $x(t)$ be a Gaussian Process with constant mean $\mu$ and kernel function $k_\phi(t_1, t_2)$, assumed to be MS differentiable.
\\\\\noindent
From the definition of GP, we know that $x(t_0 + \delta t)$ and $x(t_0)$ are jointly Gaussian distributed.
\begin{equation}
\left[\begin{matrix}
x(t_0) \\
x(t_0 + \delta t)
\end{matrix}\right] \sim \mathcal{N}\left(
\left[\begin{matrix}
\mu \\
\mu
\end{matrix}\right],
\mathbf{\Sigma}\right)
\end{equation}
where $\mathbf{\Sigma}_{i,j} = k_\phi(\mathbf{t}_i,\mathbf{ t}_j)$ using $\mathbf{t}=[t_0, t_0 + \delta t]$.\\
\\\noindent
Using the linear transformation
\begin{equation}
\mathbf{T} = \frac{1}{\delta t}\left[\begin{matrix}
\delta t & 0 \\
-1 & 1
\end{matrix}\right]
\end{equation}
one can show that
\begin{equation}
\left[\begin{matrix}
x(t_0) \\
\frac{x(t_0 + \delta t) - x(t_0)}{\delta t}
\end{matrix}\right]
\sim \mathcal{N}\left(
\left[\begin{matrix}
\mu  \\
0
\end{matrix}\right],
\mathbf{T} \mathbf{\Sigma} \mathbf{T}^T
\right)
\end{equation}
So it is clear that for all $\delta t$, $x(t_0)$ and $\frac{x(t_0 + \delta t) - x(t_0)}{\delta t}$ are jointly Gaussian distributed. Using the assumption that x is differentiable according to the definition in eq. \ref{eq:differentiability} and the fact that convergence in expectation implies convergence in distribution, it is clear that $x(t_0)$ and $\dot{x}(t_0)$ are jointly Gaussian.

This fact can be easily extended to any finite set of sample times $\mathbf{t} = [t_0, t_1, ..., t_N]$. One can use the exact same procedure to show that the resulting vectors $\mathbf{x}(\mathbf{t})$ and $\mathbf{\dot{x}}(\mathbf{t})$ are jointly Gaussian as well.

\subsubsection{Moments of the joint distribution}
As shown in the previous section, any finite set of samples $\mathbf{x}$ is jointly Gaussian together with its derivatives $\mathbf{\dot{x}}$. To calculate the full distribution, it thus suffices to calculate the mean and the covariance between the elements of the full vector 
\begin{equation}
\left[\begin{matrix}\mathbf{x}\\ \mathbf{\dot{x}}\end{matrix}\right]
\sim \mathcal{N}\left(
\left[\begin{matrix}
\boldsymbol{\mu} \\
\mathbf{0}
\end{matrix}\right],
\left[\begin{matrix}
\mathbf{C_\phi} & \mathbf{C_\phi'} \\
\mathbf{'C_\phi} & \mathbf{C_\phi''}
\end{matrix}\right]
\right)
\end{equation}
$\mathbf{C_\phi}$ is the predefined kernel matrix of the Gaussian Process.
\\\\\noindent
$\mathbf{'C_\phi}$ can be calculated by  directly using the linearity of the covariance operator.

\begin{align}
\mathbf{'C_\phi}_{i, j} &= \textrm{cov}(\dot{x}(t_i), x(t_j)) \\ \nonumber
&= \textrm{cov}\left(\frac{d}{da} x(a) \middle | _{a=t_i}, x(t_j)\right) \\ \nonumber
&= \frac{d}{da} \textrm{cov}\left( x(a) , x(t_j)\right)_{a=t_i} \\ \nonumber
&= \frac{d}{da}k_\phi(a, t_j)_{a=t_i}
\end{align}
\\\noindent
Obviously, $\mathbf{C_\phi'}$ is just the transposed of $\mathbf{'C_\phi}$, while $\mathbf{C_\phi''}$ can be calculated in exactly the same manner to obtain
\begin{equation}
\mathbf{C_\phi''}_{i, j} = \frac{d}{da} \frac{d}{db} k_\phi(a, b)_{a=t_i, b=t_j}
\end{equation}

\subsubsection{Conditional GP for derivatives}
To obtain the GP over the derivatives given the states, the joint distribution
\begin{equation}
\left[\begin{matrix}\mathbf{x}\\ \mathbf{\dot{x}}\end{matrix}\right]
\sim \mathcal{N}\left(
\left[\begin{matrix}
\boldsymbol{\mu} \\
\mathbf{0}
\end{matrix}\right],
\left[\begin{matrix}
\mathbf{C_\phi} & \mathbf{C_\phi'} \\
\mathbf{'C_\phi} & \mathbf{C_\phi''}
\end{matrix}\right]
\right)
\end{equation}
has to be transformed. This can be done using standard techniques as described e.g. in section 8.1.3 of \citet{MatrixCookbook}. There, it is written:
\\\\\noindent
Define
\begin{equation}
\mathbf{x} = \left[\begin{matrix}
\mathbf{x_a} \\
\mathbf{x_b}
\end{matrix}\right]
,
\boldsymbol{\mu} = \left[\begin{matrix}
\boldsymbol{\mu_a} \\
\boldsymbol{\mu_b}
\end{matrix}\right]
,
\boldsymbol{\Sigma} = \left[\begin{matrix}
\boldsymbol{\Sigma_a} & \boldsymbol{\Sigma_c} \\
\boldsymbol{\Sigma_c}^T & \boldsymbol{\Sigma_b}
\end{matrix}\right]
\end{equation}
Then
\begin{equation}
p(\mathbf{x_b} |\mathbf{x_a})  \sim \mathcal{N}(\hat{\boldsymbol{\mu}}_b, \hat{\boldsymbol{\Sigma}}_b)
\end{equation}
where
\begin{align}
\hat{\boldsymbol{\mu}}_b &= \boldsymbol{\mu_b} + \boldsymbol{\Sigma_c}^T\boldsymbol{\Sigma_a}^{-1}(\mathbf{x_a} - \boldsymbol{\mu_a}) \\
\hat{\boldsymbol{\Sigma}}_b &= \boldsymbol{\Sigma_b} - \boldsymbol{\Sigma_c}^T \boldsymbol{\Sigma_a}^{-1} \boldsymbol{\Sigma_c}
\end{align}
\\\noindent
Applied to the above probability distribution, this leads to
\begin{align}
p(\dot{\mathbf{x}}| \mathbf{x}) &\sim \mathcal{N}(\mathbf{D}\mathbf{x}, \mathbf{A})
\end{align}
using
\begin{align}
\mathbf{D} &= \mathbf{'C_\phi}\mathbf{C_\phi}^{-1}\\
\mathbf{A} &= \mathbf{C_\phi''} - \mathbf{'C_\phi}\mathbf{C_\phi}^{-1}\mathbf{C_\phi'}
\end{align}

\subsection{Proof of theorem 1}
\label{sec:proof1}
\begin{proof}
    The proof of this statement follows directly by combining all the previous definitions and marginalizing out all the random variables that are not part of the end result.
    
    First, one starts with the joint density over all variables as stated in equation \eqref{eq:AlternativeGraphModelEquations}
    \begin{align*}
        p(\mathbf{x}, \mathbf{\dot{x}}, &\mathbf{y}, \mathbf{F_1}, \mathbf{F_2}, \boldsymbol{\theta} | \boldsymbol{\phi}, \sigma, \gamma)= \\
        &p_{\gp}(\mathbf{x}, \mathbf{\dot{x}}, \mathbf{y}| \boldsymbol{\phi}, \sigma)
        p_{\ode}(\mathbf{F_1}, \mathbf{F_2}, \boldsymbol{\theta} |\mathbf{x}, \mathbf{\dot{x}}, \gamma),
    \end{align*}
    where
    \begin{equation}
            \nonumber p_{\gp}(\mathbf{x}, \mathbf{\dot{x}}, \mathbf{y}| \boldsymbol{\phi}, \sigma)=
            p(\mathbf{x} | \boldsymbol{\phi}) % GP prior
            p(\mathbf{\dot{x}} | \mathbf{x}, \boldsymbol{\phi}) % derivative prior
            p(\mathbf{y}|\mathbf{x}, \sigma) % observation
    \end{equation}
    and
    \begin{align*}
            p_{\ode}(&\mathbf{F_1}, \mathbf{F_2}, \boldsymbol{\theta} |\mathbf{x}, \mathbf{\dot{x}}, \gamma)= \\
            &p(\boldsymbol{\theta}) %parameter priors
            p(\mathbf{F_1} | \boldsymbol{\theta}, \mathbf{x}) % F_1 density
            p(\mathbf{F_2} | \mathbf{\dot{x}}, \gamma \mathbf{I}) %F_2 density
            \delta(\mathbf{F_1} - \mathbf{F_2}).    \nonumber         
    \end{align*}
    To simplify this formula, $p_{\ode}$ can be reduced by marginalizing out $\mathbf{F_2}$ using the properties of the Dirac delta function and the probability density defined in equation \eqref{eq:definitionOfF2Density}. The new $p_{\ode}$ is then independent of $\mathbf{F_2}$.
    \begin{align*}
        p_{\ode}(\mathbf{F_1}, \boldsymbol{\theta} |\mathbf{x}, \mathbf{\dot{x}}, \gamma) = 
        p(\boldsymbol{\theta}) %parameter priors
        p(\mathbf{F_1} | \boldsymbol{\theta}, \mathbf{x}) % F_1 density
        \mathcal{N}(\mathbf{F_1} | \mathbf{\dot{x}}, \gamma \mathbf{I}). %F_2 density
    \end{align*}
    Inserting equation \eqref{eq:NewODEDensityDef} yields
    \begin{equation*}
        p_{\ode}(\mathbf{F_1}, \boldsymbol{\theta} |\mathbf{x}, \mathbf{\dot{x}}, \gamma) =
        p(\boldsymbol{\theta}) %parameter priors
        \delta(\mathbf{F_1} - \mathbf{f}(\mathbf{x}, \boldsymbol{\theta}))
        \mathcal{N}(\mathbf{F_1} | \mathbf{\dot{x}}, \gamma \mathbf{I}). %F_2 density
    \end{equation*}
    Again, the properties of the Dirac delta function are used to marginalize out $\mathbf{F_1}$. The new $p_{\ode}$ is now independent of $\mathbf{F_1}$,
    \begin{align*}
        p_{\ode}(\boldsymbol{\theta} |\mathbf{x}, \mathbf{\dot{x}}, \gamma) = 
        p(\boldsymbol{\theta}) %parameter priors
        \mathcal{N}(\mathbf{f}(\mathbf{x}, \boldsymbol{\theta}) | \mathbf{\dot{x}}, \gamma \mathbf{I}). %F_2 density
    \end{align*}
    This reduced $p_{\ode}$ is now combined with $p_{\gp}$. Observing that the mean and the argument of a normal density are interchangeable and inserting the definition of the GP prior on the derivatives given by equation \eqref{eq:GPDerivs} leads to
    \begin{align*}
        p(\mathbf{x}, &\mathbf{\dot{x}}, \mathbf{y}, \boldsymbol{\theta} | \boldsymbol{\phi}, \sigma, \gamma) =\\
        &p(\boldsymbol{\theta}) %parameter priors
        p(\mathbf{x} | \boldsymbol{\phi}) % GP prior
        \mathcal{N}(\dot{\mathbf{x}} | \mathbf{D} \mathbf{x}, \mathbf{A}) % derivative prior
        p(\mathbf{y}|\mathbf{x}, \sigma) % observation
        \mathcal{N}(\mathbf{\dot{x}} |\mathbf{f}(\mathbf{x}, \boldsymbol{\theta}) , \gamma \mathbf{I}). %F_2 density
    \end{align*}
    $\mathbf{\dot{x}}$ can now be marginalized by observing that the product of two normal densities of the same variable is again a normal density. The formula can be found, e.g., in \citet{MatrixCookbook}. As a result, one obtains
    \begin{align*}
        p(\mathbf{x}, &\mathbf{y}, \boldsymbol{\theta} | \boldsymbol{\phi}, \sigma, \gamma) =\\
        &p(\boldsymbol{\theta}) %parameter priors
        p(\mathbf{x} | \boldsymbol{\phi}) % GP prior
        p(\mathbf{y}|\mathbf{x}, \sigma) % observation
        \mathcal{N}(\mathbf{f}(\mathbf{x}, \boldsymbol{\theta}) | \mathbf{D} \mathbf{x}, \mathbf{A} + \gamma \mathbf{I}).
    \end{align*}
    It should now be clear that after inserting equations \eqref{eq:ObsModel} and \eqref{eq:GPPrior} and renormalizing, we get the final result
    \begin{align*}
        p(&\mathbf{x}, \boldsymbol{\theta}|\mathbf{y}, \boldsymbol{\phi}, \gamma, \sigma) \propto \\
        &p(\boldsymbol{\theta}) 
        \mathcal{N}(\mathbf{x} | \mathbf{0}, \boldsymbol{C_\phi})
        \mathcal{N}(\mathbf{y} | \mathbf{x}, \sigma^2 \mathbf{I})
        \mathcal{N}(\mathbf{f}(\mathbf{x}, \boldsymbol{\theta}) | \mathbf{D} \mathbf{x}, \mathbf{A} + \gamma \mathbf{I}),
    \end{align*}
    concluding the proof of this theorem.
\end{proof}

\subsection{Hyperparameter and kernel selection}
\label{sec:HyperAndKernelSelection}
As discussed before, the Gaussian process model is defined by a kernel function $k_\phi(t_i, t_j)$. For both the hyperparameters $\boldsymbol{\phi}$ and the functional form of $k$ there exist many possible choices. Even though the exact choice might not be too important for consistency guarantees in GP regression \citep{consistencyPaper}, this choice directly influences the amount of observations that are needed for reasonable performance. While there exist some interesting approaches to learn the kernel directly from the data, e.g., \citet{AutomaticKernelBuilder} and \citet{KernelSelection}, these methods can not be applied due to the very low amount of observations of the systems considered in this paper. As in previous approaches, the kernel functional form is thus restricted to simple kernels with few hyperparameters, whose behaviors have already been investigated by the community, e.g., in the kernel cookbook by \citet{DuvenaudThesis}. Once a reasonable kernel is chosen, it is necessary to fit the hyperparameter and depending on the amount of expert knowledge available, there are different methodologies.

\subsubsection{Maximizing the data likelihood}
\label{sec:MaxEvTheory}
As mentioned e.g. in \citet{RasmussenGPBook}, it is possible for a Gaussian process model to analytically calculate the marginal likelihood of the observations $\mathbf{y}$ given the evaluation times $\mathbf{t}$ and hyperparameters $\boldsymbol{\phi}$ and $\sigma$.
\begin{align}
\label{eq:marginalHyperparams}
\log (&p(\mathbf{y} | \mathbf{t}, \boldsymbol{\phi}, \sigma)) = \nonumber\\
&-\frac{1}{2} \mathbf{y}^T (\mathbf{C}_\phi + \sigma \mathbf{I})^{-1}\nonumber\\
&-\frac{1}{2} \log | \mathbf{C}_\phi + \sigma \mathbf{I}|\nonumber\\
&-\frac{n}{2} \log 2 \pi
\end{align}

where $\sigma$ is the GPs estimate for the standard deviation of the observation noise and n is the amount of observations.

This equation is completely independent of the ODE system one would like to analyze and depends only on the observations of the states. To fit the GP model to the data, equation \eqref{eq:marginalHyperparams} can be maximized w.r.t. $\boldsymbol{\phi}$ and $\sigma$, without incorporating any prior knowledge.

\subsubsection{Concurrent optimization}
In $\mathrm{AGM}$ of \citet{Dondelinger}, the hyperparameters are not calculated independent of the ODEs. Instead, a prior is defined and their posterior distribution is determined simultaneously with the posterior distribution over states and parameters by sampling from equation \eqref{eq:DondelingerDensity}.

This approach has several drawbacks. As we shall see in section \ref{sec:Experiments}, its empirical performance is significantly depending on the hyperpriors. Furthermore, optimizing the joint distribution given equation \eqref{eq:DondelingerDensity} requires calculating the inverse of the covariance matrices $\mathbf{C_\phi}$ and $\mathbf{A}$, which has to be done again and again for each new set of hyperparameters. Due to the computational complexity of matrix inversion, this is significantly slowing down the optimization.

For these reasons, if strong prior knowledge about the hyperparameters is available, it might be better to incorporate it into the likelihood optimization presented in the previous section. There, it could be used as a hyperprior to regularize the likelihood.

\subsubsection{Manual tuning}
In the variational inference approach \citet{VGM}, the hyperparameters were assumed to be provided by an expert. If such expert knowledge is available, it should definitely be used since it can improve the accuracy drastically.

\subsection{Adjusting the GP model}
\label{sec:StandardizingData}
To make $\mathrm{VGM}$ more comparable to $\mathrm{AGM}$, the hyperparameters of the kernel must be learned from the data. However, maximizing the data likelihood described in equation \eqref{eq:marginalHyperparams} directly using the prior defined in equation \eqref{eq:GPPrior} will lead to very bad results.

\subsubsection{Zero mean observations}
The main reason for the poor performance without any pretreatment of $\mathbf{y}$ is the fact that the zero mean assumption in equation \eqref{eq:marginalHyperparams} is a very strong regularization for the amount of data available. As its effect directly depends on the distance of the true values to zero, it will be different for different states in multidimensional systems, further complicating the problem. Thus, it is common in GP regression to manipulate the observations such that they have zero mean.

This procedure can be directly incorporated into the joint density given by equation \eqref{eq:DondelingerDensity}. It should be noted that for multidimensional systems this joint density will factorize over each state k, whose contribution will be given by
\begin{align}
p(\mathbf{x_k}|&\mathbf{y_k},\boldsymbol{\theta},  \boldsymbol{\phi}, \gamma, \sigma) \propto \mathcal{N}(\mathbf{\tilde{x}_k} | \mathbf{0}, \boldsymbol{C_{\phi_k}})\nonumber\\
& \times \mathcal{N}(\mathbf{y_k} | \mathbf{x_k}, \sigma^2 \mathbf{I})\nonumber\\
& \times \mathcal{N}(\mathbf{f}_k(\mathbf{x_k}, \boldsymbol{\theta}) | \mathbf{D_k} \mathbf{\tilde{x}_k}, \mathbf{A_k} + \gamma \mathbf{I})
\end{align}
where
\begin{align}
\mathbf{\tilde{x}_k}&=\mathbf{x_k}- \mu_{y,k}\mathbf{1} \nonumber
\end{align} 
using $\mu_{y,k}$ to denote the mean of the observations of the k-th state and $\mathbf{1}$ to denote a vector of ones with appropriate length.

It is important to note that this transformation is not just equivalent to exchanging $\mathbf{x_k}$ and $\mathbf{\tilde{x}_k}$. While the transformation is not necessary for the observation term, as $\mathbf{x_k}$ and $\mathbf{y_k}$ would be shifted equally, the original $\mathbf{x_k}$ is needed as input to the ODEs. This allows for this transformation without the need to manually account for this in the structural form of the differential equations.

This trick will get rid of some of the bias introduced by the GP prior. In the simulations, this made a difference for all systems, including the most simple one presented in section \ref{sec:LotkaVolterra}.

\subsubsection{Standardized states}
If the systems get more complex, the states might be magnitudes apart from each other. If one were to use the same hyperparameters $\boldsymbol{\phi}$ for all states, then a deviation {$(\mathbf{F_k}-\mathbf{D_k}\mathbf{\tilde{x}_k}) = 10^{-4}$} would contribute equally to the change in probability, independent of whether the states $\mathbf{\tilde{x}_k}$ are of magnitude $10^{-8}$ or $10^{3}$. Thus, small relative deviations from the mean of states with large values will lead to stronger changes in the joint probability than large relative deviations of states with small values. This is not a desirable property, which can be partially alleviated by calculating a new set of hyperparameters for each state. However, this problem can be completely nullified by standardizing the data $\mathbf{y}$. For the k-th state, this would change its contribution to the joint density to
\begin{align}
p(\mathbf{x_k}|&\mathbf{y_k},\boldsymbol{\theta},  \boldsymbol{\phi}, \gamma, \sigma) \propto 
\mathcal{N}\left(\frac{1}{\sigma_{y, k}}\mathbf{\tilde{x}_k} \middle| \mathbf{0}, \boldsymbol{C_{\phi_k}}\right) \\\nonumber
&\times\mathcal{N}\left(\frac{1}{\sigma_{y, k}}\mathbf{y_k} \middle| \frac{1}{\sigma_{y, k}}\mathbf{x_k}, \sigma^2 \mathbf{I}\right) \\\nonumber
&\times\mathcal{N}\left(\frac{1}{\sigma_{y, k}}\mathbf{f}_k(\mathbf{x_k}, \boldsymbol{\theta}) \middle| \frac{1}{\sigma_{y, k}}\mathbf{D_k} \mathbf{\tilde{x}_k}, \mathbf{A_k} + \gamma \mathbf{I}\right)
\end{align}
where
\begin{align}
\mathbf{\tilde{x}_k}&=\mathbf{x_k}- \mu_{y,k}\mathbbm{1} \nonumber
\end{align} 
and $\sigma_{y, k}$ is the standard deviation of the observations of the k-th state.

For complex systems with states on different orders of magnitude, standardization is a must to obtain reasonable performance. Even for the system presented in section \ref{sec:ProteinTransduction}, standardization has a significantly beneficial effect, although the states do not differ greatly in magnitude.

\subsection{Algorithmic details}
\label{sec:AppendixAlgoDetails}

For all experiments involving $\mathrm{AGM}$, the R toolbox \mbox{deGradInfer} \citep{codeAGM} published alongside \citet{MacdonaldThesis} was used.  For comparability, no priors were used, but the toolbox needed to be supplied with a value for the standard deviation of the observational noise and the true standard deviation of the noise was used. For all other parameters, e.g., amount of chains or samples, the values reported in \citet{Dondelinger} were used.

For $\mathrm{MVGM}$ and $\mathrm{FGPGM}$, the hyperparameters of the GP model were determined in a preprocessing step identical for both algorithms. After calculating the hyperparameters, the $\mathrm{MVGM}$ parameters were inferred using the implementation used by \citet{VGM}.

Both $\mathrm{MVGM}$ and $\mathrm{FGPGM}$ need to be supplied with $\gamma$, which was treated as a tuning parameter. In principle, this parameter could be found by evaluating multiple candidates in parallel and choosing based on data fit.

For both experiments, the amount of iterations recommended by the $\mathrm{AGM}$ toolbox \citep{codeAGM} have been used, namely 100'000 iterations for Lotka Volterra and 300'000 iterations for Protein Transduction. This is the same setup that has been used to obtain the parameter estimates shown throughout the paper. The simpler sampling setup of $\mathrm{FGPGM}$ clearly leads to running time savings of about a third compared to $\mathrm{AGM}$. Thus, $\mathrm{FGPGM}$ is not only significantly more accurate, but also significantly faster than $\mathrm{AGM}$. \citet{Dondelinger} have shown that this also implies order of magnitude improvements if compared to the running time of approaches based on numerical integration.

All experiments were performed using the ETH cluster.\footnote{https://scicomp.ethz.ch/wiki/Euler\#Euler\_I}.
It should be noted that the algorithms were implemented in different programming languages and the cluster consists of cores with varying computation power, adding some variability to the running time estimates.

\subsection{Lotka Volterra}
\label{sec:AppendixExperiments}
\begin{figure}[tbh!]
    \centering
    \begin{subfigure}[t]{.48\textwidth}
        \centering
        \includegraphics[width=\textwidth]{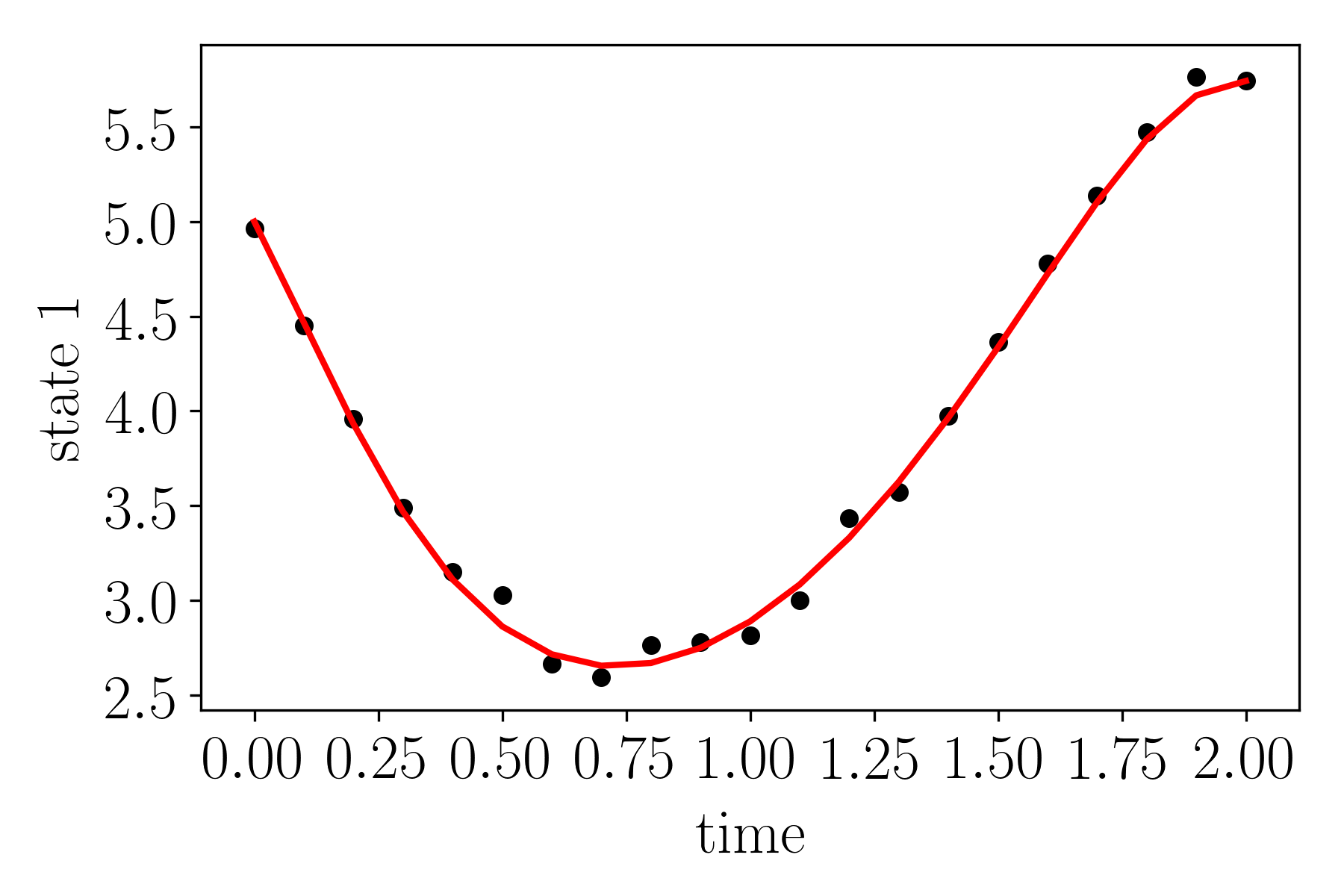}
    \end{subfigure}
    \begin{subfigure}[t]{.48\textwidth}
        \centering
        \includegraphics[width=\textwidth]{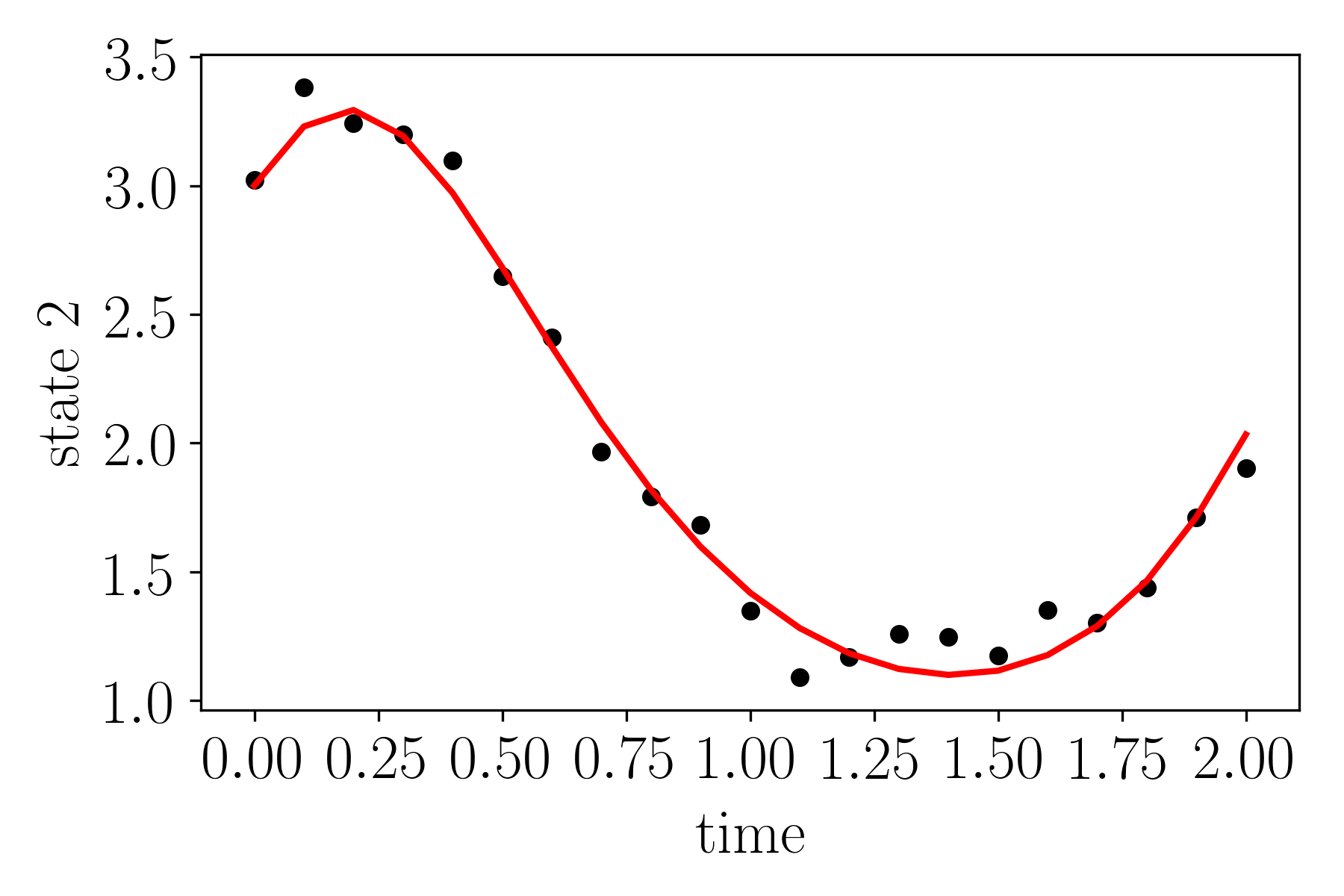}
    \end{subfigure}
    \caption{Example rollout of the Lotka Volterra system showing the state evolution over time. The dots denote the observations while the line represents the ground truth obtained by numerical integration.}
    \label{fig:LVProblemSetting}
\end{figure}

As in the previous publications, a squared exponential kernel was used. For $\mathrm{FGPGM}$ and $\mathrm{MVGM}$, $\gamma$ was set to 0.3, while $\mathrm{AGM}$ was provided with the true observation noise standard deviations $\sigma$. The standard deviation of the proposal distribution of $\mathrm{FGPGM}$ was chosen as 0.075 for state proposals and as 0.09 for parameter proposals to roughly achieve an acceptance rate of 0.234. For all algorithms, it was decided to use only one GP to fit both states. This effectively doubles the amount of observations and leads to more stable hyperparameter estimates. As the state dynamics are very similar, this approximation is feasible.

\begin{figure}[!tbh]
    \begin{subfigure}[t]{.48\textwidth}
        \centering
        \includegraphics[width=\textwidth]{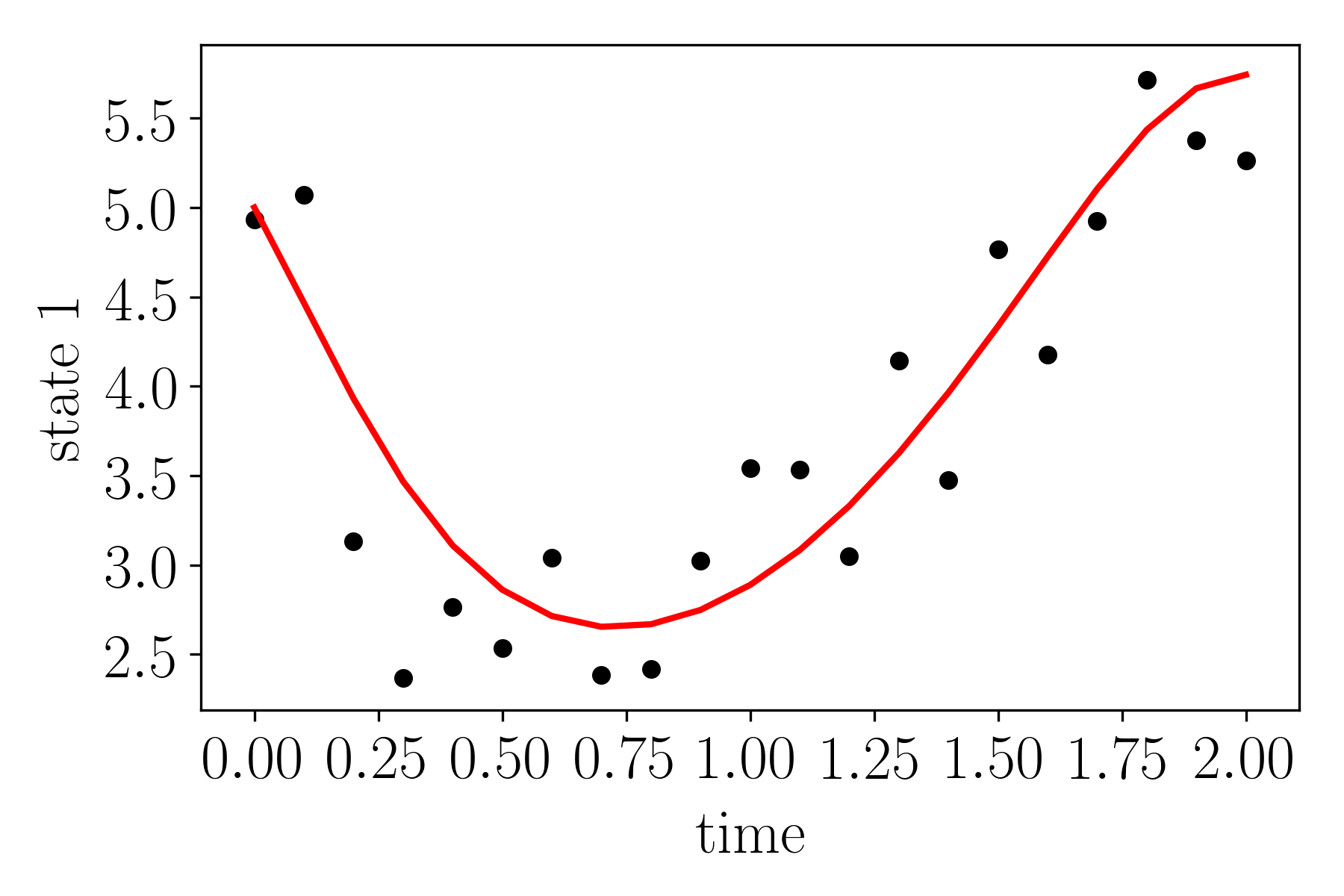}
    \end{subfigure}
    \begin{subfigure}[t]{.48\textwidth}
        \centering
        \includegraphics[width=\textwidth]{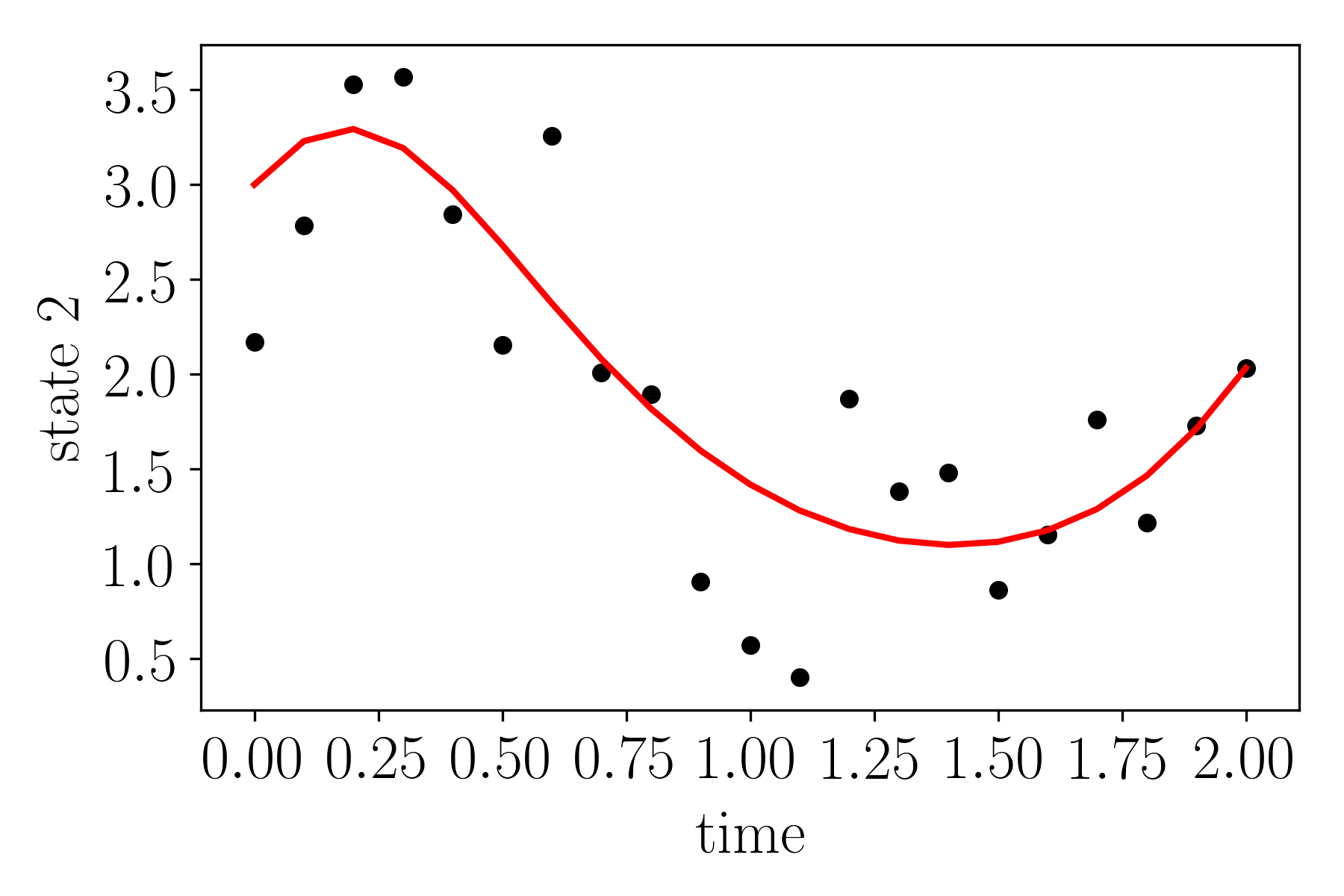}
    \end{subfigure}
    \caption{Example rollout of the Lotka Volterra system showing the state evolution over time for the high noise case. The dots denote the observations while the line represents the ground truth obtained by numerical integration.}
    \label{fig:LVProblemSettingHighNoise}
\end{figure}

\begin{figure*}[tbh!]
    \centering
    \begin{subfigure}[t]{.32\textwidth}
        \centering
        \includegraphics[width=\textwidth]{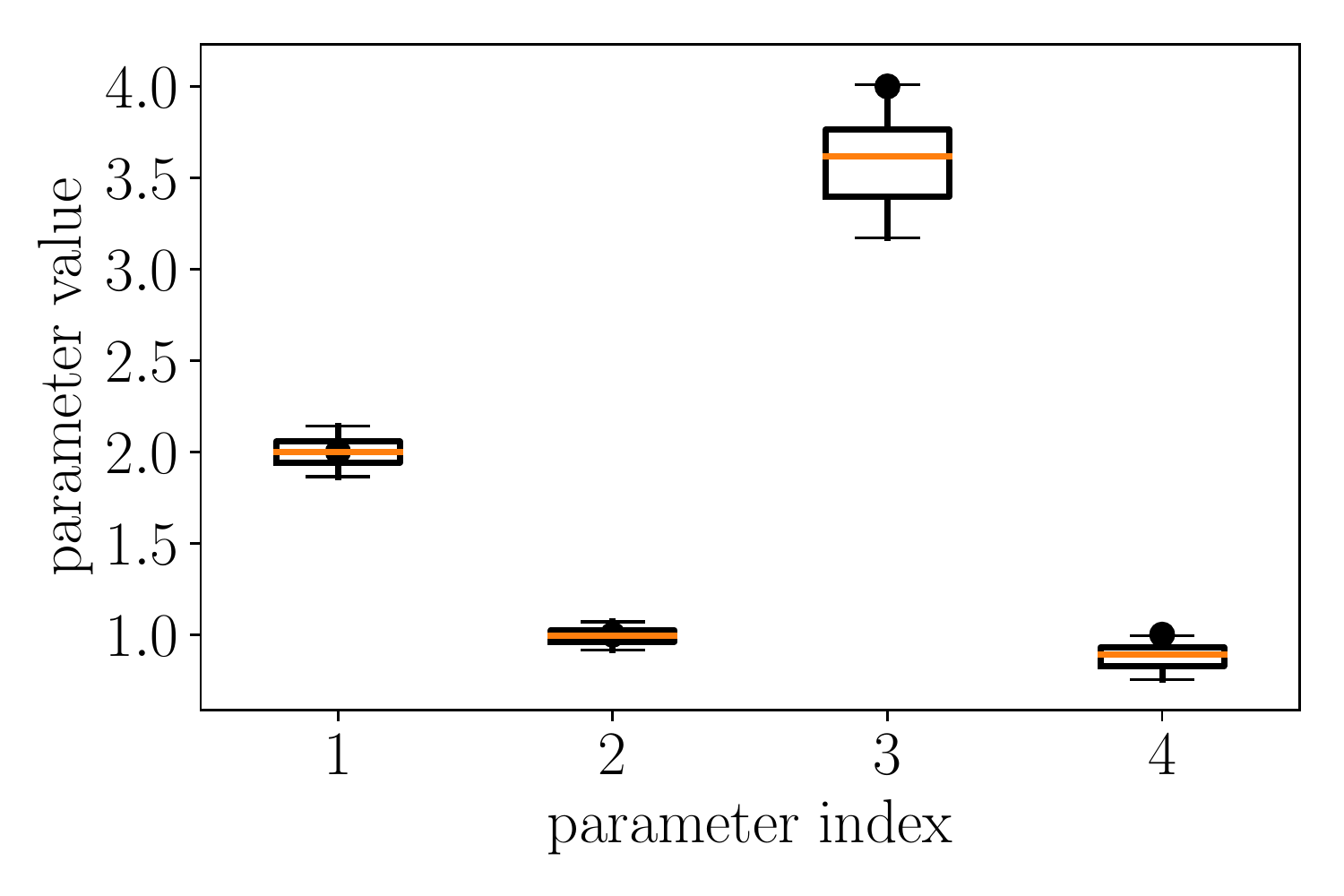}
        \caption{AGM}
    \end{subfigure}
    \begin{subfigure}[t]{.32\textwidth}
        \centering
        \includegraphics[width=\textwidth]{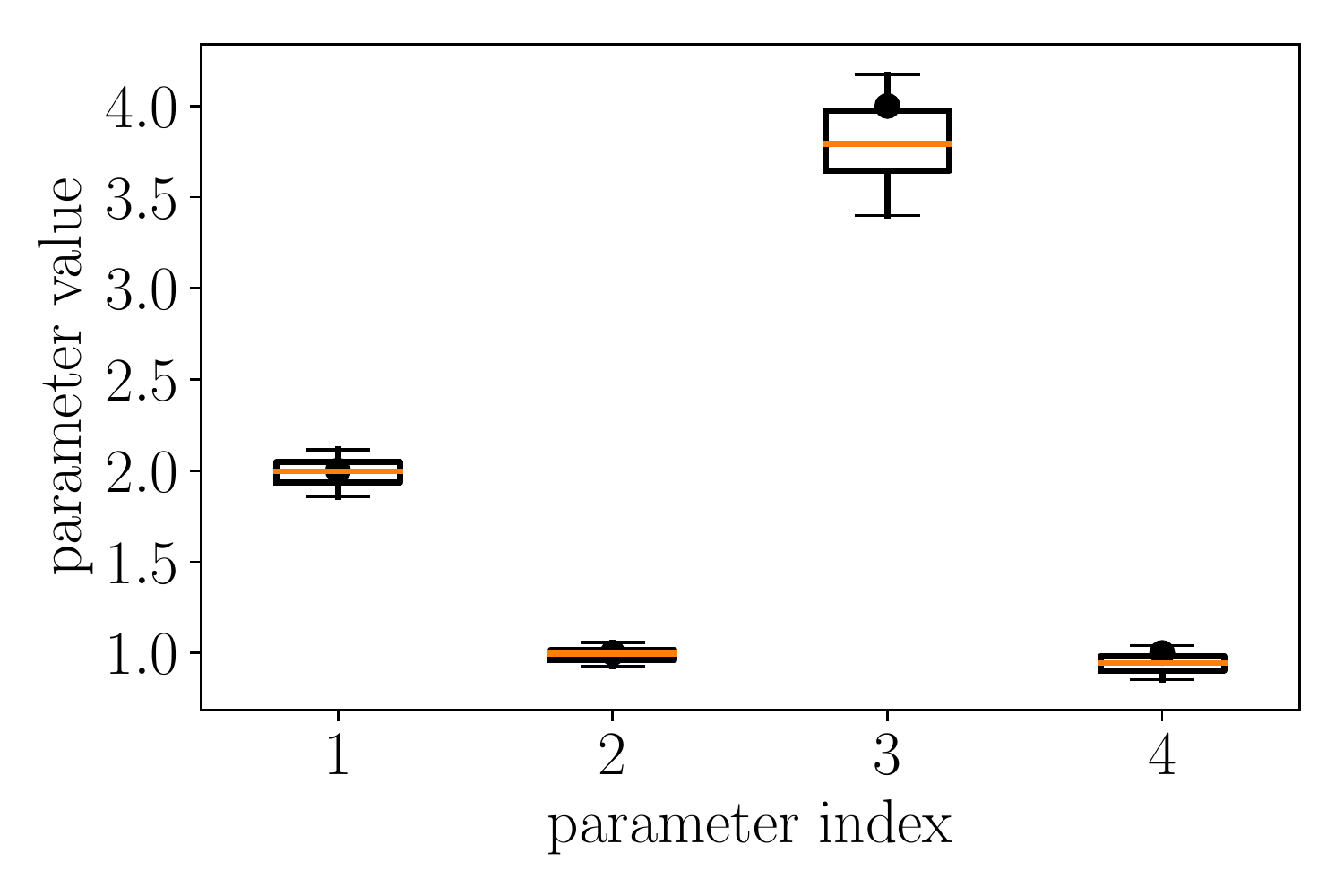}
        \caption{FGPGM}
    \end{subfigure}
    \begin{subfigure}[t]{.32\textwidth}
        \centering
        \includegraphics[width=\textwidth]{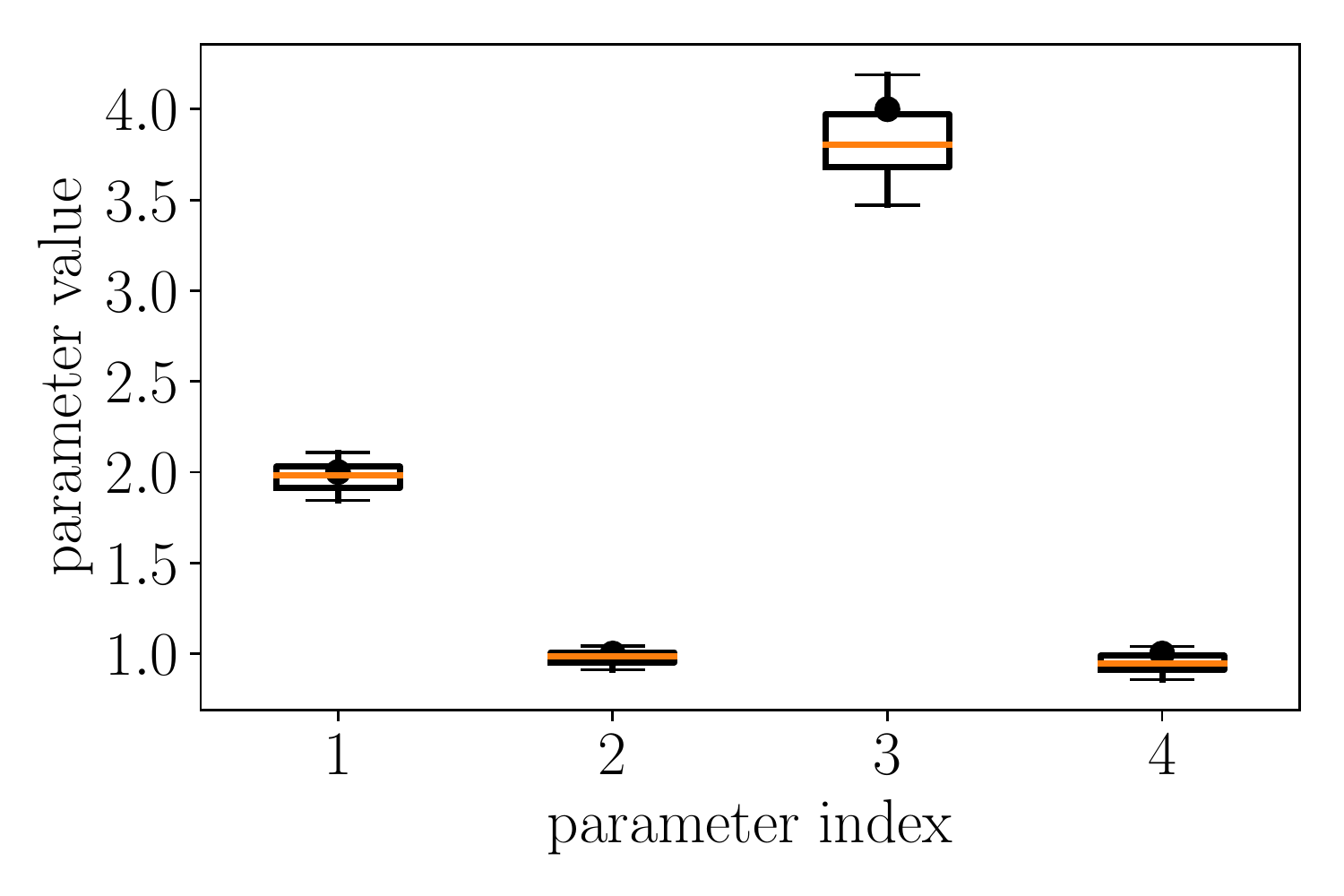}
        \caption{MVGM}
    \end{subfigure}
    \caption{Inferred parameters for the low noise case. Ground truth (black dots), median (orange line), 50\% (boxes) and 75\% (whiskers) quantiles evaluated over 100 independent noise realizations are shown.}
    \label{fig:LVLowNoise}
\end{figure*}

\begin{figure*}[!tbh]
    \begin{subfigure}[t]{.32\textwidth}
        \centering
        \includegraphics[width=\textwidth]{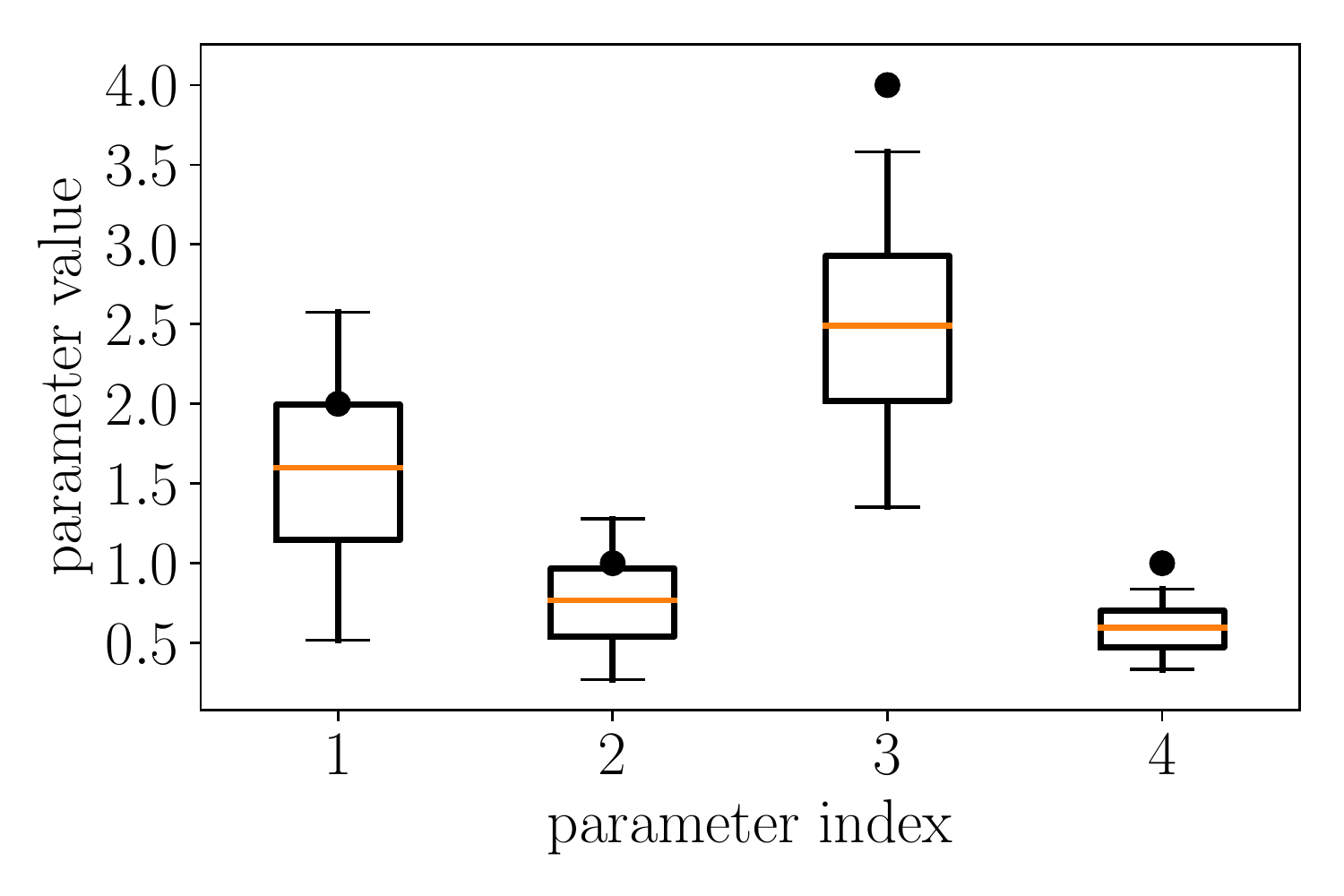}
        \caption{AGM}
    \end{subfigure}
    \begin{subfigure}[t]{.32\textwidth}
        \centering
        \includegraphics[width=\textwidth]{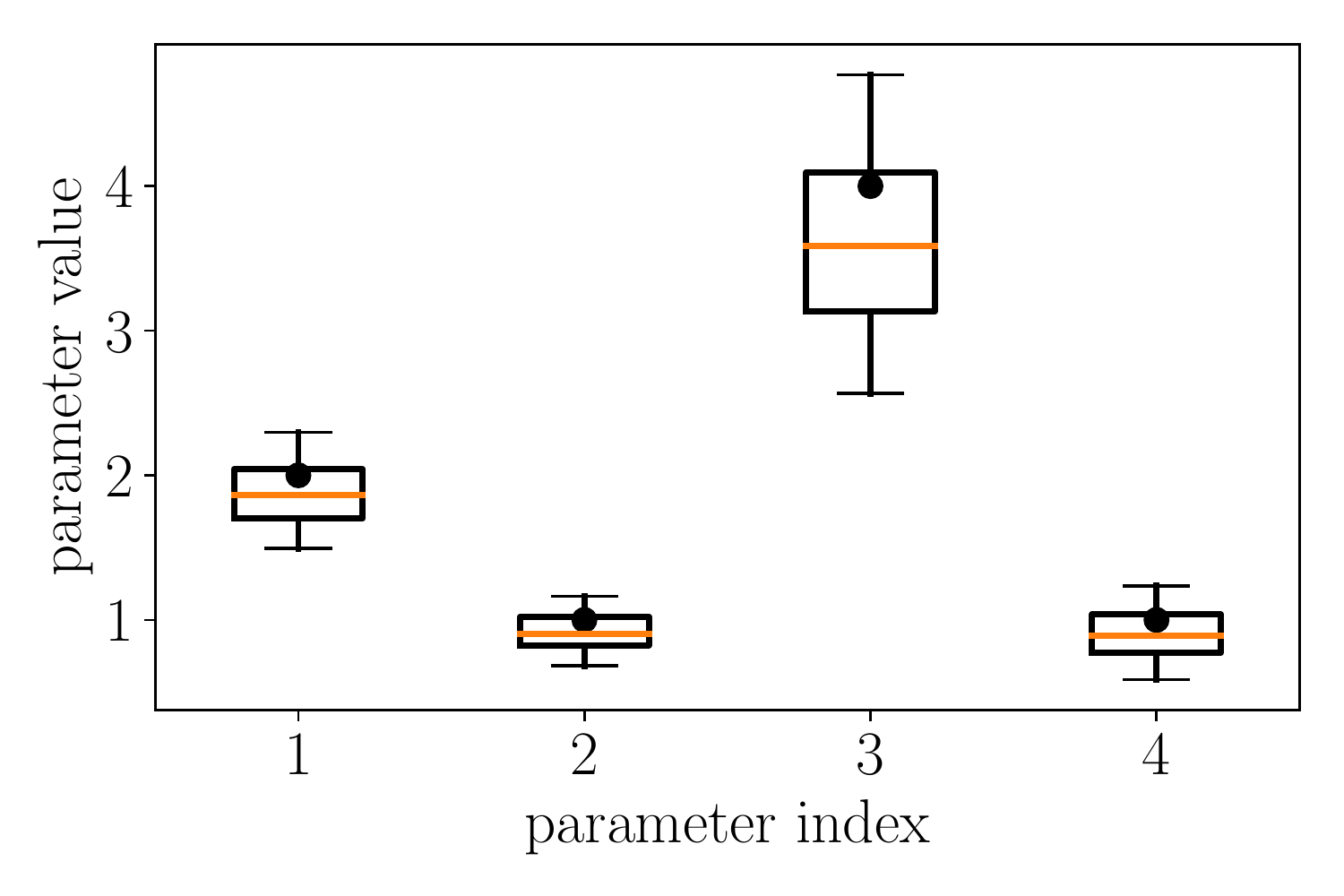}
        \caption{FGPGM}
    \end{subfigure}
    \begin{subfigure}[t]{.32\textwidth}
        \centering
        \includegraphics[width=\textwidth]{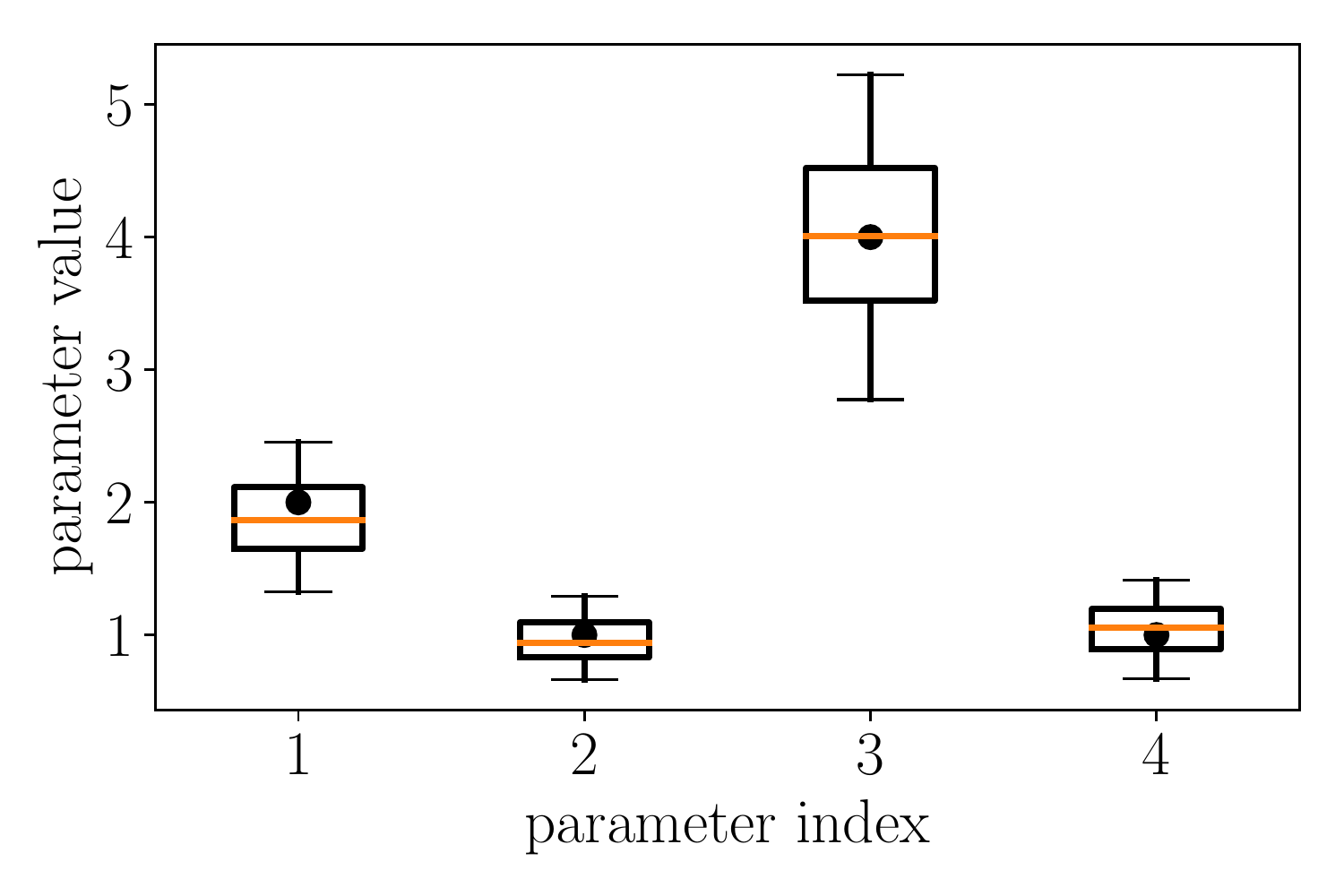}
        \caption{MVGM}
    \end{subfigure}
    \caption{Boxplots showing the inferred parameters over 100 runs for the Lotka Volterra dynamics for the high noise case. The black dots represent the ground truth, the orange line denotes the median of the 100 parameter estimates while the boxes and whiskers denote 50\% and 75\% quantiles.}
    \label{fig:LV0.5Noise}
\end{figure*}

\begin{figure*}[tbh!]
    \begin{tabular}[c]{ccc}
        \begin{subfigure}[t]{.32\textwidth}
            \centering
            \includegraphics[width=\textwidth]{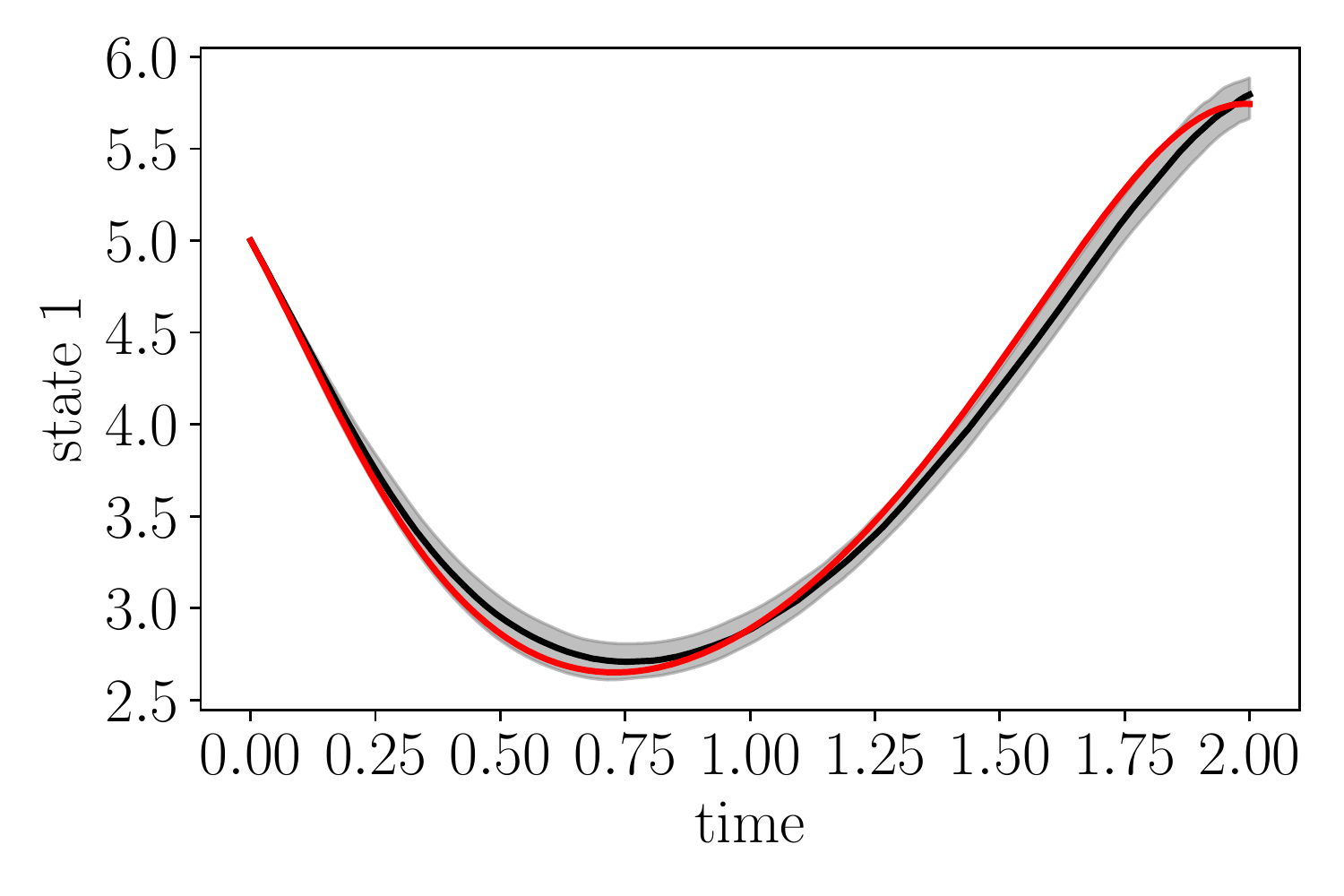}
            \caption{$\mathrm{AGM}$}
        \end{subfigure}&
        \begin{subfigure}[t]{.32\textwidth}
            \centering
            \includegraphics[width=\textwidth]{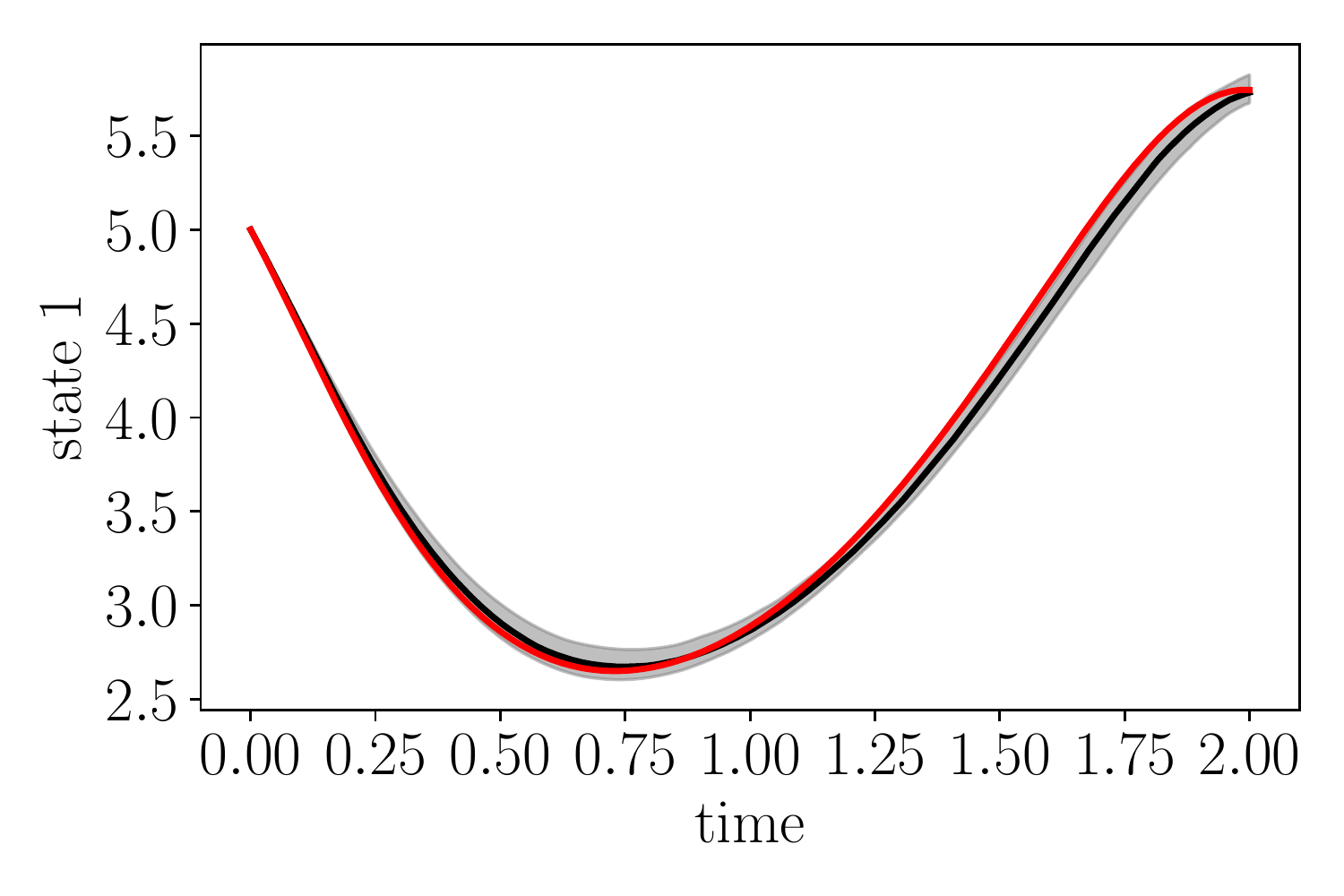}
            \caption{$\mathrm{VGM}$}
        \end{subfigure}&
        \begin{subfigure}[t]{.32\textwidth}
            \centering
            \includegraphics[width=\textwidth]{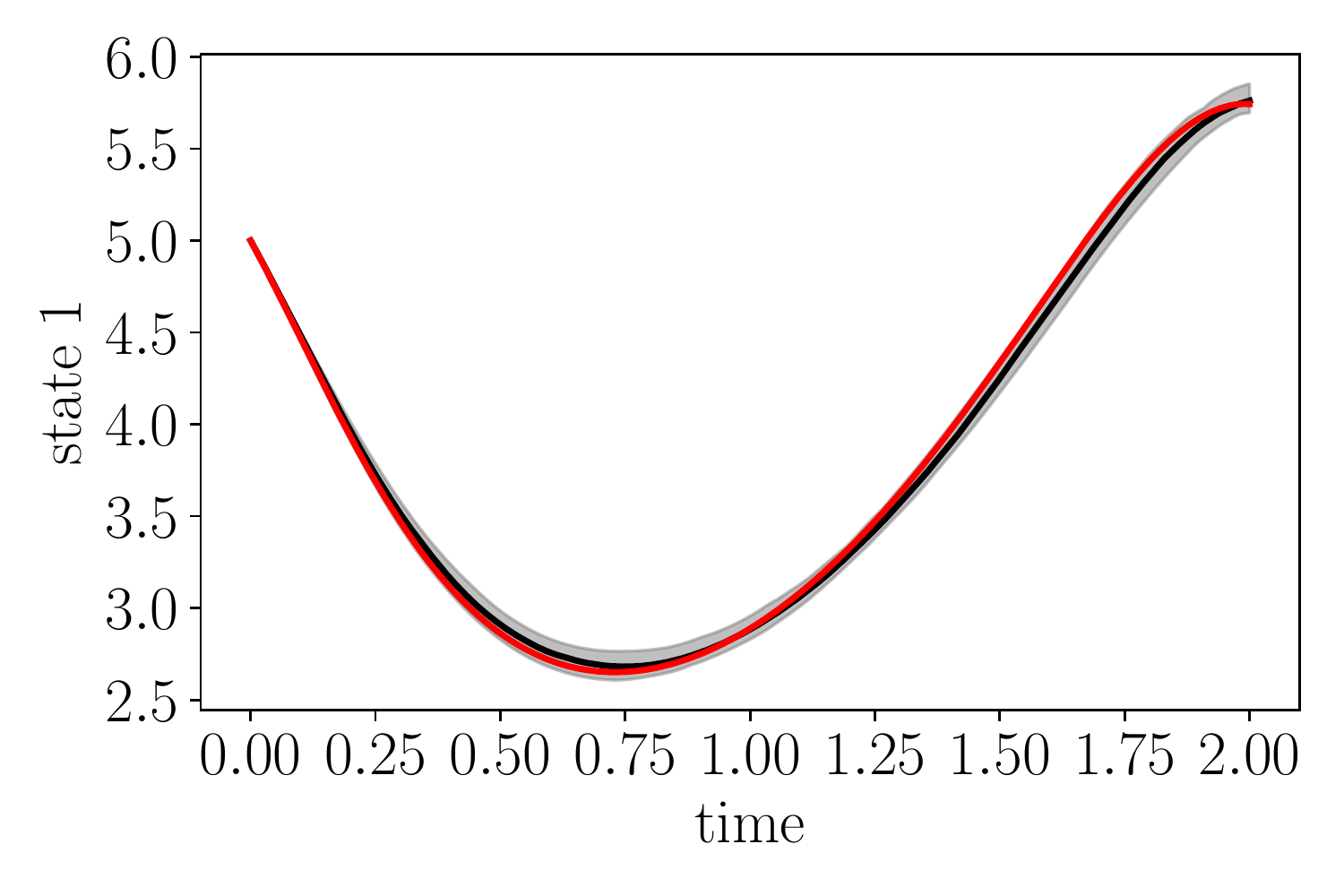}
            \caption{$\mathrm{FGPGM}$}
        \end{subfigure}\\
        \begin{subfigure}[t]{.32\textwidth}
            \centering
            \includegraphics[width=\textwidth]{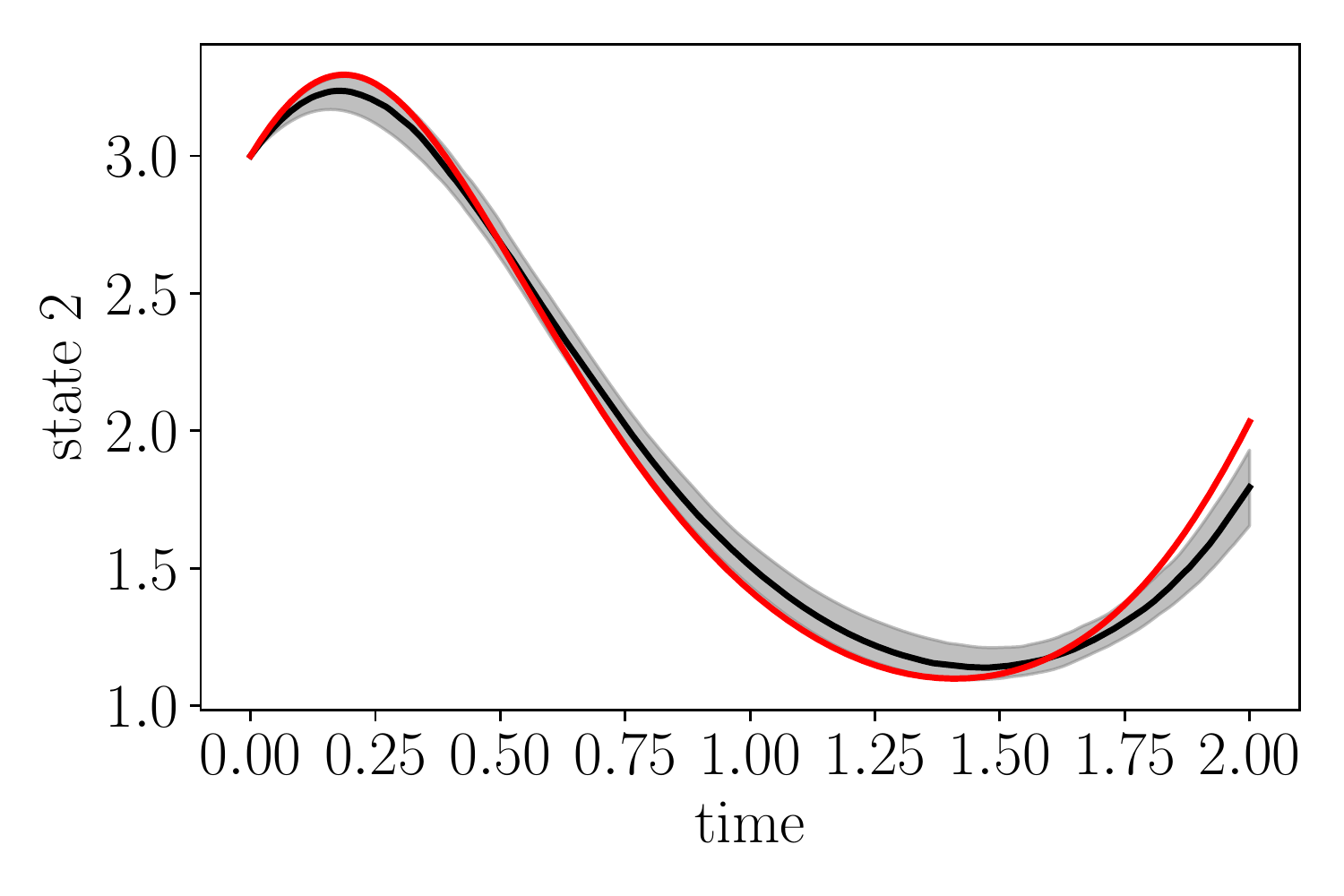}
        \end{subfigure}&
        \begin{subfigure}[t]{.32\textwidth}
            \centering
            \includegraphics[width=\textwidth]{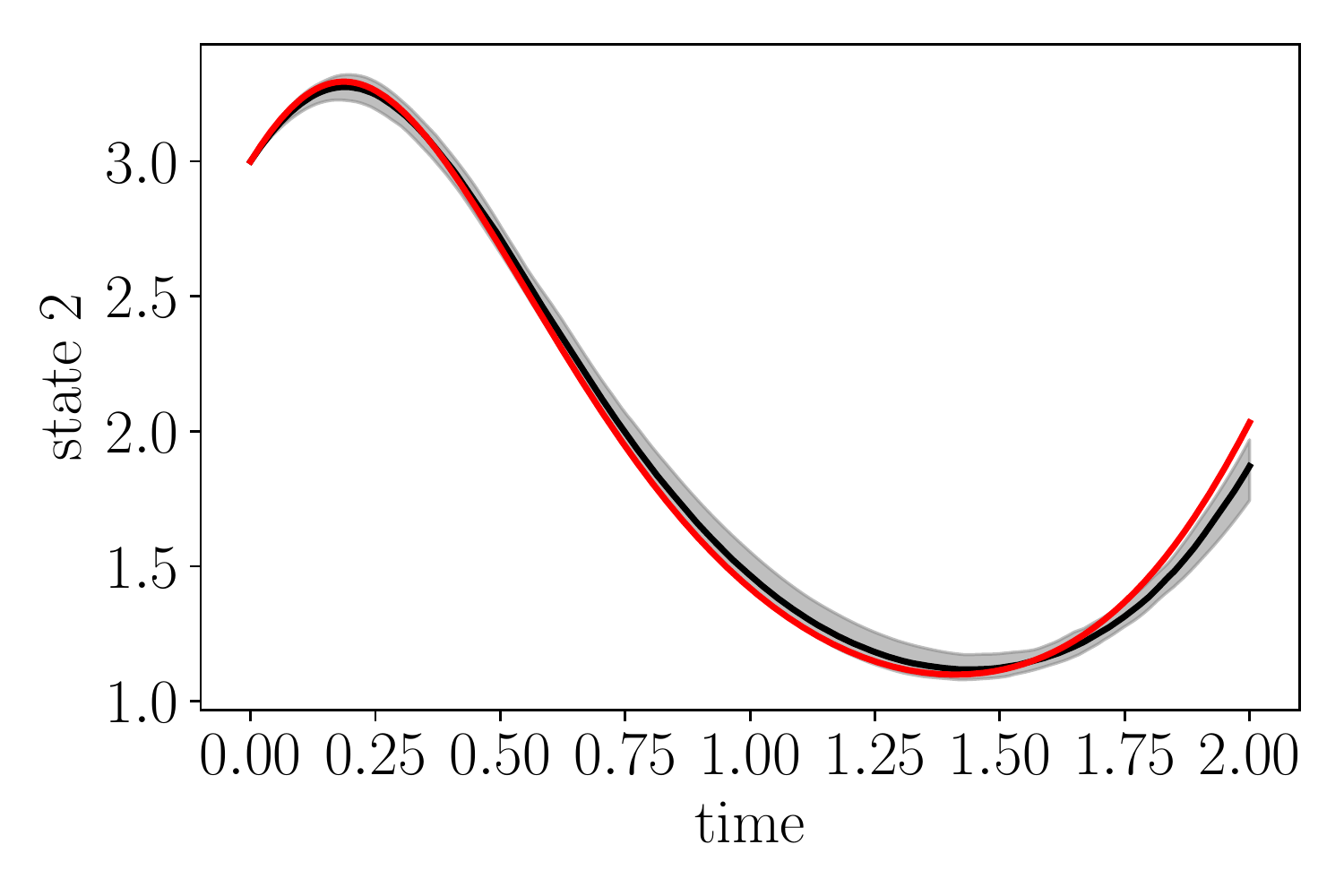}
        \end{subfigure}&
        \begin{subfigure}[t]{.32\textwidth}
            \centering
            \includegraphics[width=\textwidth]{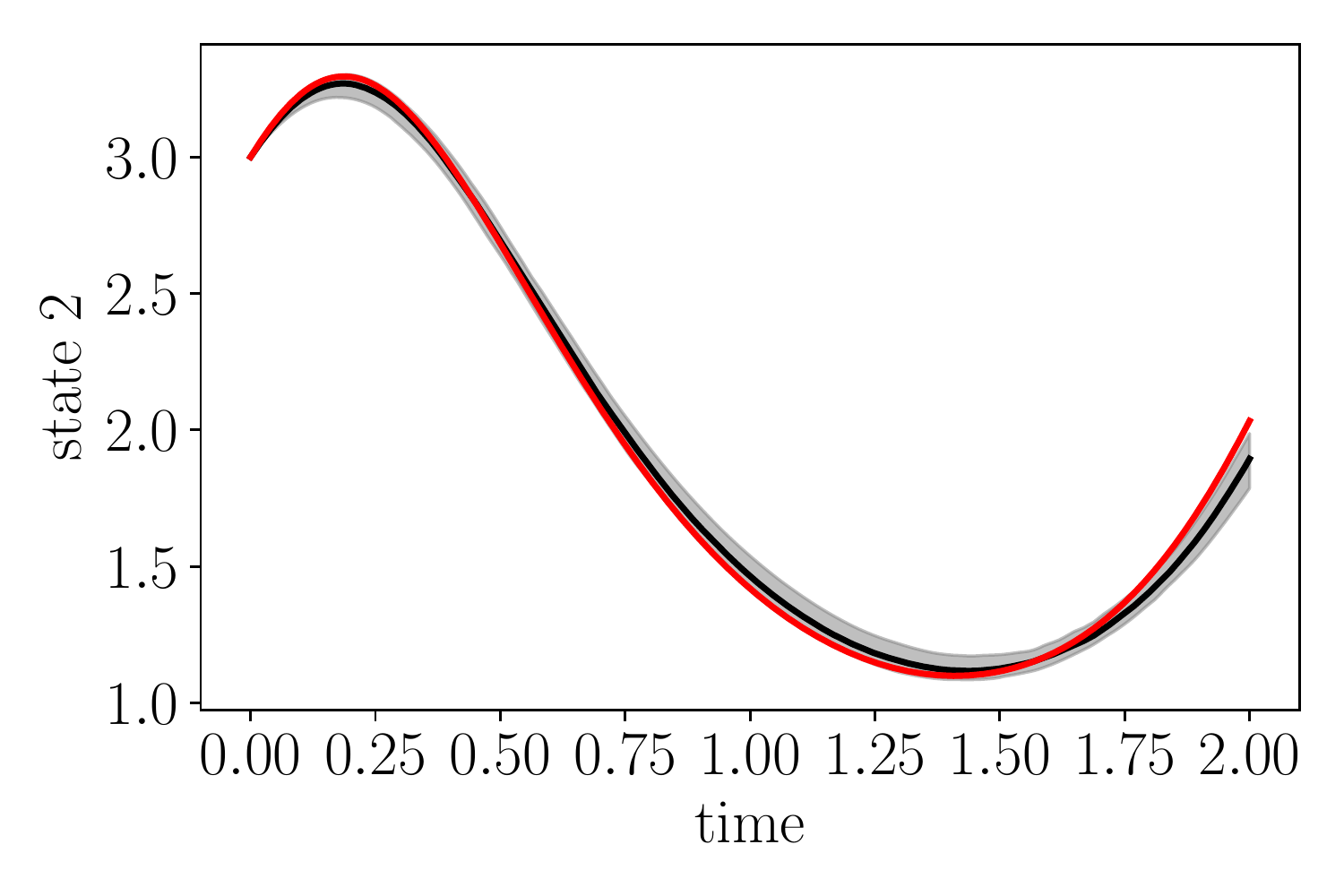}
        \end{subfigure}
    \end{tabular}
    \caption{Median Plots for all states of Lotka Volterra with low noise. State 1 is in the top row, state 2 is in the bottom row. The red line is the ground truth, while the black line and the shaded area denote the median and the 75\% quantiles of the results of 100 independent noise realizations. As was already to be expected by the parameter estimates, $\mathrm{FGPGM}$ and $\mathrm{VGM}$ are almost indistinguishable while $\mathrm{AGM}$ falls off a little bit.}
    \label{fig:LVowNoiseMedian}
\end{figure*}

\subsection{Parameter Distribution}
\label{sec:AppendixParamDist}
\begin{figure*}[tbh!]
    \centering
    \begin{subfigure}[t]{.24\textwidth}
        \centering
        \includegraphics[width=\textwidth]{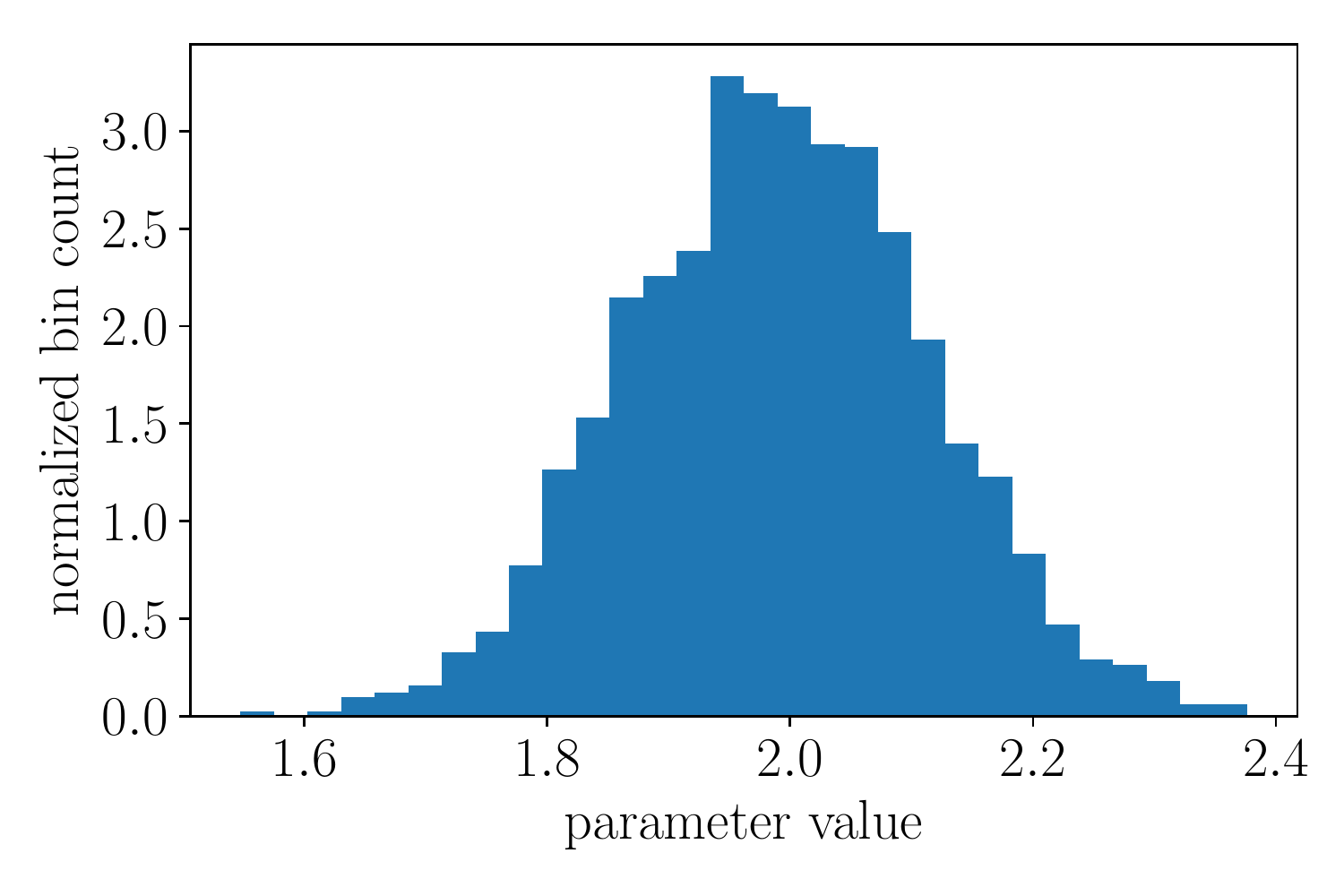}
    \end{subfigure}
    \begin{subfigure}[t]{.24\textwidth}
        \centering
        \includegraphics[width=\textwidth]{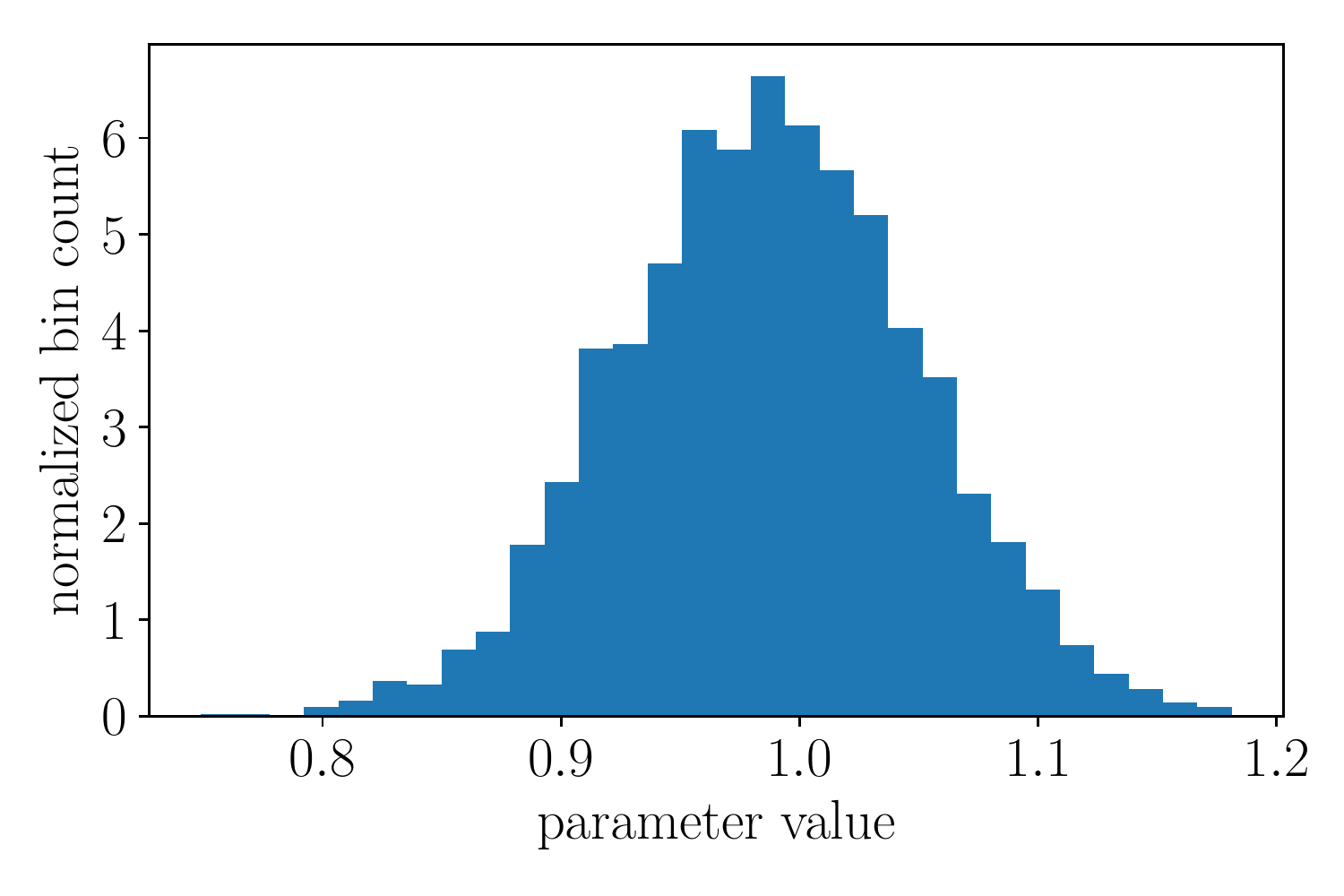}
    \end{subfigure}
    \begin{subfigure}[t]{.24\textwidth}
        \centering
        \includegraphics[width=\textwidth]{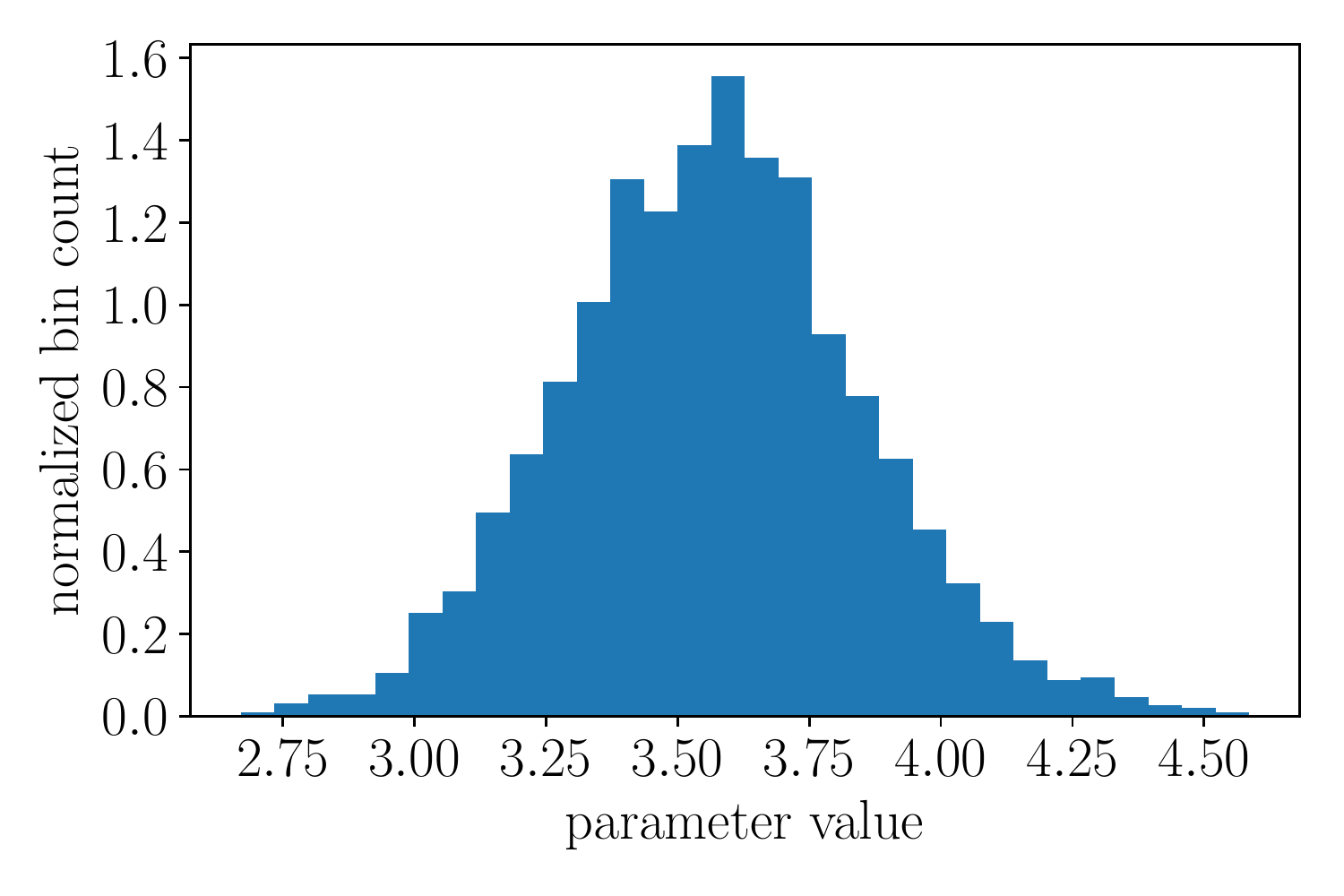}
    \end{subfigure}
    \begin{subfigure}[t]{.24\textwidth}
        \centering
        \includegraphics[width=\textwidth]{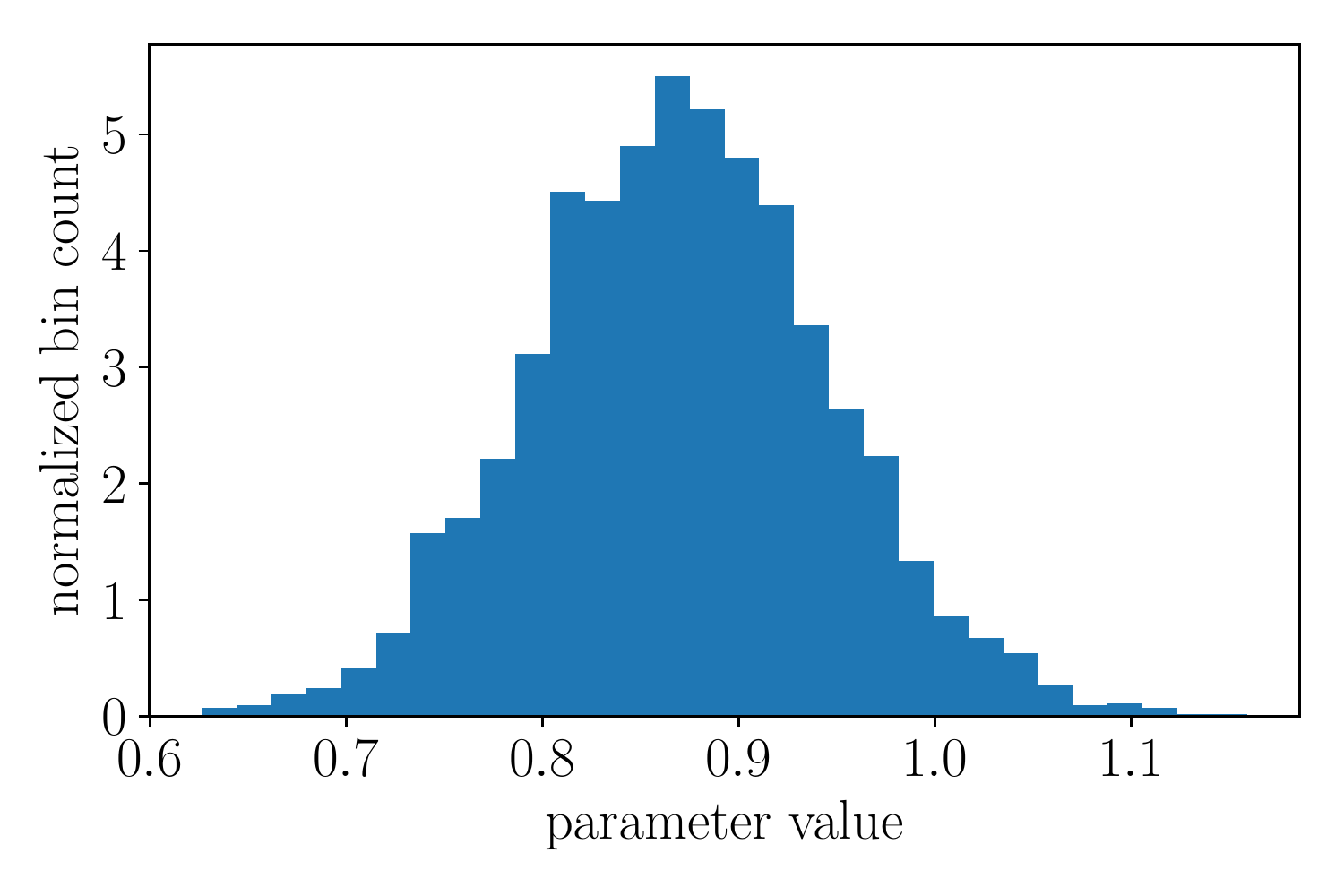}
    \end{subfigure}
    \caption{Histograms representing the MCMC samples obtained for one example run of the Lotka Volterra system. Each histogram represents the marginal distribution of one ODE parameter.}
    \label{fig:LV0.1Histograms}
\end{figure*}
The MCMC approach of $\mathrm{FGPGM}$ allows to infer the probability distribution over parameters. This is shown for one example rollout in Figure \ref{fig:LV0.1Histograms}. The inferred distributions are close to Gaussian in shape. This likely explains the sampling-like performance of the variational approach $\mathrm{MVGM}$, as their assumptions of using a factorized Gaussian proxy distribution over the parameters seems to be a good fit for the true distribution.

\subsection{Protein Transduction}

\begin{figure*}[tbh!]
    \begin{tabular}[c]{ccccc}
        \begin{subfigure}[t]{.17\textwidth}
            \centering
            \includegraphics[width=\textwidth]{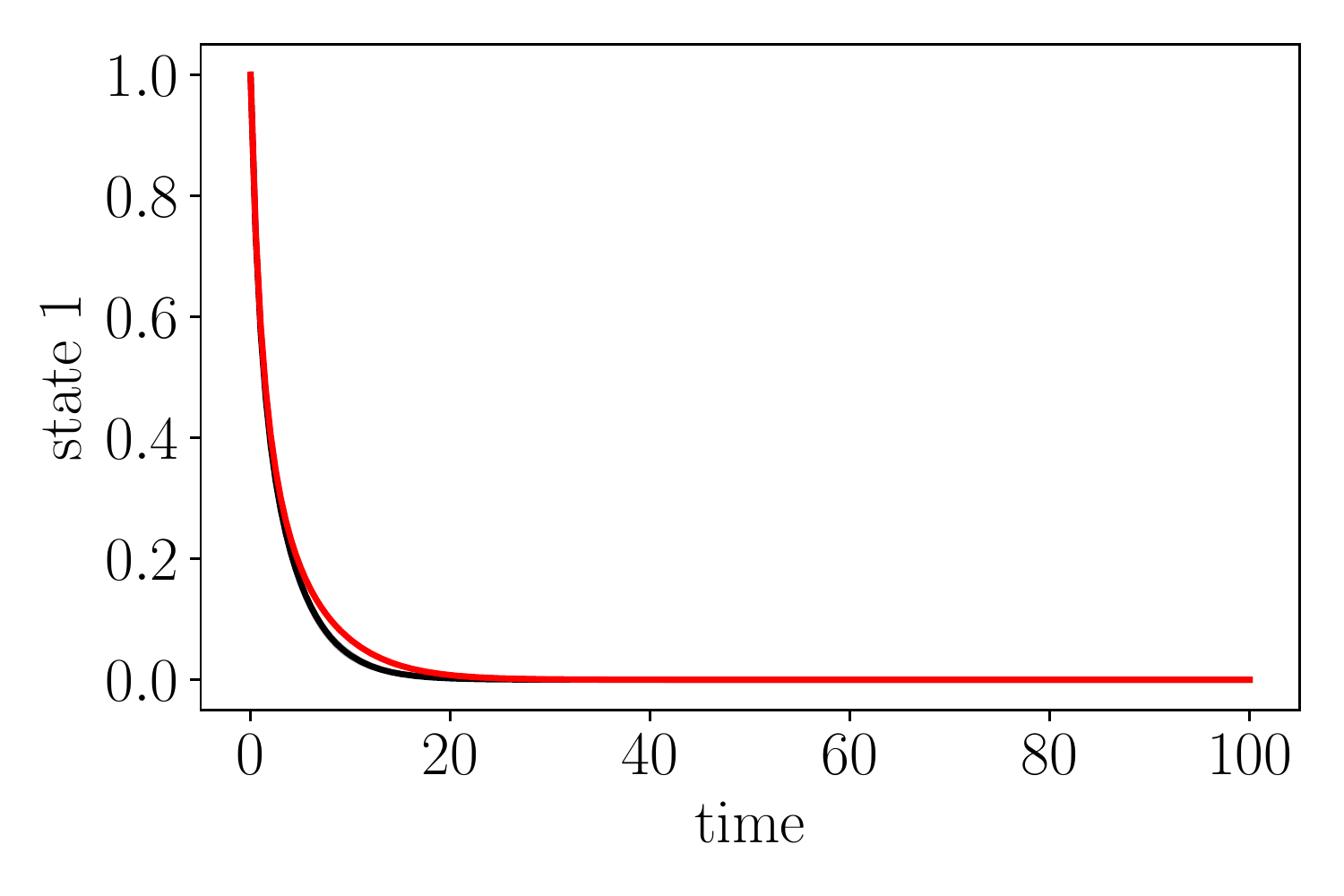}
        \end{subfigure}&
        \begin{subfigure}[t]{.17\textwidth}
            \centering
            \includegraphics[width=\textwidth]{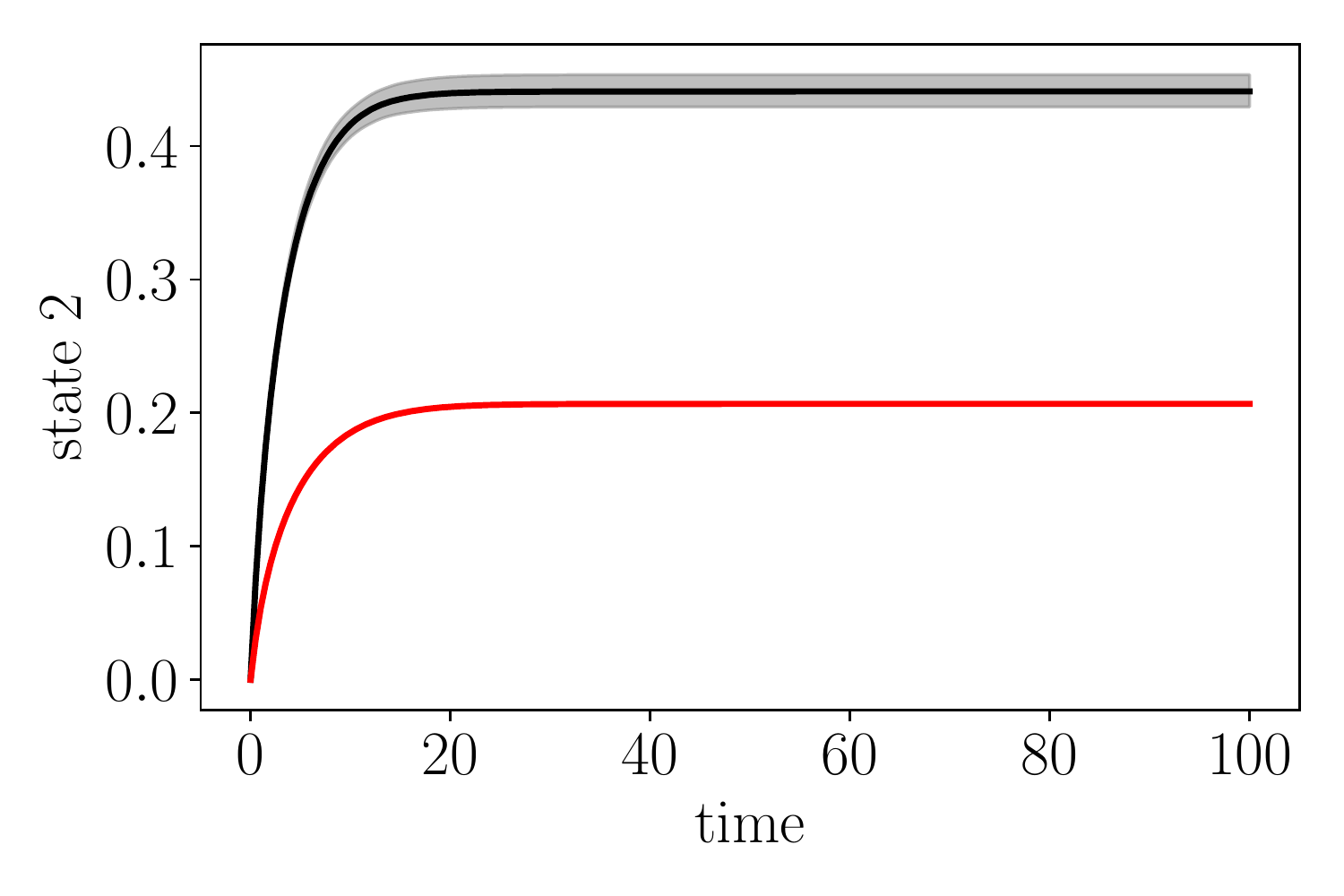}
        \end{subfigure}&
        \begin{subfigure}[t]{.17\textwidth}
            \centering
            \includegraphics[width=\textwidth]{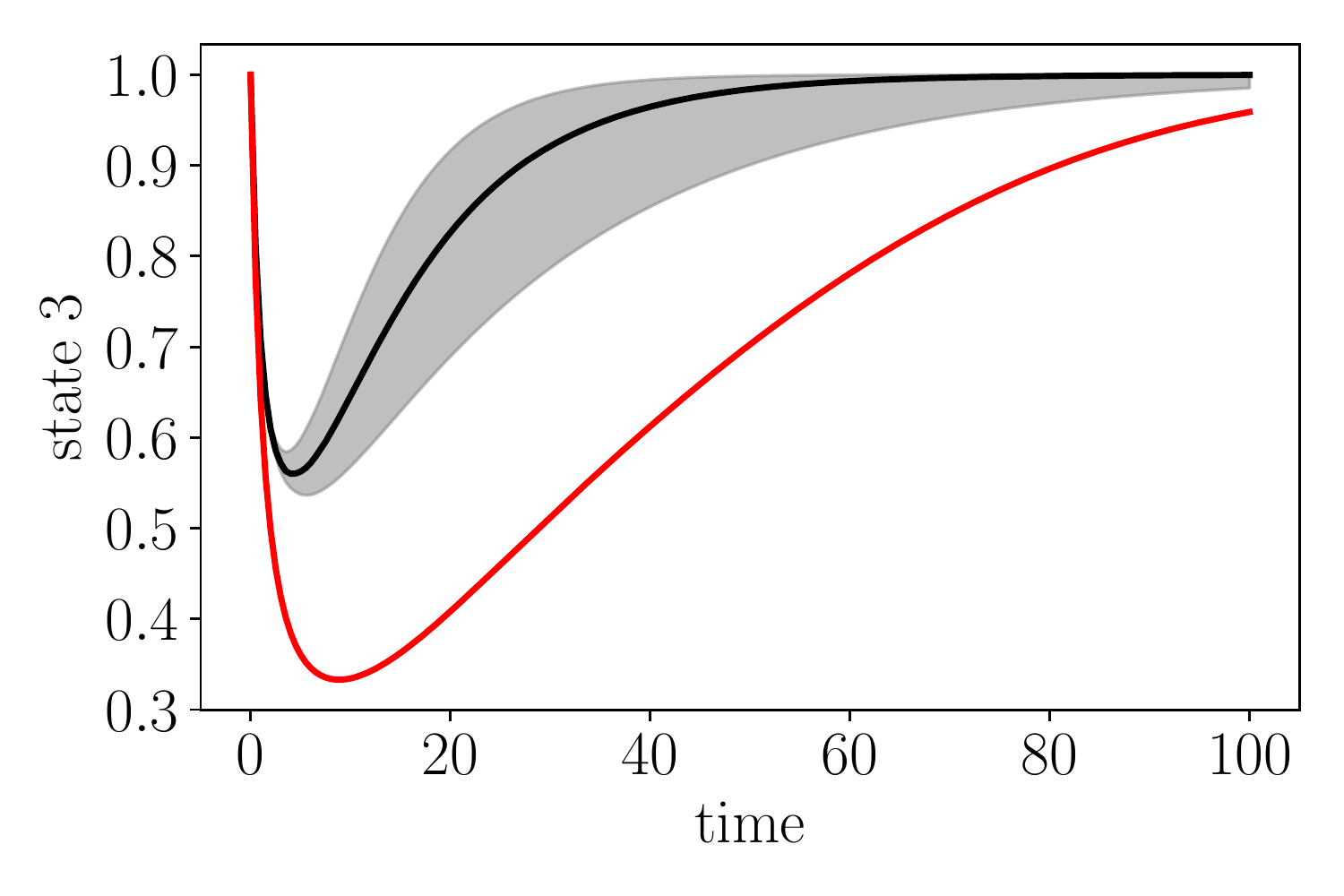}
        \end{subfigure}&
        \begin{subfigure}[t]{.17\textwidth}
            \centering
            \includegraphics[width=\textwidth]{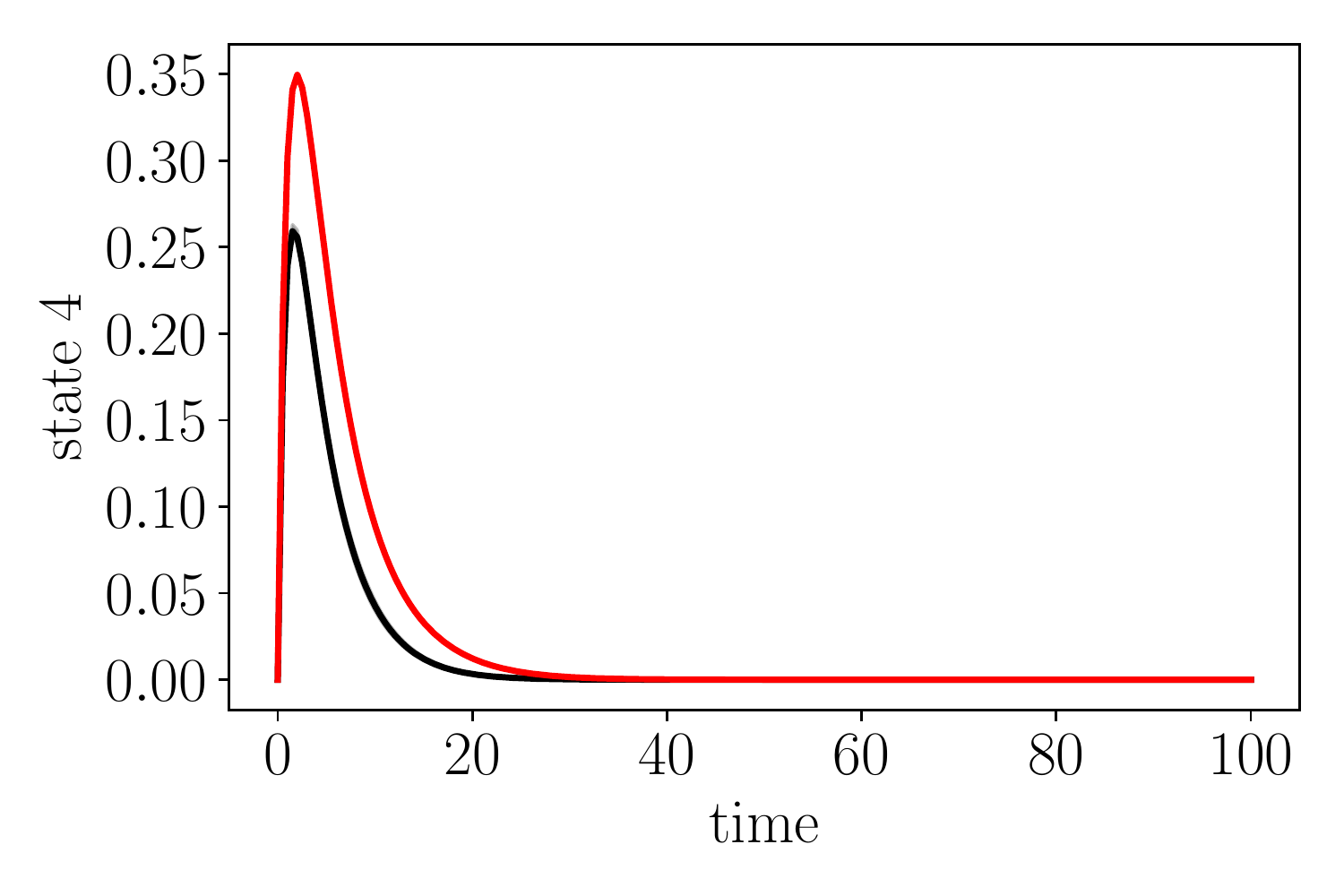}
        \end{subfigure}&
        \begin{subfigure}[t]{.17\textwidth}
            \centering
            \includegraphics[width=\textwidth]{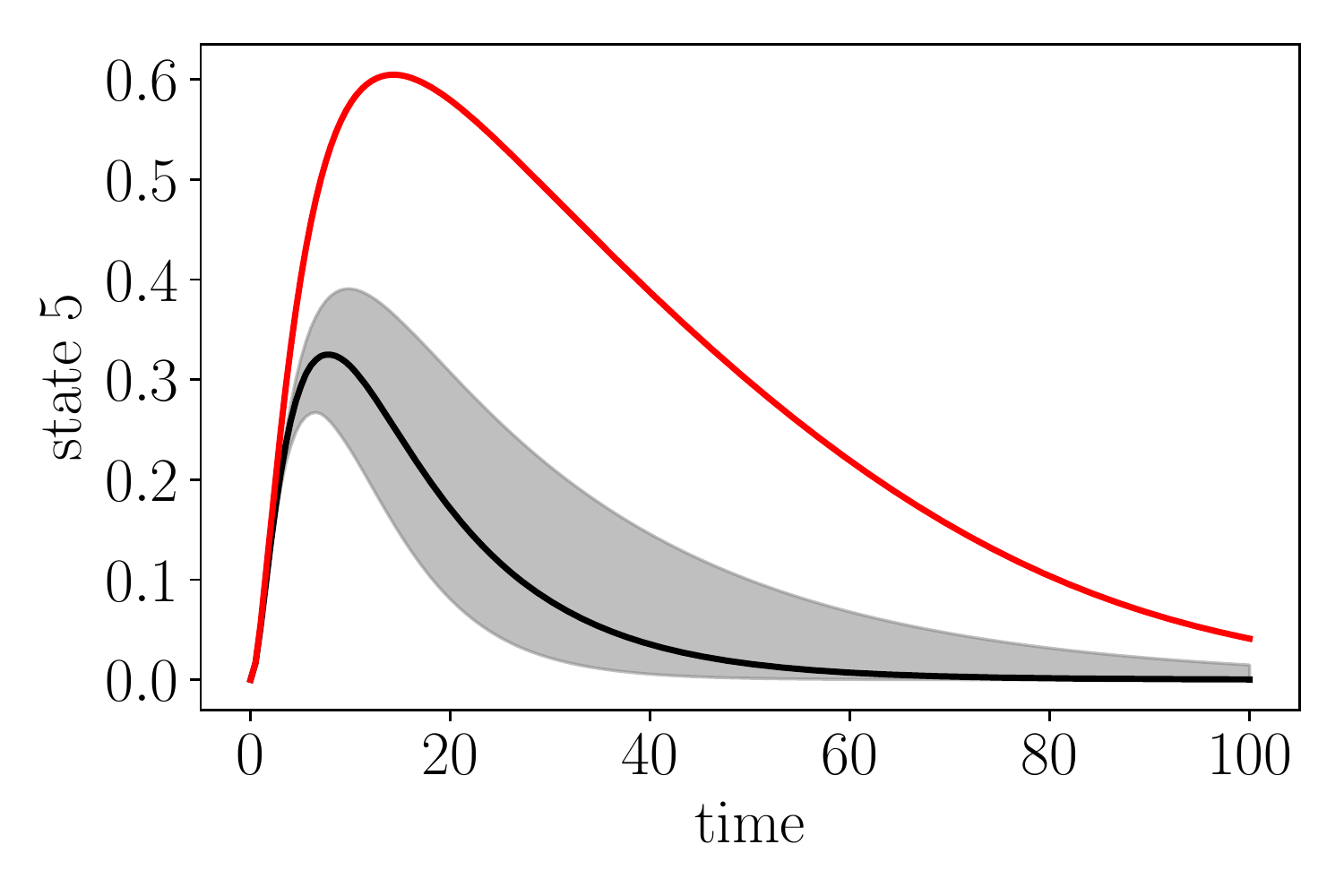}
        \end{subfigure}\\
        \begin{subfigure}[t]{.17\textwidth}
            \centering
            \includegraphics[width=\textwidth]{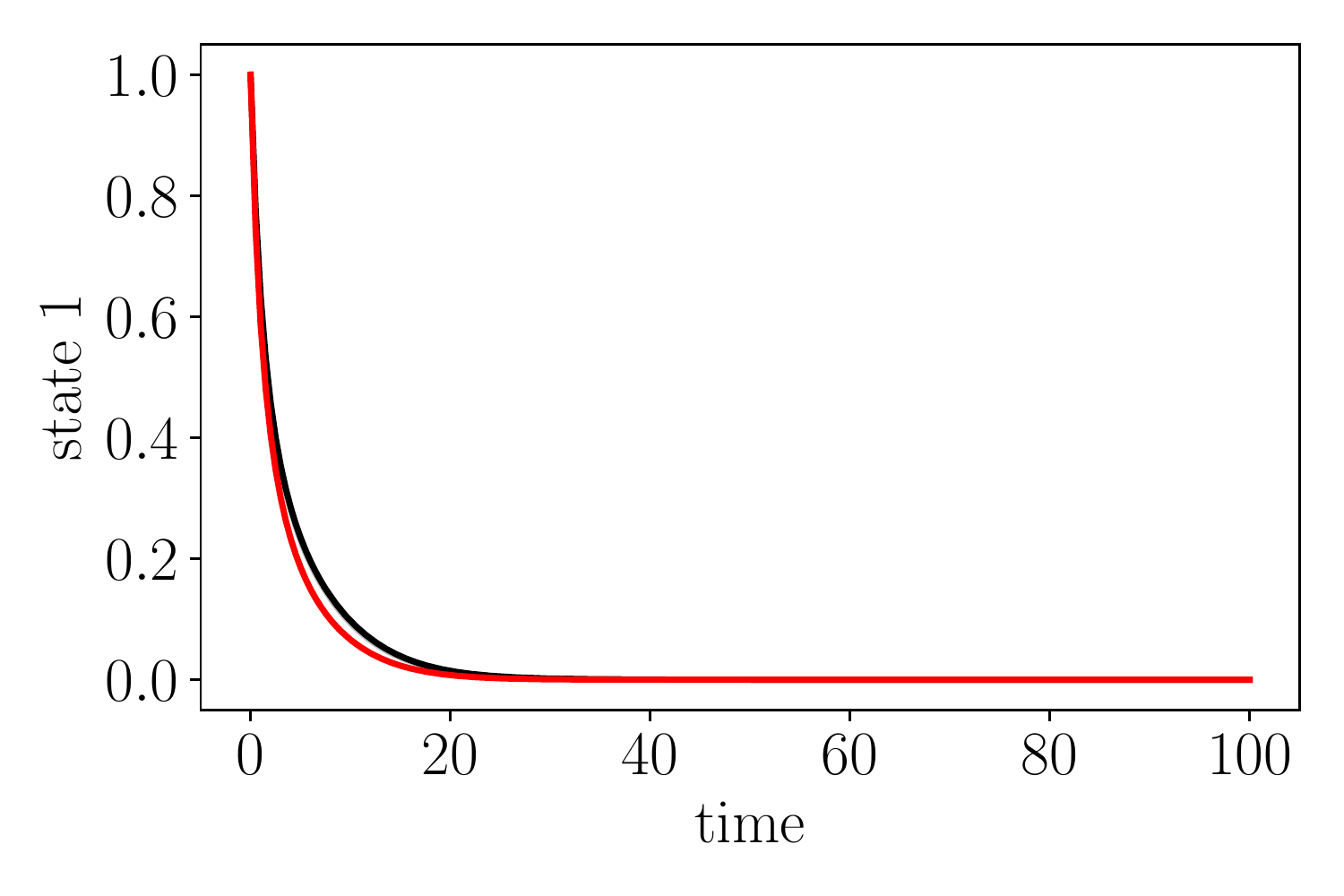}
        \end{subfigure}&
        \begin{subfigure}[t]{.17\textwidth}
            \centering
            \includegraphics[width=\textwidth]{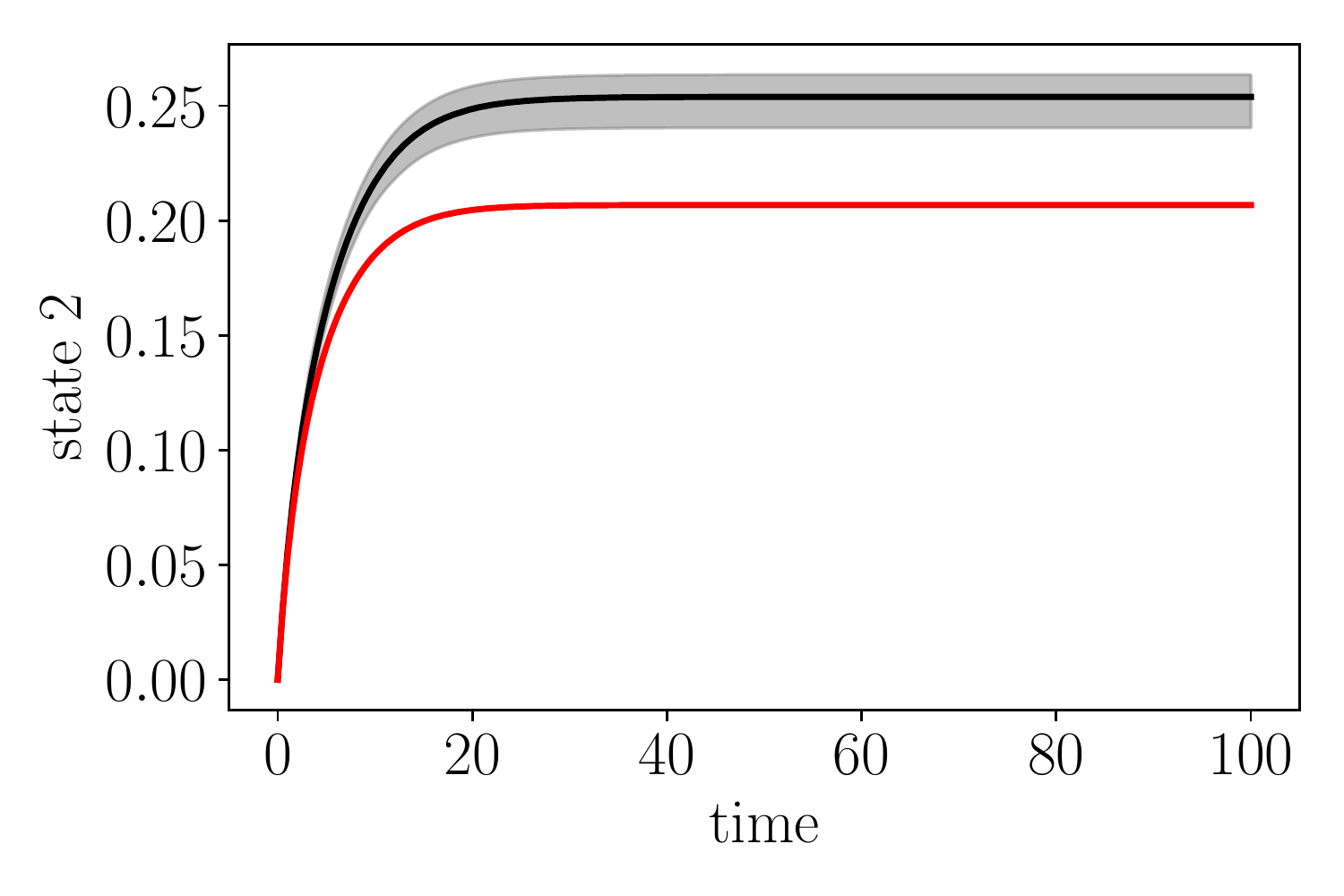}
        \end{subfigure}&
        \begin{subfigure}[t]{.17\textwidth}
            \centering
            \includegraphics[width=\textwidth]{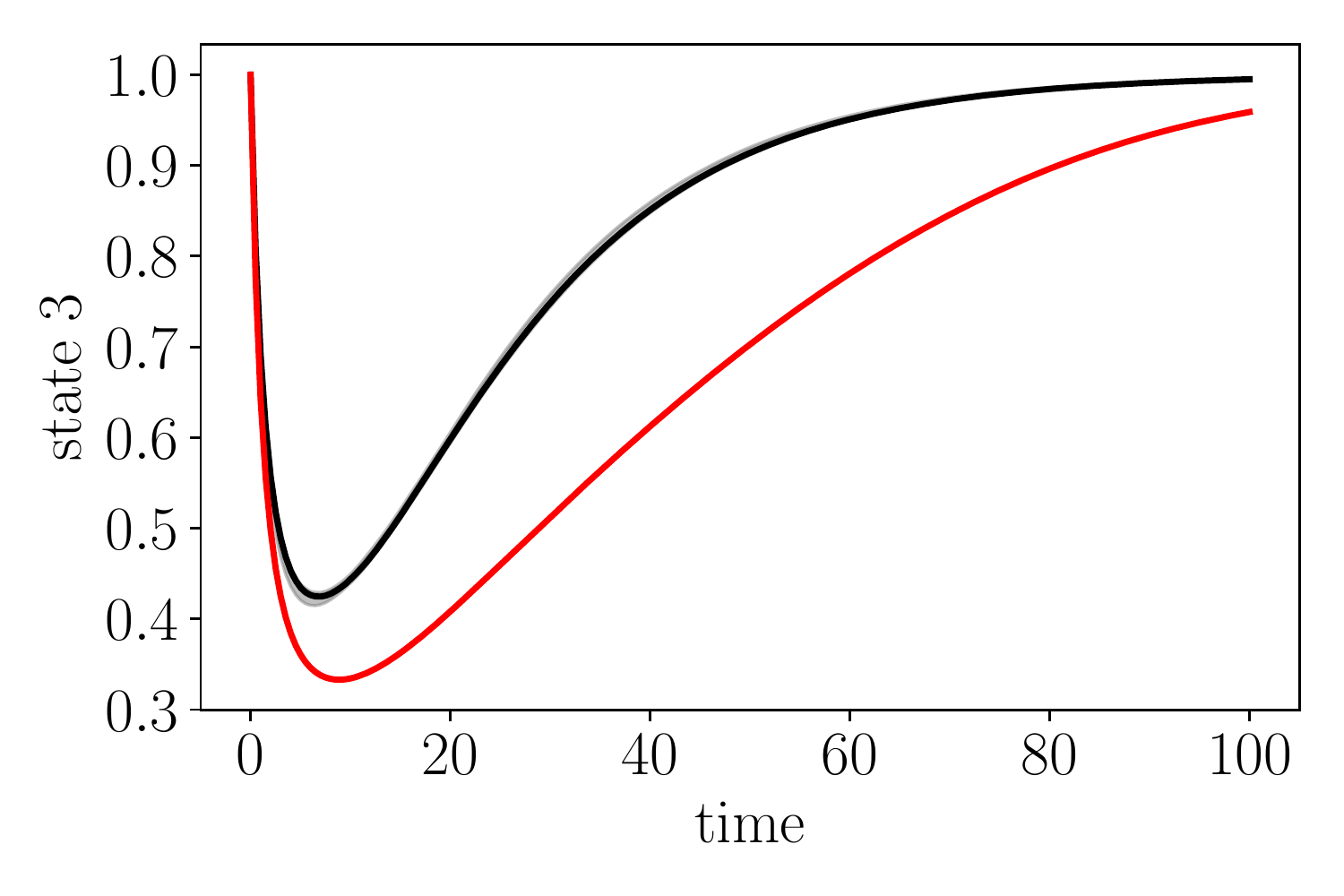}
        \end{subfigure}&
        \begin{subfigure}[t]{.17\textwidth}
            \centering
            \includegraphics[width=\textwidth]{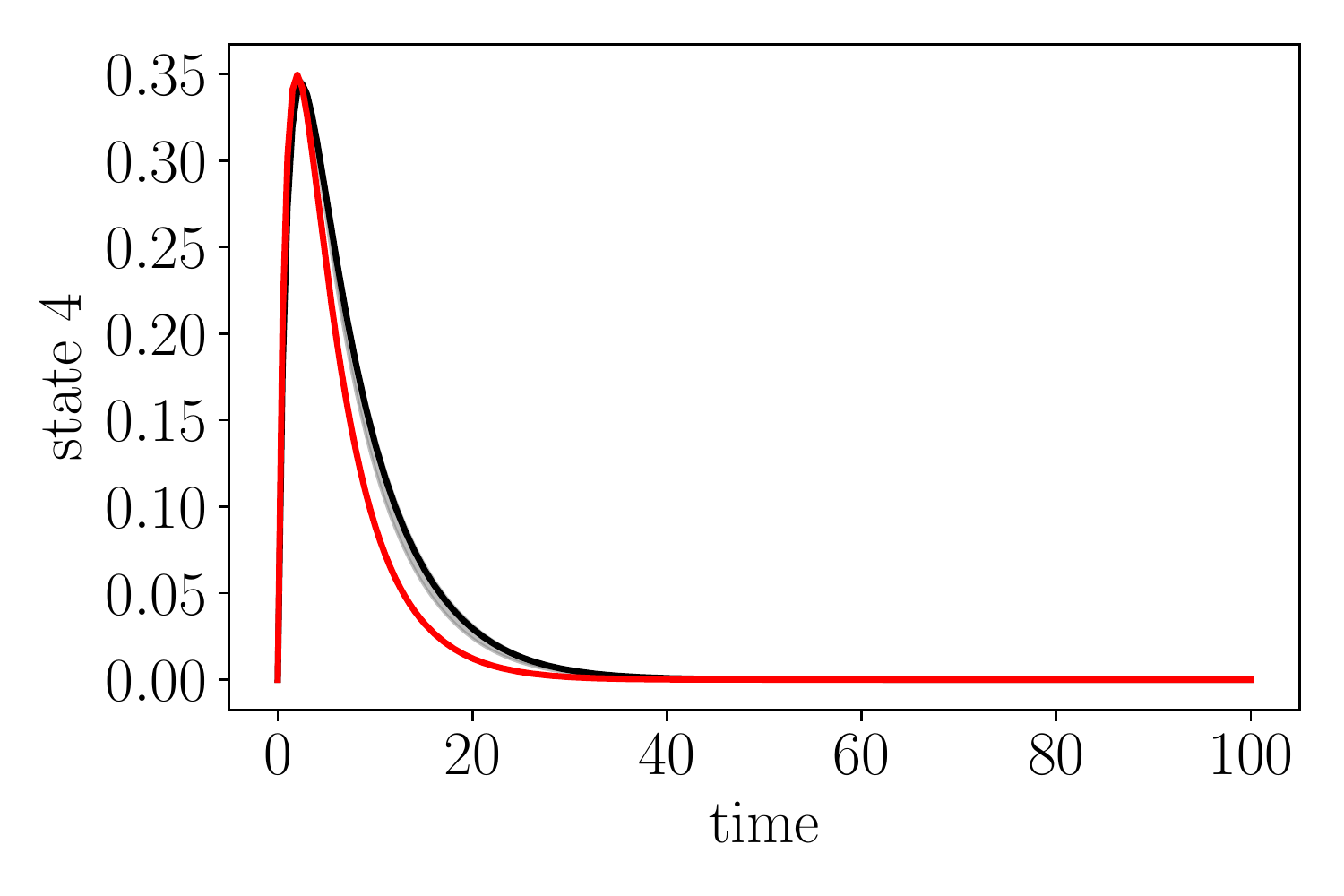}
        \end{subfigure}&
        \begin{subfigure}[t]{.17\textwidth}
            \centering
            \includegraphics[width=\textwidth]{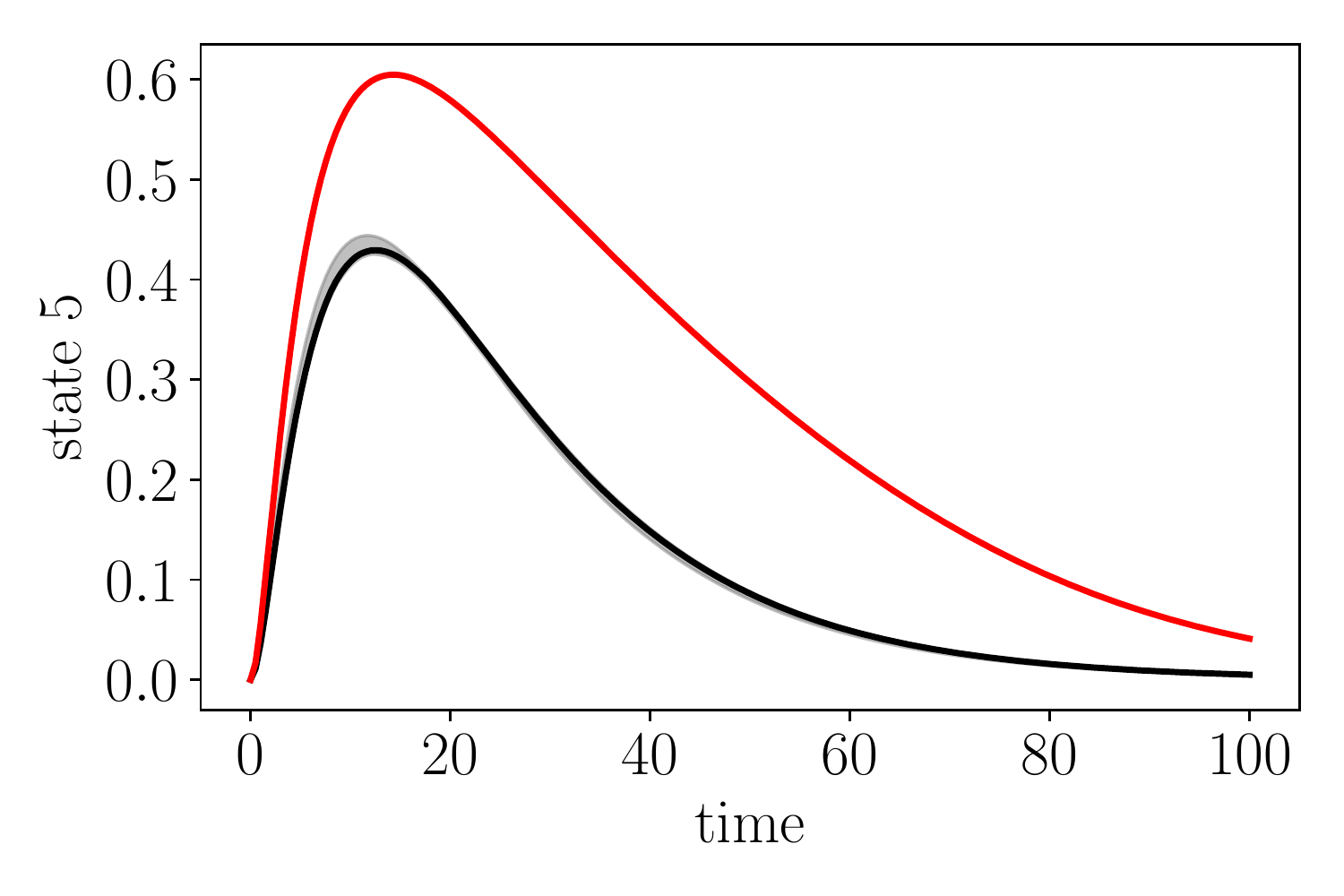}
        \end{subfigure}
    \end{tabular}
    \caption{Median plots for all states of the most difficult benchmark system in the literature, Protein Transduction. The red line is the ground truth, while the black line and the shaded area denote the median and the 75\% quantiles of the results of 100 independent noise realizations. $\mathrm{FGPGM}$ (middle) is clearly able to find more accurate parameter estimates than $\mathrm{AGM}$ (top).}
    \label{fig:PTLowNoise}
\end{figure*}

\begin{figure*}[!tbh]
    \begin{tabular}[c]{ccccc}
        \begin{subfigure}[t]{.17\textwidth}
            \centering
            \includegraphics[width=\textwidth]{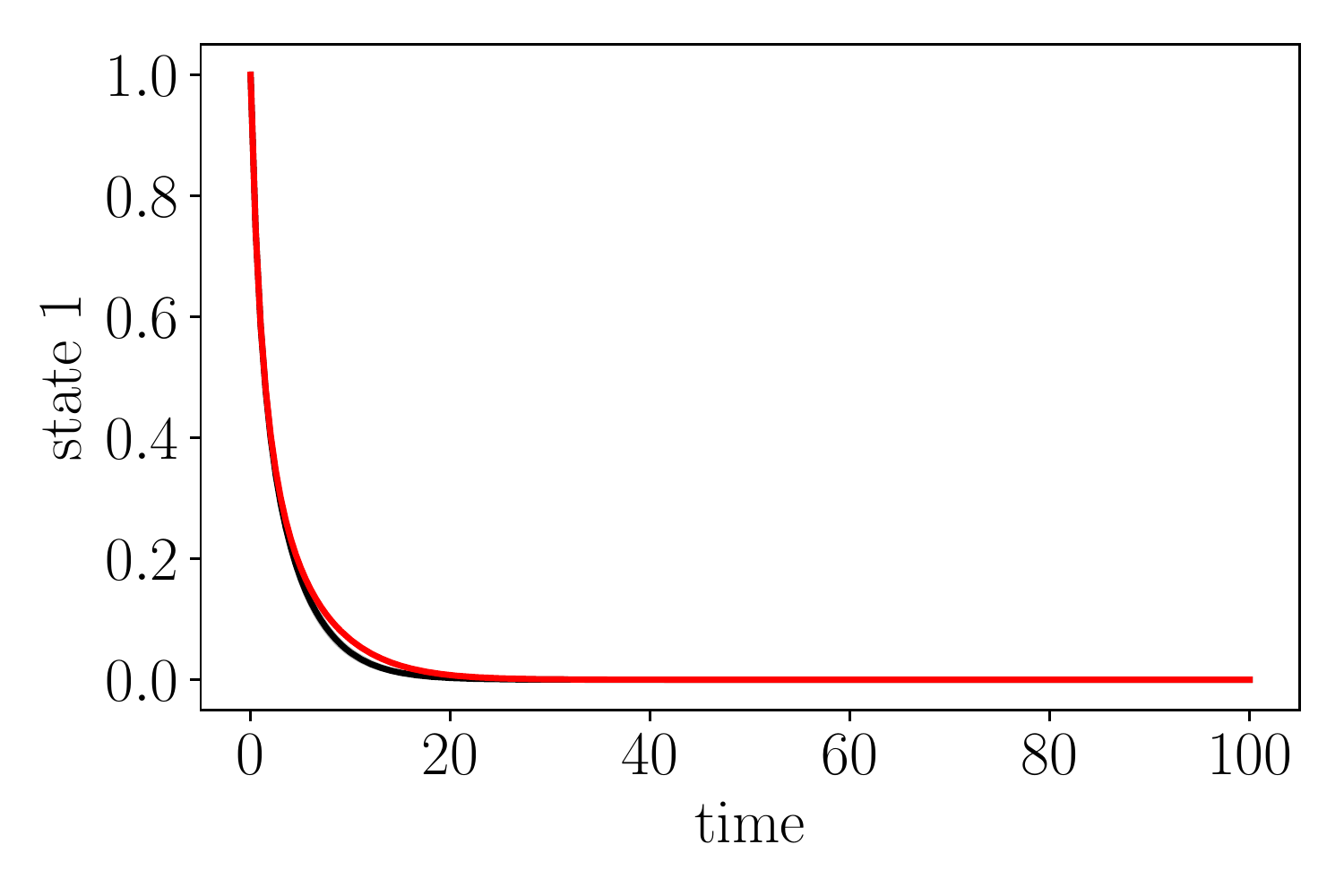}
        \end{subfigure}&
        \begin{subfigure}[t]{.17\textwidth}
            \centering
            \includegraphics[width=\textwidth]{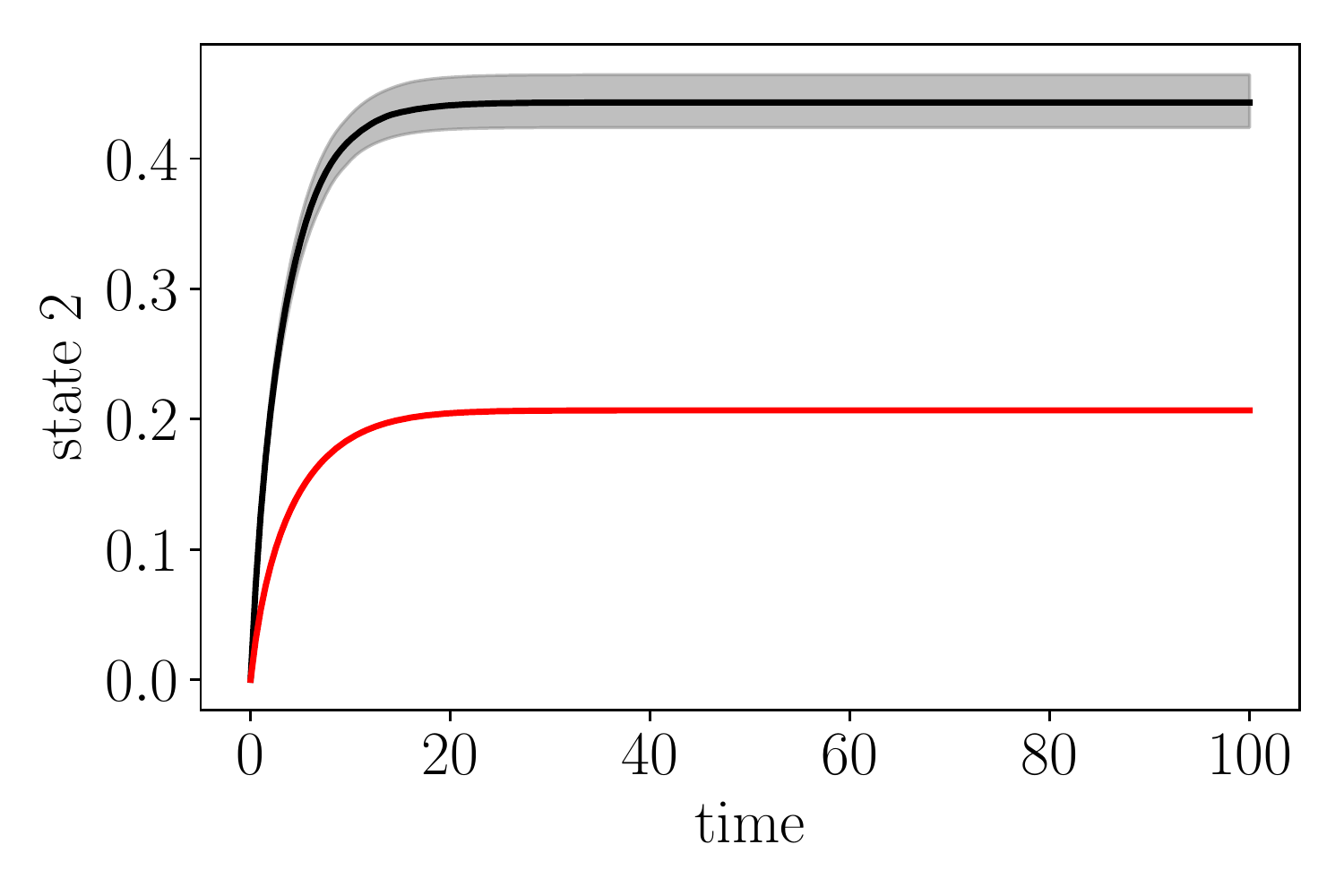}
        \end{subfigure}&
        \begin{subfigure}[t]{.17\textwidth}
            \centering
            \includegraphics[width=\textwidth]{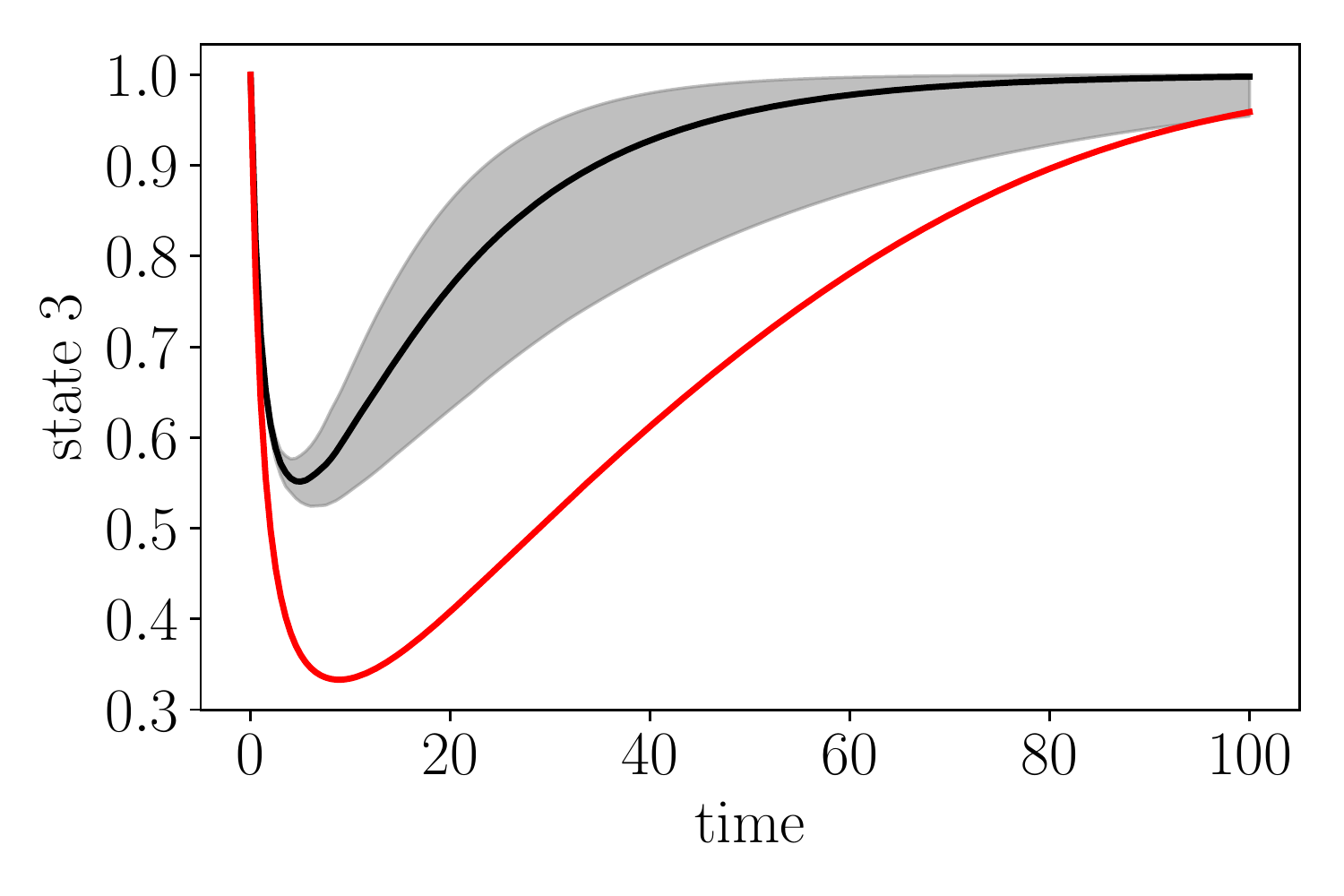}
        \end{subfigure}&
        \begin{subfigure}[t]{.17\textwidth}
            \centering
            \includegraphics[width=\textwidth]{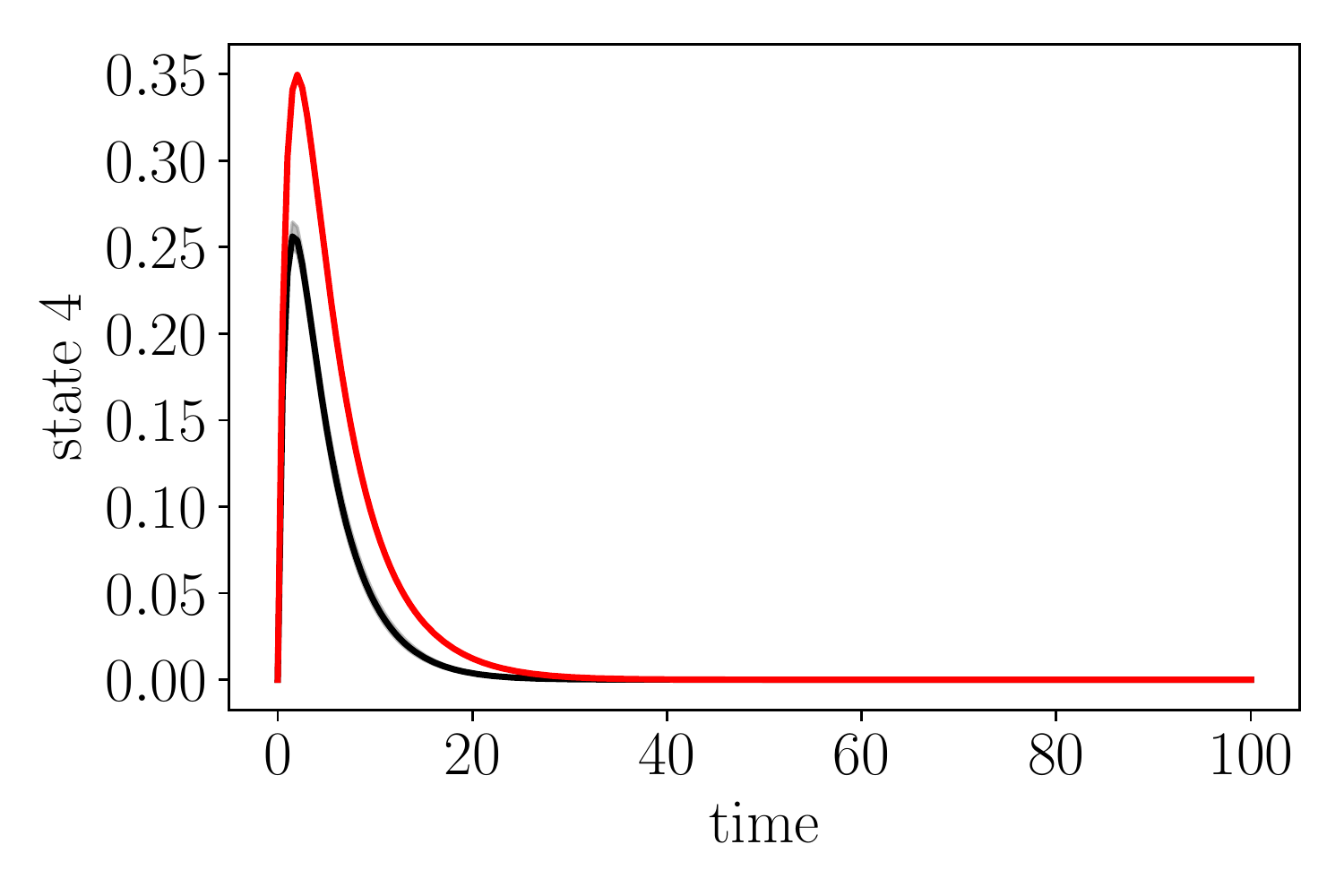}
        \end{subfigure}&
        \begin{subfigure}[t]{.17\textwidth}
            \centering
            \includegraphics[width=\textwidth]{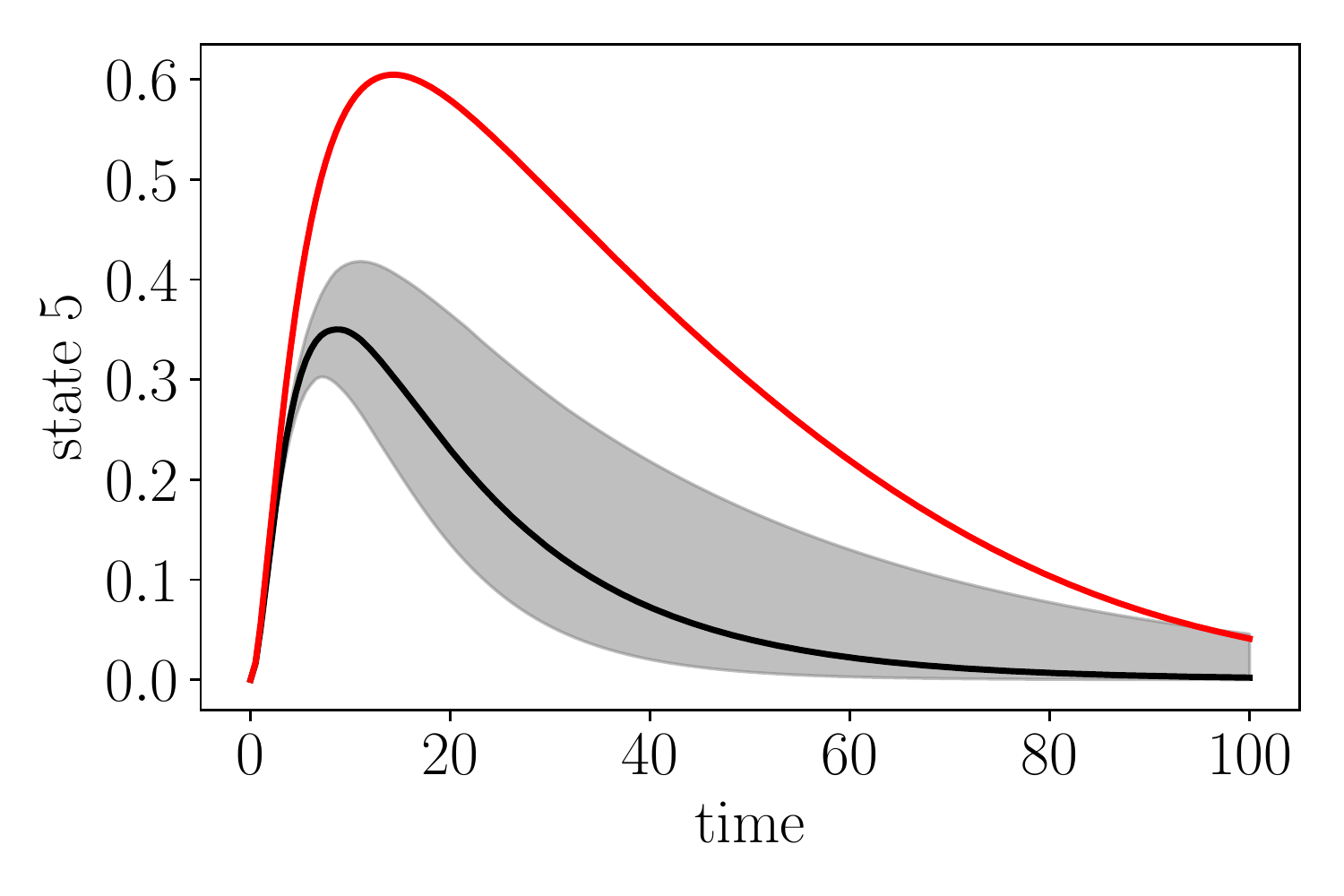}
        \end{subfigure}\\
        \begin{subfigure}[t]{.17\textwidth}
            \centering
            \includegraphics[width=\textwidth]{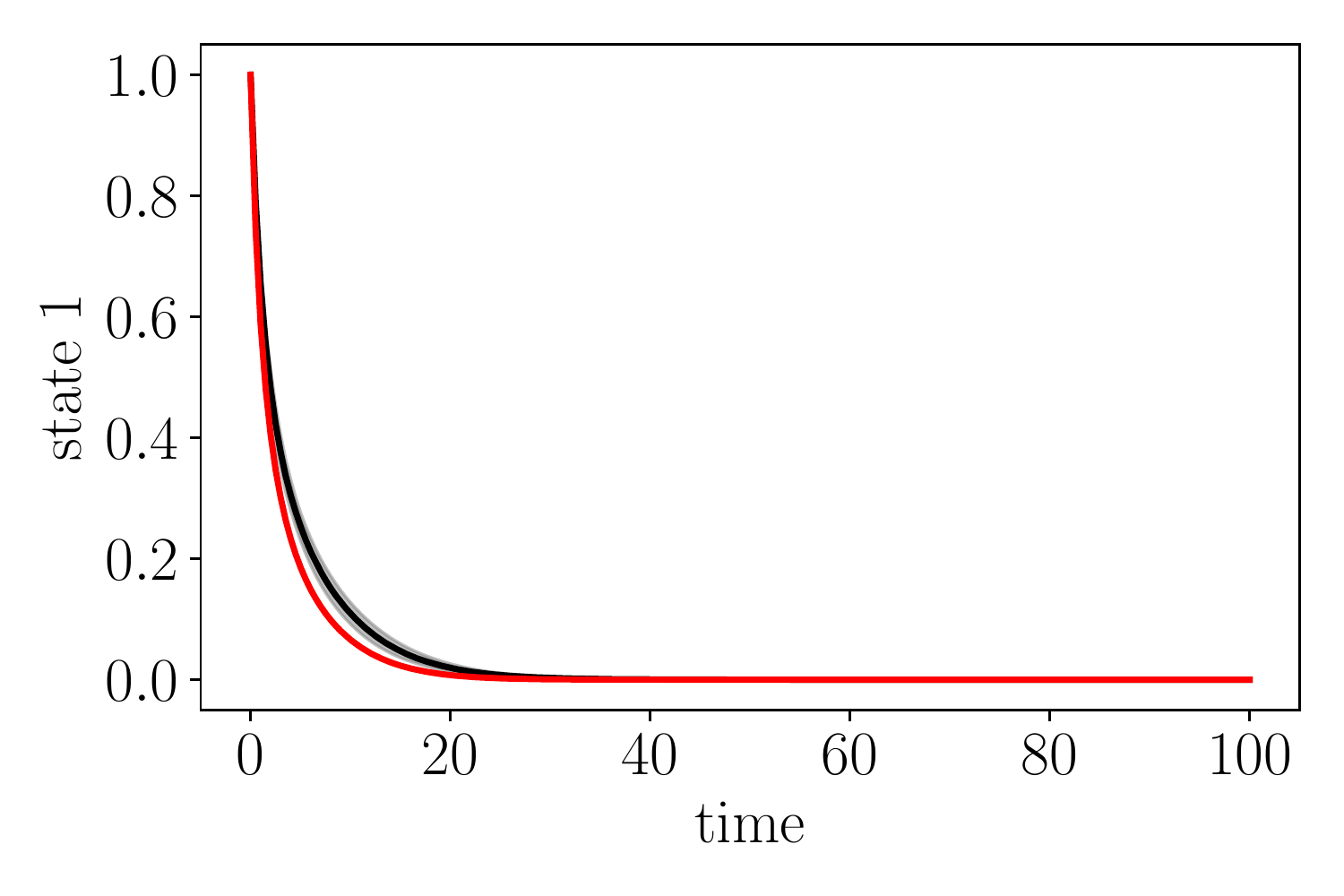}
        \end{subfigure}&
        \begin{subfigure}[t]{.17\textwidth}
            \centering
            \includegraphics[width=\textwidth]{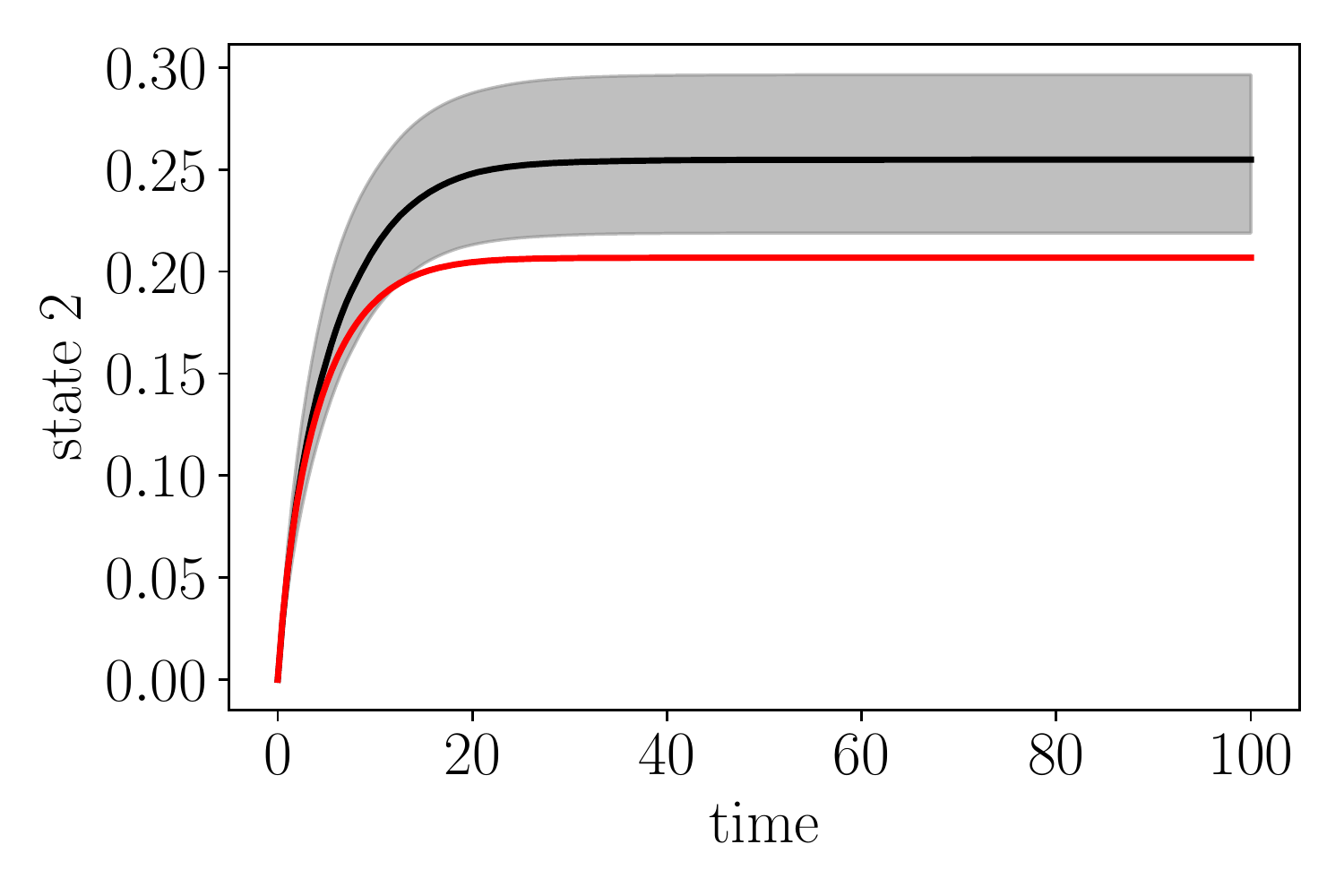}
        \end{subfigure}&
        \begin{subfigure}[t]{.17\textwidth}
            \centering
            \includegraphics[width=\textwidth]{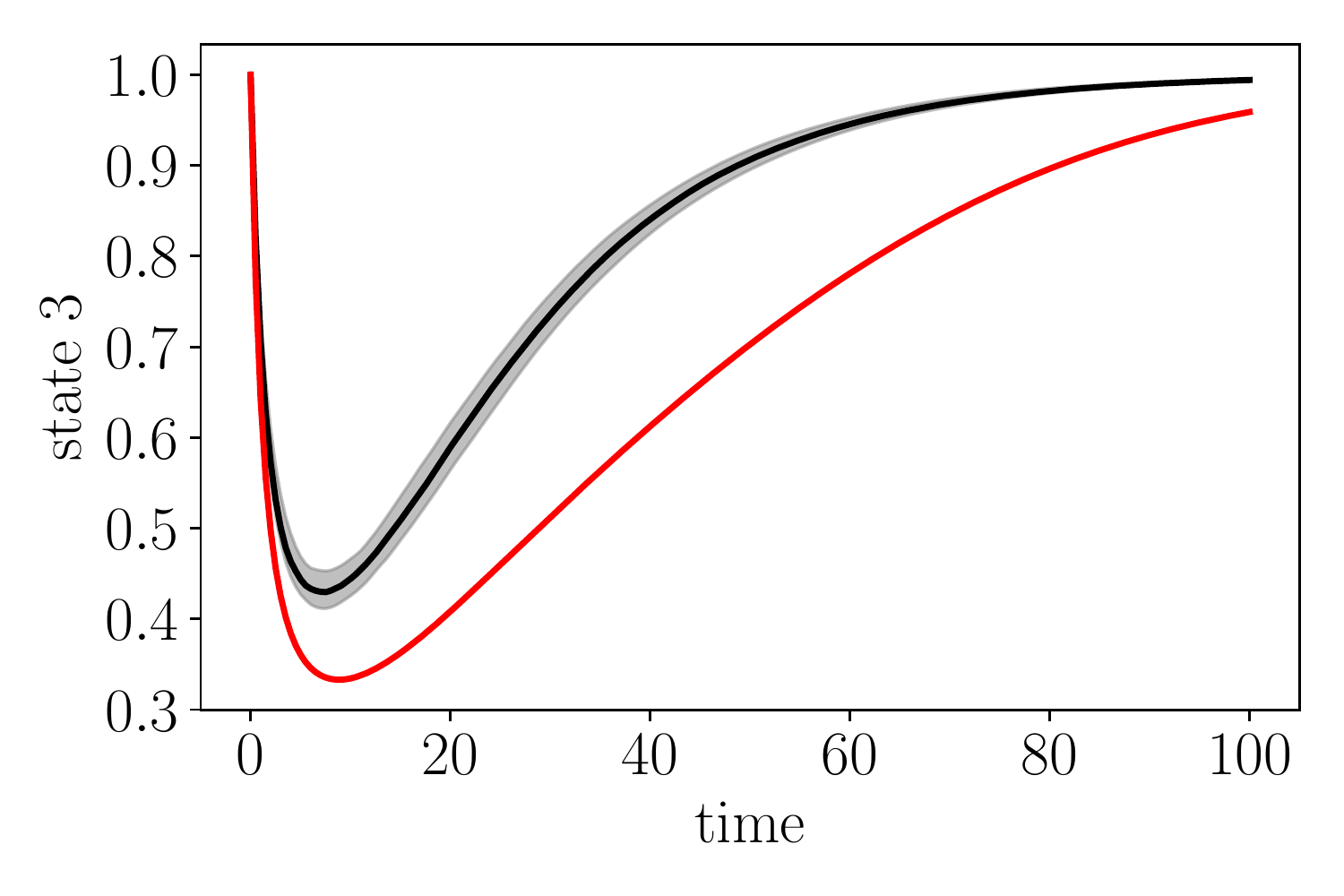}
        \end{subfigure}&
        \begin{subfigure}[t]{.17\textwidth}
            \centering
            \includegraphics[width=\textwidth]{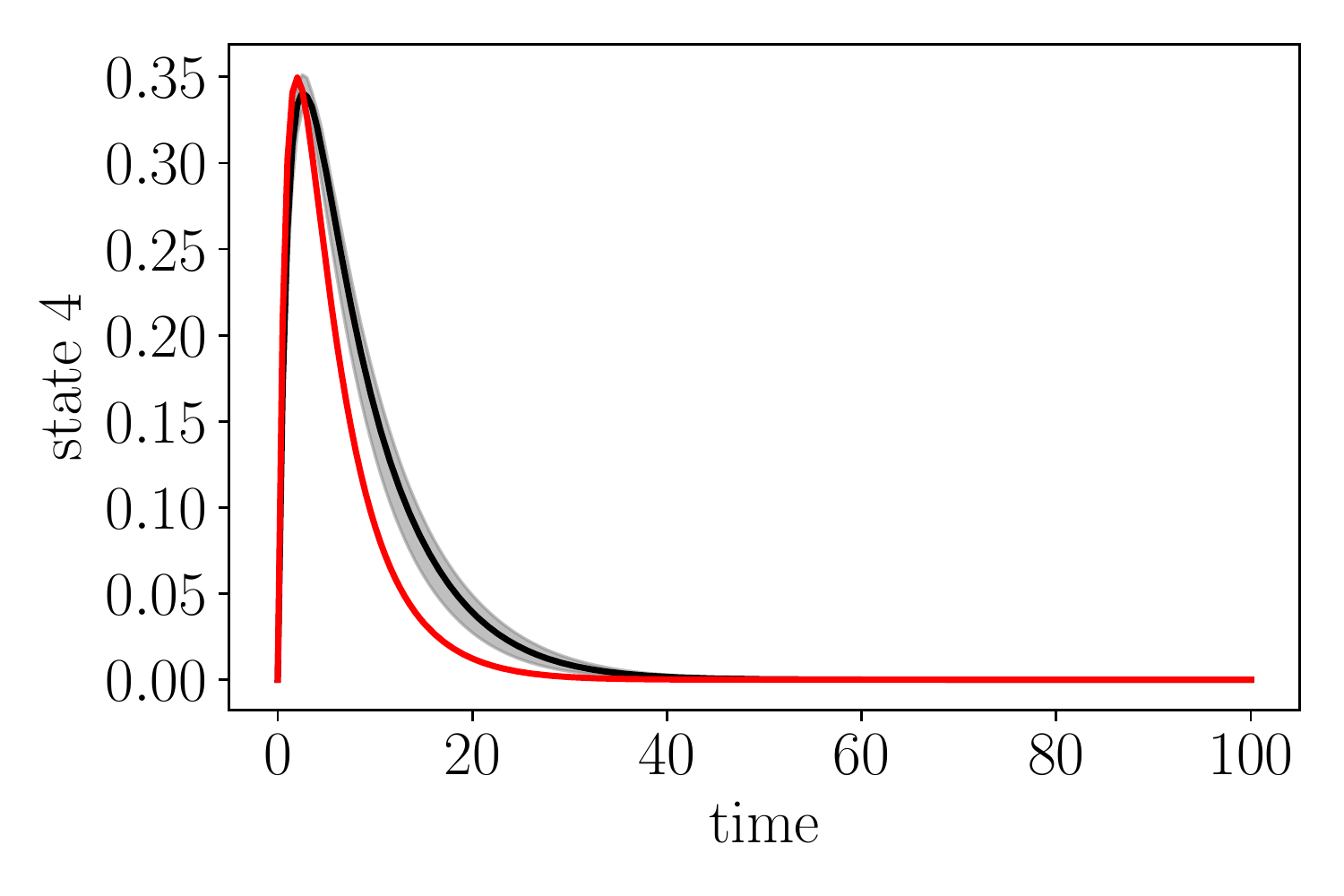}
        \end{subfigure}&
        \begin{subfigure}[t]{.17\textwidth}
            \centering
            \includegraphics[width=\textwidth]{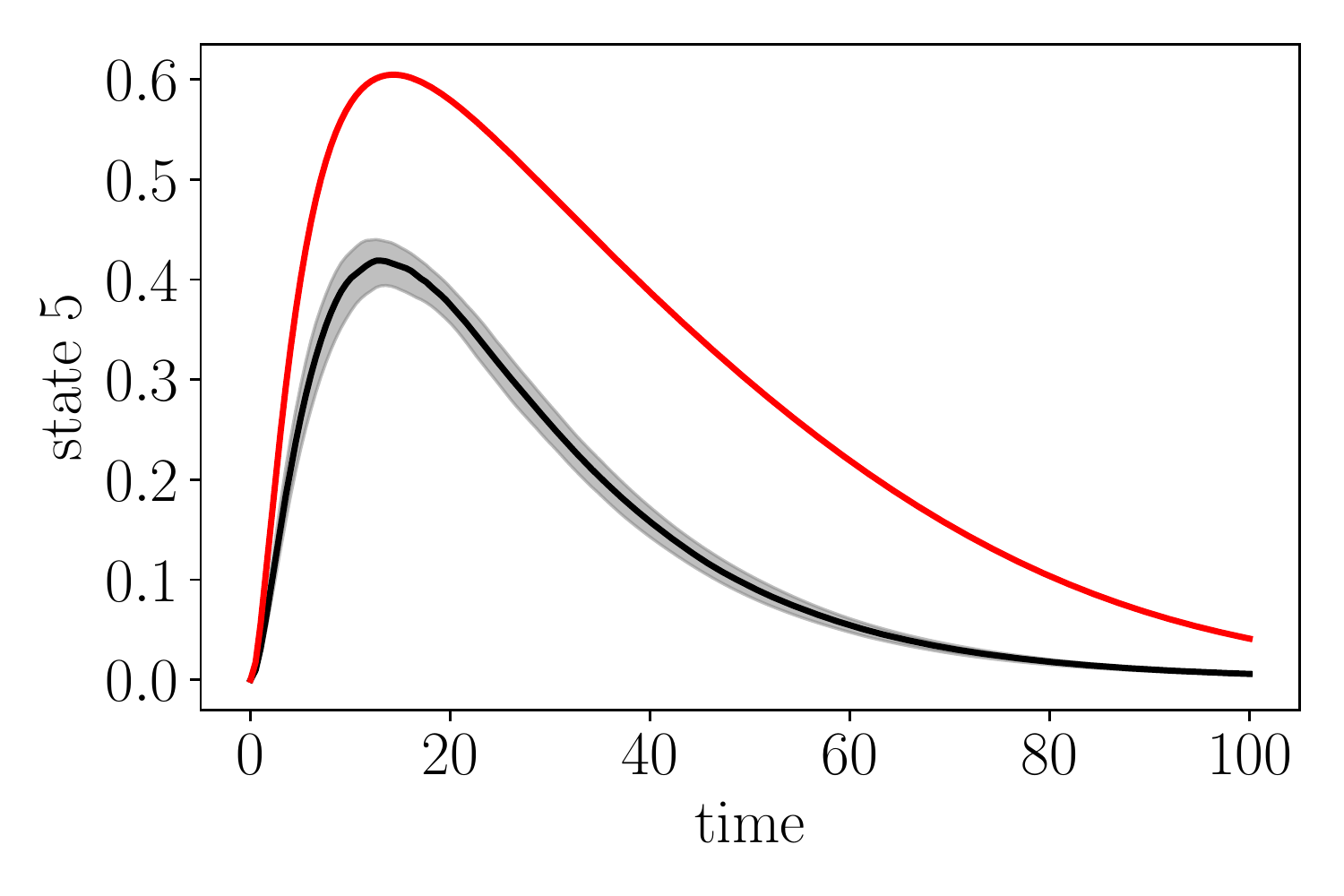}
        \end{subfigure}
   \end{tabular}
    \caption{Median Plots for all states of the most difficult benchmark system in the literature, Protein Transduction, for the high noise case.The red line is the ground truth, while the black line and the shaded area denote the median and the 75\% quantiles of the results of 100 independent noise realizations. $\mathrm{FGPGM}$ (middle) is clearly able to find more accurate parameter estimates than $\mathrm{AGM}$ (top).}
    \label{fig:PTHighNoise}
\end{figure*}

Figure \ref{fig:PTLowNoise} and Figure \ref{fig:PTHighNoise} show median plots for the states obtained by numerical integration of the inferred parameters of the Protein Transduction system.

\begin{figure*}
	\begin{minipage}{\textwidth}
		\centering
		\subcaptionbox{10 observations}{\includegraphics[width=.24\textwidth]{graphs/FHN/new/FHN-1-0.003-100/state_0_FGPGM.png}}
		\subcaptionbox{25 observations}{\includegraphics[width=.24\textwidth]{graphs/FHN/new/FHN-0.5-0.003-100/state_0_FGPGM.png}}
		\subcaptionbox{50 observations}{\includegraphics[width=.24\textwidth]{graphs/FHN/new/FHN-0.25-0.003-100/state_0_FGPGM.png}}
		\subcaptionbox{100 observations}{\includegraphics[width=.24\textwidth]{graphs/FHN/new/FHN-0.1-0.003-100/state_0_FGPGM.png}}
		\subcaptionbox*{}{\includegraphics[width=.24\textwidth]{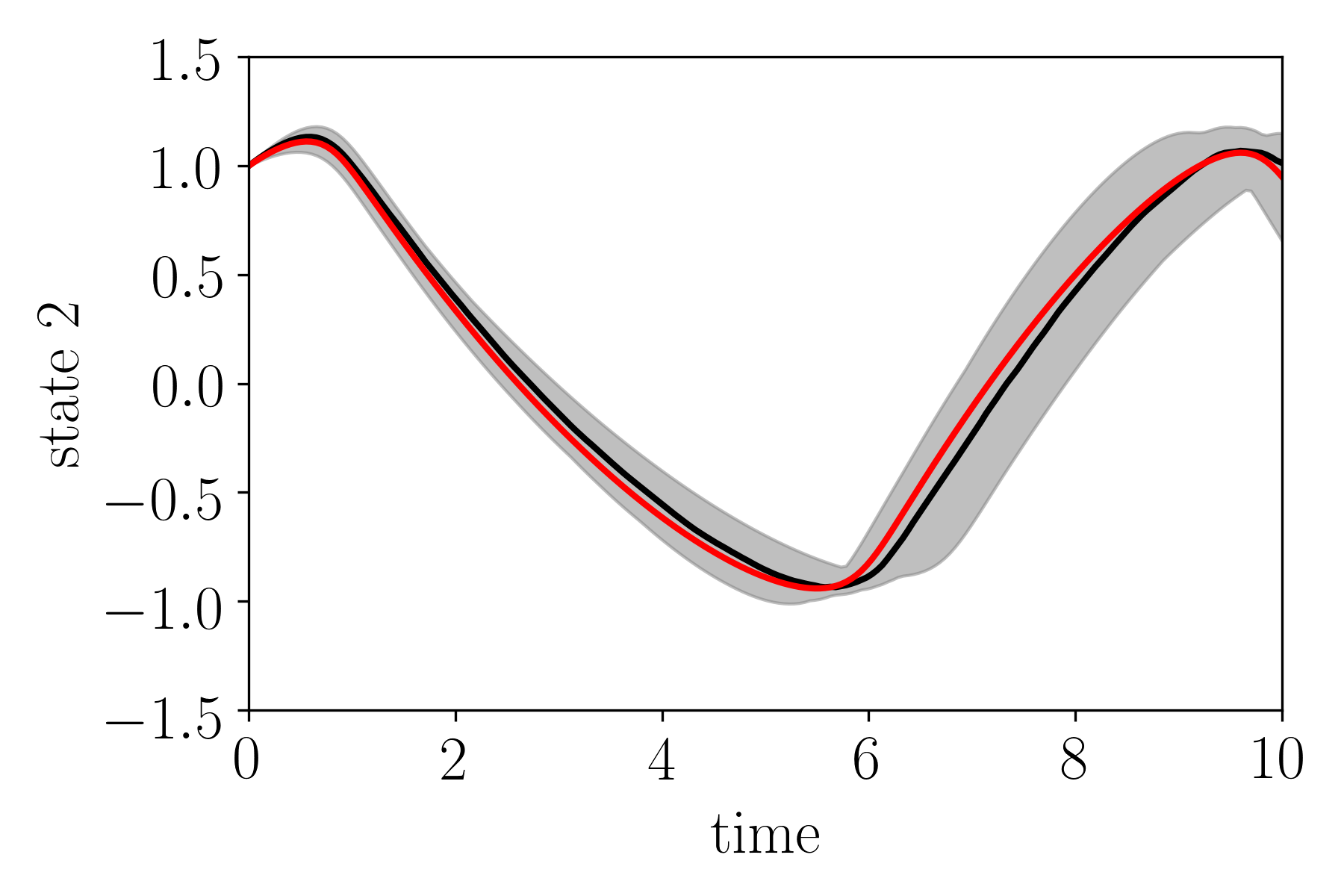}}
		\subcaptionbox*{}{\includegraphics[width=.24\textwidth]{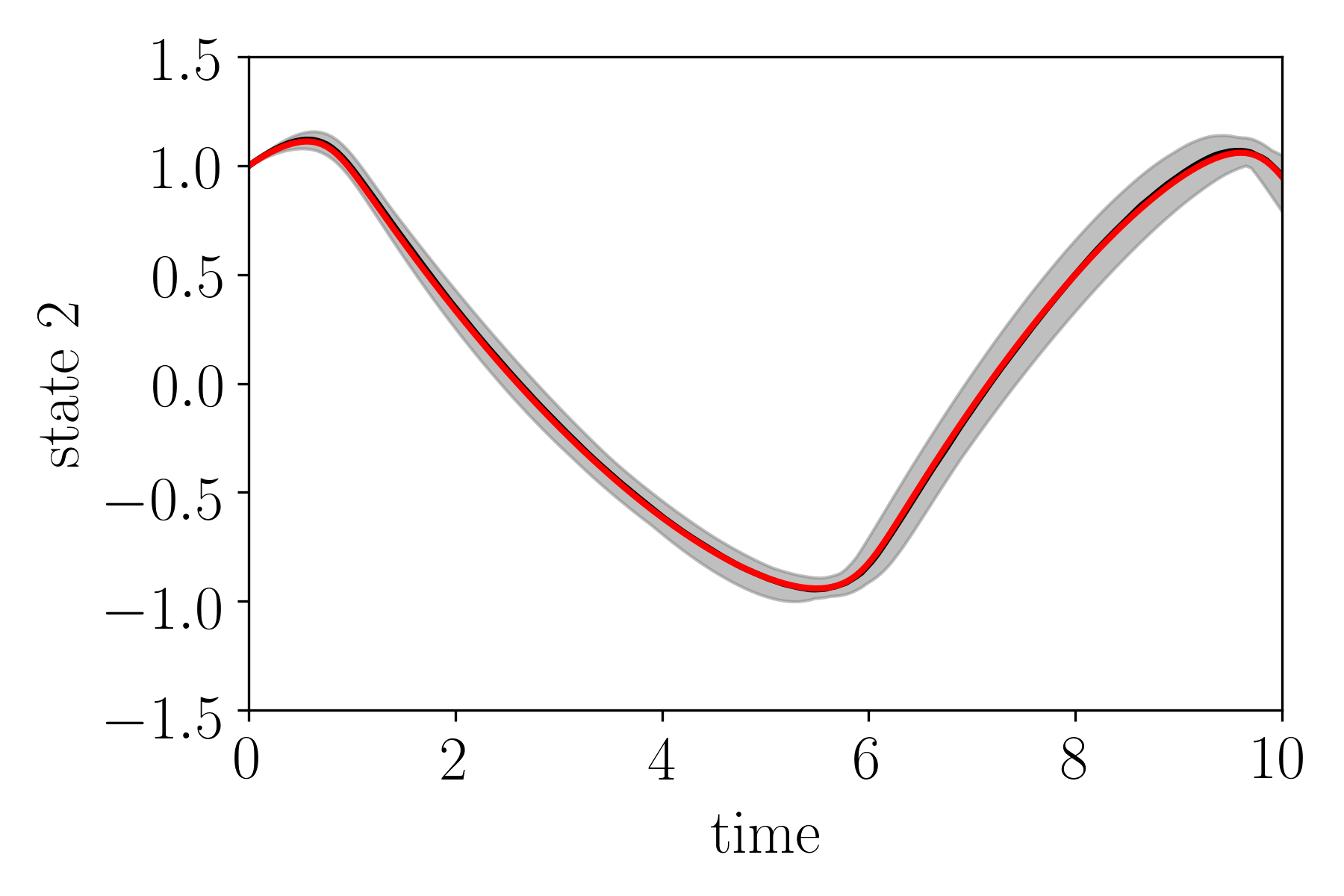}}
		\subcaptionbox*{}{\includegraphics[width=.24\textwidth]{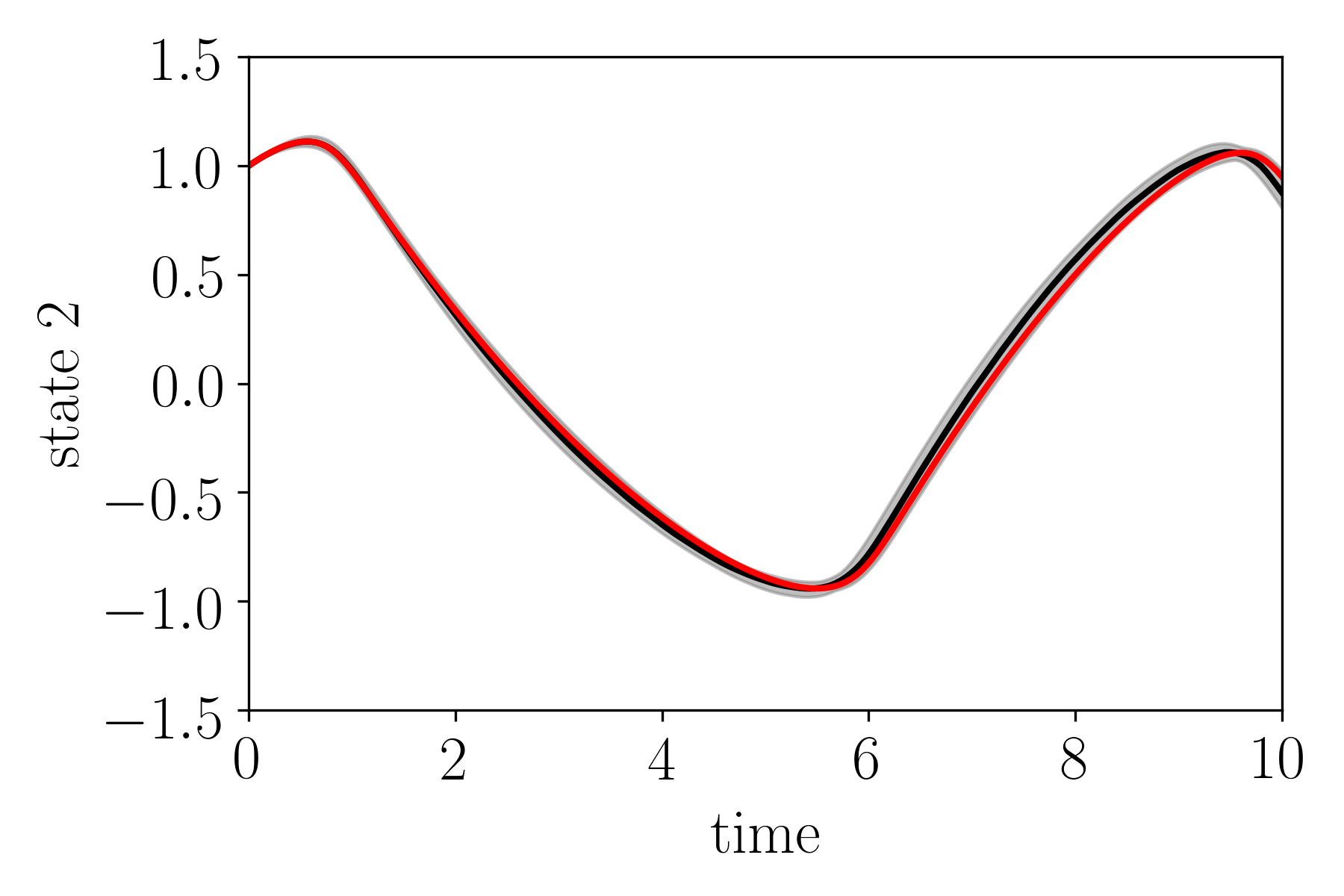}}
		\subcaptionbox*{}{\includegraphics[width=.24\textwidth]{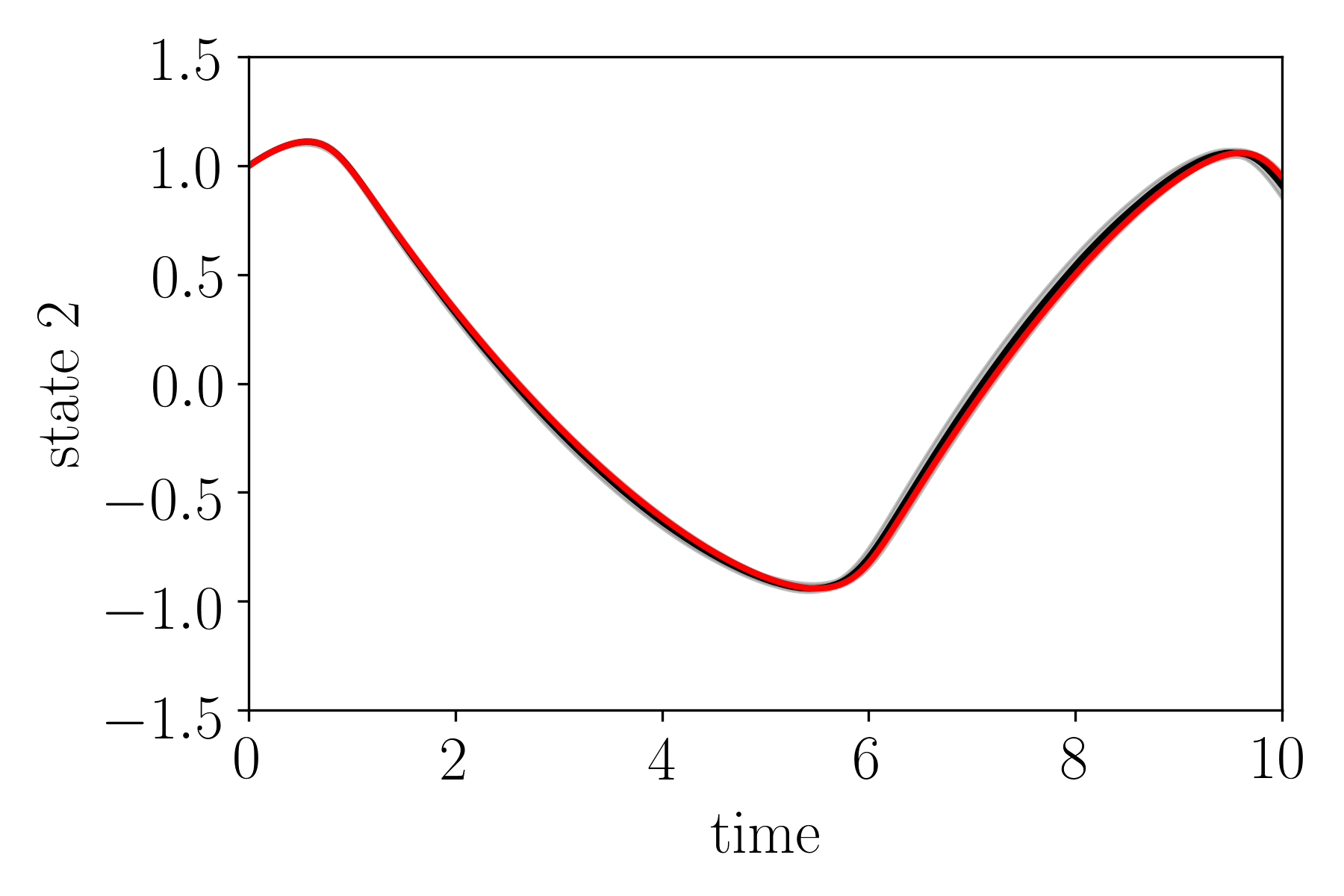}}
		\vspace{-20pt}      
		\caption{Median plots of the numerically integrated states after parameter inference for the FHN system with SNR 10. Ground truth (red), median (black) and 75\% quantiles (gray) over 100 independent noise realizations.}
		\label{fig:FHNLowNoiseAppendix}
	\end{minipage}
\end{figure*}

\end{document}